\documentclass{article}
\usepackage{arxiv}
\pdfoutput=1
\usepackage[utf8]{inputenc} 
\usepackage[T1]{fontenc}    
\usepackage{url}            
\usepackage{booktabs}       
\usepackage{amsfonts}       
\usepackage{nicefrac}       
\usepackage{microtype}      
\usepackage{lipsum}
\usepackage{color}
\usepackage{times}
\usepackage{epsfig}
\usepackage{graphicx}
\usepackage{amsmath}
\usepackage{amssymb}
\DeclareGraphicsExtensions{.pdf,.jpeg,.png}
\usepackage{epstopdf}
\usepackage{multirow}
\usepackage{makecell}

\usepackage{amsmath}
\usepackage{amssymb}
\usepackage{multirow}
\usepackage{subfigure,graphicx}
\usepackage{courier}
\usepackage{booktabs}
\usepackage{latexsym,amsmath,amssymb,graphicx}
\usepackage{url}
\usepackage{color}
\usepackage{bm}
\usepackage{helvet}
\usepackage{algorithm}
\usepackage{algorithmic}

\title{A Triple-Double Convolutional Neural Network for Panchromatic Sharpening}

\author{
Tian-Jing Zhang\\
Yingcai Honors College\\
University of Electronic Science and Technology of China\\
Chengdu, 611731, China\\
\texttt{zhangtianjinguestc@163.com} \\
	\And
  Liang-Jian Deng\protect{\footnote{Corresponding author}}\\
  School of Mathematical Sciences\\
  University of Electronic Science and Technology of China\\
  Chengdu, 611731, China\\
  \texttt{liangjian.deng@uestc.edu.cn}   
\And
Ting-Zhu Huang\\
School of Mathematical Sciences\\
University of Electronic Science and Technology of China\\
Chengdu, 611731, China\\
\texttt{tingzhuhuang@126.com}   
\And

Jocelyn~Chanussot\\
Univ. Grenoble Alpes, Inria, CNRS, Grenoble INP, LJK\\
   Grenoble, 38000, France\\
   \texttt{jocelyn.chanussot@grenoble-inp.fr} \And
   
Gemine Vivone\\
   National Research Council - Institute of Methodologies for Environmental Analysis,\\

   CNR-IMAA, 85050 Tito Scalo, Italy\\
   \texttt{gemine.vivone@imaa.cnr.it} \\
}

\begin{document}
\maketitle

\begin{abstract}
Pansharpening refers to the fusion of a panchromatic image with a high spatial resolution and a multispectral image with a low spatial resolution, aiming to obtain a high spatial resolution multispectral image. In this paper, we propose a novel deep neural network architecture with level-domain based loss function for pansharpening by taking into account the following double-type structures, \emph{i.e.,} double-level, double-branch, and double-direction, called as triple-double network (TDNet).
By using the structure of TDNet, the spatial details of the panchromatic image can be fully exploited and utilized to progressively inject into the low spatial resolution multispectral image, thus yielding the high spatial resolution output. The specific network design is motivated by the physical formula of the traditional multi-resolution analysis (MRA) methods. Hence, an effective MRA fusion module is also integrated into the TDNet. Besides, we adopt a few ResNet blocks and some multi-scale convolution kernels to deepen and widen the network to effectively enhance the feature extraction and the robustness of the proposed TDNet.
Extensive experiments on reduced- and full-resolution datasets acquired by WorldView-3, QuickBird, and GaoFen-2 sensors demonstrate the superiority of the proposed TDNet compared with some recent state-of-the-art pansharpening approaches. An ablation study has also corroborated the effectiveness of the proposed approach.
\end{abstract}


\section{Introduction} \label{sec:intro}

Remote sensing satellites are dedicated to collect image data from the Earth's surface. However, because of some constraints on the signal-to-noise ratio (SNR) for the sensor hardware, we cannot get high spatial and spectral resolutions in a unique acquisition. Thus, satellites, such as, IKONOS, GaoFen, QuickBird, and WorldView-3, usually capture images containing several spectral bands, called multispectral (MS) images, together with panchromatic (PAN) images having high spatial resolution, \textit{i.e.}, containing many image details. Hence, the fusion of these kinds of data is often required to get very high spatio-spectral resolution products. Pansharpening is the fusion of a PAN image and an MS image to obtain the final outcome with the same spatial resolution as the PAN image and the same spectral resolution as the MS image. This research topic has been rapidly developed in recent years and has been proved to be an effective image fusion method~\cite{gemine2019tgrs2}. The results of pansharpening have been widely used in ground object detection, mapping, and image data pre-processing for various high-level applications~\cite{souza2003mapping,wu2017}.

\begin{figure}[!t]
	\begin{center}
		\includegraphics[width=3.2in,height=3.3in]{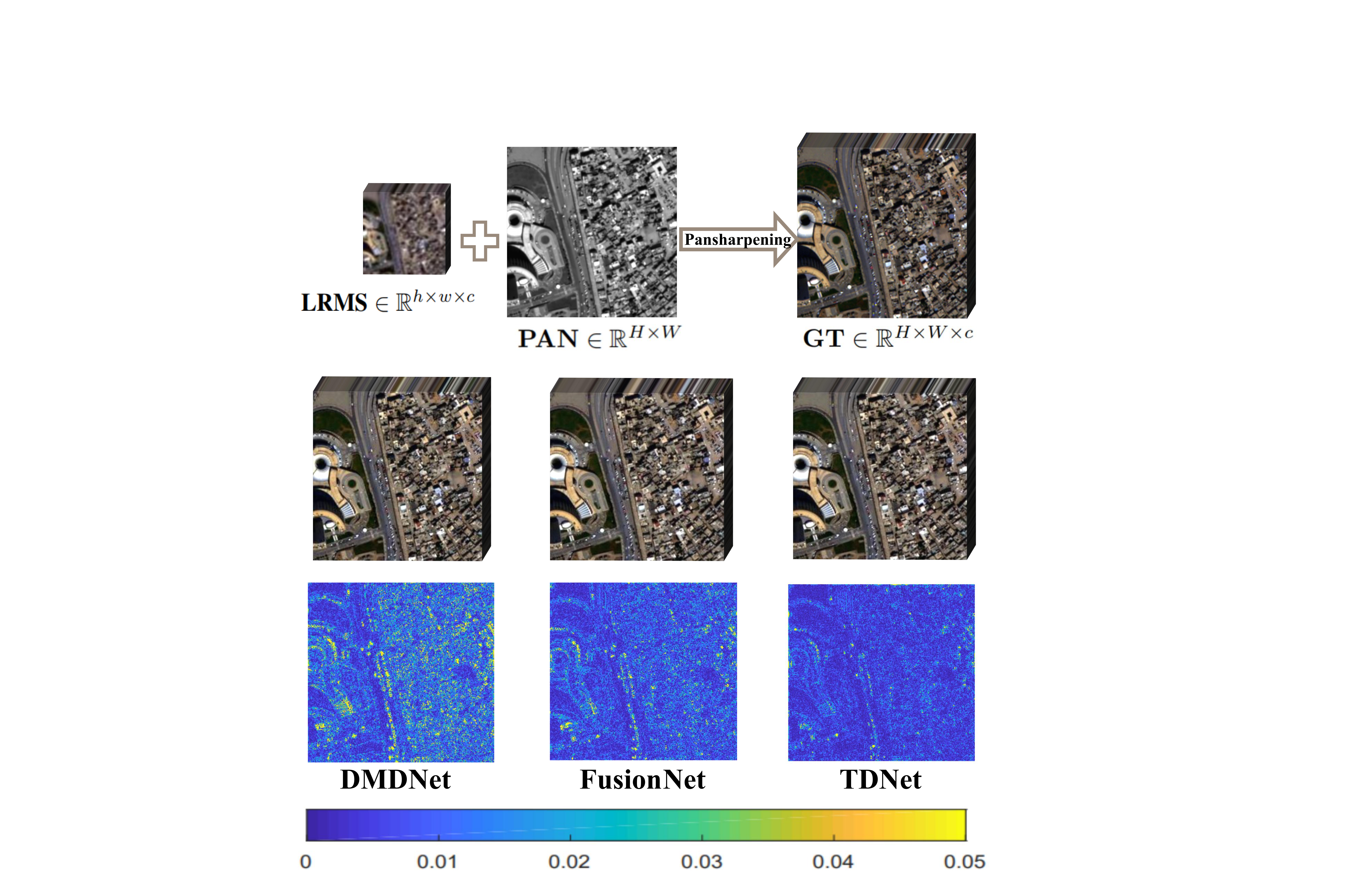} 
	\end{center}
	\caption{First row: pansharpening on a WorldView-3 dataset. This row includes the low spatial resolution MS (LRMS) image, the PAN image, and the desired ground truth (GT) image. Second row: the pansharpened results by three representative DL-based methods, \textit{i.e.}, DMDNet (SAM/ERGAS/Q8=2.9355/1.8119/0.9690)~\cite{fu2020}, FusionNet (SAM/ERGAS/Q8=2.8338/1.7510/0.9714)~\cite{fusionnet}, and the TDNet (SAM/ERGAS/Q8=2.7373/1.6733/0.9764). Third row: the corresponding error maps, which show that TDNet produces less errors than the other two approaches.}
	\label{fig:front}
\end{figure}
Over the past decades, many different approaches have been proposed for the pansharpening problem, and these techniques can be roughly divided into four categories~\cite{thomas2008synthesis,vivone2015critical,MENG2019102}, \emph{i.e.,} component substitution (CS) methods, multi-resolution analysis (MRA) methods, variational optimization (VO) approaches, and deep learning (DL) techniques, respectively. In this work, our approach is based on convolutional neural networks (CNNs), thus belonging to DL techniques. In what follows, we will introduce the representative approaches for each category.

CS-based methods are usually simple approaches belonging to traditional techniques. They project the original MS image into a transformation domain, whose purpose is to simplify the replacement of part or all the spatial information, making easier the replacement of the spatial structure components with the PAN image. It is worth mentioning that many pioneering pansharpening methods are based on the CS philosophy because approaches in this category usually have simple and efficient implementations. Some representative examples into this class are the partial replacement adaptive CS (PRACS)~\cite{pracs}, the Gram-Schmidt (GS) spectral sharpening~\cite{laben2000process}, and the band-dependent spatial-detail with local parameter estimation (BDSD)~\cite{bdsd}. Note that the CS-based methods can generally get products with a better rendering paying it with a greater spectral distortion.

MRA methods are another class of traditional approaches whose goal is to inject the spatial details extracted from the PAN image into the MS image that is interpolated to the size of the PAN image. The MRA-based fused results are superior to those of the CS-based methods considering the spectral quality. However, these methods can easily generate artifacts, thus often introducing spatial distortion. Some methods belonging to this class are, for instance, the smoothing filter-based intensity modulation (SFIM)~\cite{sfim1}, the additive wavelet luminance proportional (AWLP)~\cite{awlp1}, the modulation transfer function generalized Laplacian Pyramid with high-pass modulation injection model (GLP-HPM) \cite{aiazzi2003mtf}, and the modulation transfer function generalized laplacian pyramid with full resolution regression-based injection model (GLP-Reg)~\cite{vivone2018full}.

Unlike the above-mentioned traditional approaches, VO-based methods have been developed by imposing pre-specified prior terms to regularize the underlying high-resolution multispectral (HRMS) image~\cite{fang2013,duran2014,deng2019if}.
These methods show an elegant mathematical formulation and have a good performance on spatio-spectral preservation~\cite{wei2016,deng2017icip,deng2018rs} compared with some state-of-the-art CS and MRA techniques. The main drawback of VO-based methods is the heavy computational burden, including the tuning of many hyper-parameters. Therefore, CS and MRA approaches are still nowadays used for benchmarking purposes.

Recently, deep learning (DL) techniques have gained much attention due to their powerful ability to implicitly learn the priors from big data. Undoubtedly, methods based on deep learning have been widely used in the field of remote sensing images~\cite{lan2020global,mucnn,Hu2021Hyperspectral}. As a newly developed category to solve pansharpening, deep learning requires physical support at a higher level.  The structure design is of critical importance since it is closely related to the performance gain of the model. 
By building a convolutional neural network (CNN) with a certain structure and functional units, ({\it e.g.}, deep residual network~\cite{2017PanNet}, multiscale and multidepth network~\cite{Yuan2018A}), the DL method can reproduce the nonlinear relationship between MS images, PAN images, and ideal fusion images through the training on satellite datasets. The groundbreaking attempt was made by Masi \textit{et al.}~\cite{pnn}, in 2016, with a three-layer CNN designed specifically for pansharpening, achieving promising results. Inspired by PNN, many researchers developed various structures relied upon CNNs. Among them, the residual module in ResNet~\cite{resnet} is widely used for pansharpening~\cite{2017PanNet,fu2020,BAM}. However, the learning process is difficult to be explained and the neural network often gets into the dilemma of vanishing gradient when the parameters are hard to update. In particular, some essential properties and prior information of the images, such as the uniqueness of high-frequency information, and the intrinsic relationship of the spectrum are often ignored by these types of ``black box" deep models, leaving big room for further improvement. Therefore, we argue that the network framework should be designed based on some characteristics of the problem at hand underlining the unique relationships between the input images~\cite{moller,fang2013}.

In this paper, we propose a novel DL approach for pansharpening, which can exploit a multi-scale spatial details strategy, progressively injecting PAN details into the low-resolution MS image. A novel triple-double network (TDNet) structure is designed based on the MRA formulation. The main contributions of this work can be summarized as follows:
\begin{enumerate}
	\item We propose an overall structure of the network with double-level, double-branch, and double-direction, which injects the latent multi-scale spatial details of the PAN image to the MS image in a hierarchical and bidirectional way. Under this framework, we adopted a level-domain-based loss function to pose constraints on multi-level outcomes, which ensure reasonable final fusion results.
	
	\item Following the traditional MRA methods, an MRA block (MRAB) embedded in the TDNet structure is designed. The MRAB can better complete the extraction of structural information from the PAN image. The design of this block structure also introduces the idea of the attention mechanism, which is more flexible and robust than traditional methods.
	
	\item Considering the pansharpening problem, which requires the injection of different objects at various scales, a multi-scale convolution kernel module is adopted to deepen and widen the proposed network to improve the capability of the nonlinear fitting. The results, shown in Fig. \ref{fig:front}, demonstrate the superiority of the proposed method.
\end{enumerate}

The remaining of this paper is organized as follows. In Sect.~\ref{sec:related}, the background and related works will be briefly introduced. The proposed network is presented in Sect.~\ref{sec:net}. Afterwards, the experimental results and discussion are provided in Sect.~\ref{sec:result}. Finally, conclusions are drawn in Sect.~\ref{sec:conclusion}.
\section{Notation and Related Works}\label{sec:related}

\subsection{Notation}\label{sec:notations}
For convenience, the notation used throughout this paper is presented first. The low-resolution multispectral (LRMS) image and the high-resolution (HR) panchromatic (PAN) image are denoted as $\mathbf{MS}\in \mathbb{R}^{h\times w\times c}$ and $\mathbf{P}\in \mathbb{R}^{H\times W}$, respectively. The desired high-resolution multispectral (HRMS) image is defined as $\widehat{\mathbf{MS}}\in \mathbb{R}^{H\times W\times c}$. The multispectral image upsampled at PAN image scale is represented by $\widetilde{\mathbf{MS}}\in \mathbb{R}^{H\times W\times c}$, and the ground truth image is represented as $\mathbf{GT}\in \mathbb{R}^{H\times W\times c}$.

\subsection{Background}\label{sec:background}
As introduced in Sect. \ref{sec:intro}, due to limitations of hardware devices, LRMS and PAN images are only acquired. Considering the goal of pansharpening, which is to generate multispectral images with high spatial resolution, the general fusion formula can be summarized as follows,
\begin{equation} \label{eq:pansharpening}
\mathbf{\widetilde{MS}} =\mathcal{F}_{\theta}(\mathbf{P},~\mathbf{MS}),
\end{equation}
where $\mathcal{F}_{\theta}(\cdot)$ is used to depict the latent relationship between the involved images. The common idea behind many pansharpening approaches (both traditional and DL-based) is to find the befitting way to characterize the relationship between the known LRMS and PAN images and the desired HRMS image.

\subsection{Overview of MRA Methods}\label{sec:MRA}

\begin{figure}[!t]
	\begin{center}
		{\includegraphics[width=0.6\linewidth]{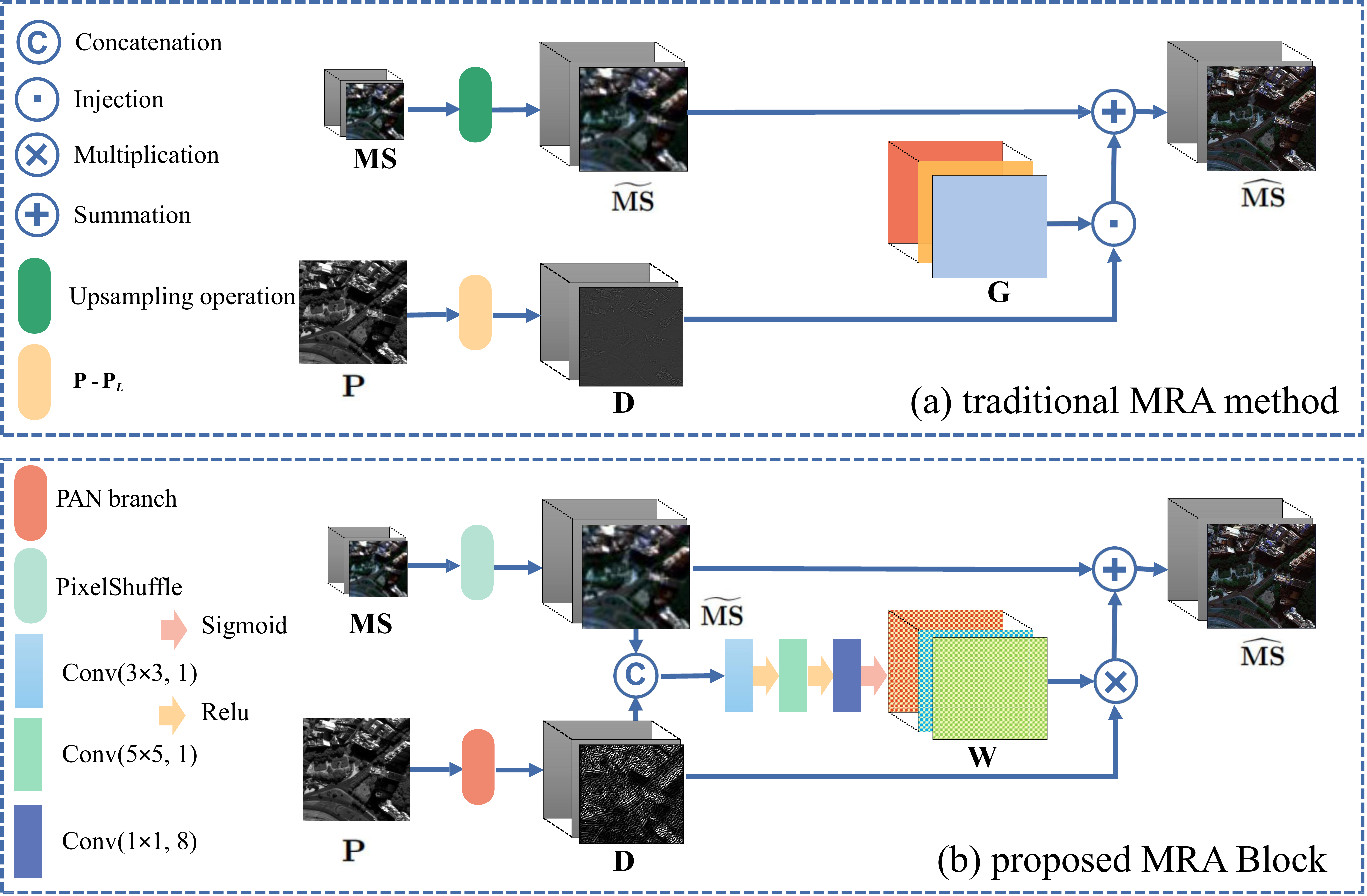}}
	\end{center}
	\caption{(a) The diagram of traditional MRA methods. (b) MRA block (MRAB) was designed based on traditional MRA methods. Please note that the upsampling operation in (a) is a polynomial kernel with 23 coefficients~\cite{glp}. 8-bands datasets are considered to define the number of convolution kernels in (b).}\label{fig:mra}
\end{figure} 

Traditional MRA methods competitively perform in pansharpening. The schematic diagram of general MRA methods is shown in Fig.~\ref{fig:mra}(a). It can be seen that the MRA methods have two main processes, \textit{i.e.}, extracting spatial structure details from the PAN image, $\mathbf{P}$, and injecting information obtained from the $\mathbf{P}$ into $\mathbf{\widetilde{MS}}$ through certain strategies. The mathematical formulation of MRA methods is given by:
\begin{equation}\label{eq:mra}
\widehat{\mathbf{MS}} = \mathbf{\widetilde{MS}} + \mathbf{G\odot(P-P_{L})},
\end{equation}
where $\mathbf{G}\in{\mathbb{R}}^{H\times W\times c}$ is the general form of the injection coefficient gain, $\mathbf{P_{L}}$ stands for the low-pass version of the PAN image $\mathbf{P}$, and $\odot$ represents the element-wise multiplication. Please, refer to~\cite{2020A} for more details. In (\ref{eq:mra}), the spatial structure can be obtained by the difference $\mathbf{P - P_{L}}$, where $\mathbf{P_{L}}$ can be obtained by different filters, see \emph{e.g.}, \cite{mra1}, \cite{mra2}, and \cite{mra3}. The related literature also presents various attempts about the detail injection process, see, \emph{e.g.}, \cite{mra42002Liu}, \cite{mra5ATWT}. The traditional MRA methods can preserve the spectral information, but paying it with the possible introduction of spatial distortion.

\subsection{CNNs for Pansharpening}\label{sec:CNN}
Among DL methods for pansharpening, the techniques based on CNNs have been deeply explored thanks to their excellent ability in the feature extraction phase. Existing CNN-based frameworks addressing the pansharpening problem can be roughly summarized by minimizing the following loss function:
\begin{equation}\label{eq:cnn}
\min_{\Theta}{\mathcal{L} = \| \mathbf{GT} - \mathcal{N}(\mathbf{P}, \mathbf{MS}; \Theta) \|} ,
\end{equation}
where $ \mathcal{N}(\cdot; \Theta)$ represents the functional mapping, through the unknown parameter $\Theta$, between the inputs and an ideal HRMS output, and the $\|\cdot\|$  is a function to describe the distance between the outcome of the network (HRMS) and the GT image. The basic structure for pansharpening can be expressed as follows,
\begin{equation}\label{eq:bcnn}
\begin{split}
&\mathbf{C}^{0}=\mathbf{\{P, ~MS\}},\\
&\mathbf{C}^{1}=\sigma({\mathbf{W}^{1}}\otimes{\mathbf{C}^{0}}+\mathbf{b}^{1}),\\
&\mathbf{C}^{n}=\sigma({\mathbf{W}^{n}}\otimes{\mathbf{C}^{n-1}}+\mathbf{b}^{n}),  n = 2, \cdots, L,
\end{split}
\end{equation}
where the initial $\{\mathbf{P,~MS}\}$ generated by concatenation or other strategies is fed into the network as input. $\mathbf{C}^{i}$, $i=1, 2, \cdots, L,$ represents the $i$-th convolution layer with the corresponding weight $\mathbf{W}^{i}$ and bias $\mathbf{b}^{i}$, where $L$ is the total number of layers. ${\sigma(\cdot)}$ is usually a nonlinear activation function, \emph{e.g.,} ReLU. 

Many effective and promising CNNs are proposed for the task of pansharpening based on the above-mentioned strategy. In~\cite{pnn}, a modified super-resolution network that maps relations through a simple three-layer convolution is proposed by Masi \textit{et al.}. Another typical example is PanNet proposed by Yang \textit{et al.} in \cite{2017PanNet}. It considers spectral and spatial fidelity on high-pass features and introduces the ResNet structure to deepen the given network. In \cite{2017MSDCNN}, Yuan \textit{et al.} propose the use of multi-scale convolution kernels to extract features on different image scales achieving satisfactory results compared with the single-scale convolution kernel. Unlike feeding together $\mathbf{P}$ and $\mathbf{MS}$ into the network, Zhang \textit{et al.} in \cite{2019BDPN} propose a novel network architecture named BDPN, in which $\mathbf{P}$ and $\mathbf{MS}$ are processed using different branches, by exploiting a bi-directional pyramid structure.  

\subsection{Motivation}\label{sec:motivation}
Although various CNN-based approaches have achieved promising results, there is still room for improvements, \emph{e.g.,} physically interpretable architectures, the use of multi-scale structures, and so forth. Recently, unlike other methods that take CNNs as black boxes, Deng \textit{et al.} in \cite{fusionnet} propose FusionNet inspired by traditional CS and MRA methods, which motivates us to regard the formula of traditional methods such as MRA as a guide for the design of the proposed network. The module inspired by traditional methods can be embedded into the CNN network to have a better details extraction and injection.

Besides, existing CNN-based techniques do not fully explore and utilize the multi-scale information in the PAN and MS images loosing some possible information in the process of enhancing the LRMS image. This inspires us to focus on the information injection in hierarchical and bidirectional ways, which is the original intention of the triple-double structure.

\section{The Proposed Network}\label{sec:net}

\begin{figure*}[!htp]
	\begin{center}
		{\includegraphics[width=1\linewidth]{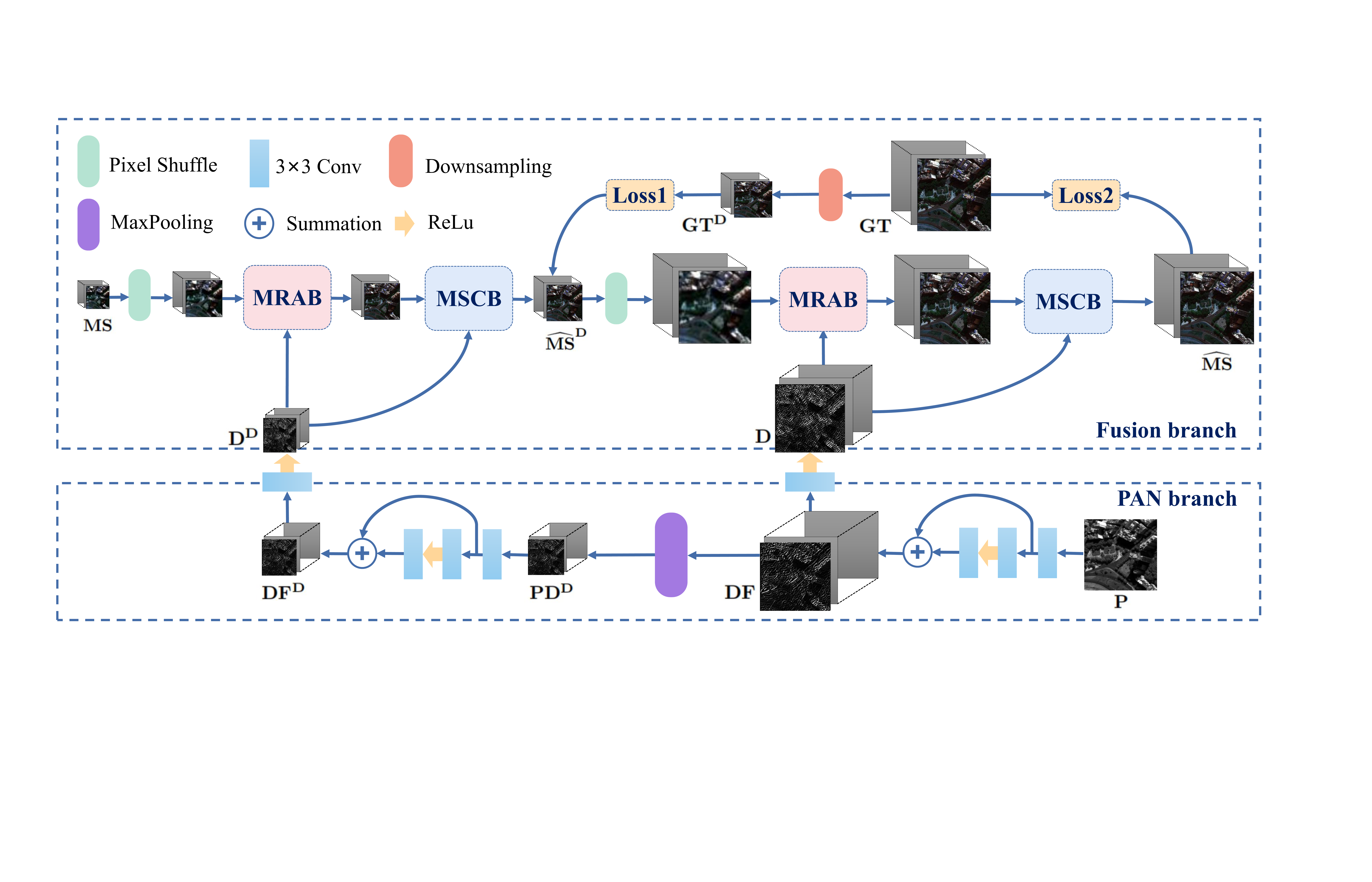}}
	\end{center}
	\caption{Flowchart of the proposed TDNet consisting in two branches, \emph{i.e.,} PAN branch and fusion branch. For convenience, the LRMS image and PAN image are denoted as $\mathbf{MS}$ and $\mathbf{P}$, respectively. The $\widehat{\mathbf{MS}}^{\mathbf{D}}$ is the output of the first-level fusion, and $\widehat{\mathbf{MS}}$ is the final HRMS image. $\mathbf{D}^{\mathbf{D}}$ and $\mathbf{D}$ are the output of the PAN branch. And the ground truth image and its downsampled version are denoted as $\mathbf{GT}$ and $\mathbf{GT}^{\mathbf{D}}$, respectively. The parameters and details of the network can be found in Sect.~\ref{sec:net}. The number of convolution kernels used in the convolution operation is specified in Fig.~\ref{fig:panbranch}.}\label{fig:structure}
	
\end{figure*}

As stated before, our model is inspired by traditional MRA methods, where the spatial structure information extracted from the PAN image was added to the upsampled LRMS image. The overall flowchart of the proposed network has been shown in Fig.~\ref{fig:structure}, which includes the following parts: 1) the MRA block (MRAB), whose structure is based on the MRA general formulation; 2) the multi-scale convolutional feature extraction block (MSCB) is used to further improve the quality of the fused image and to strengthen the learning potential of the network; 3) the triple-double architecture, \textit{i.e.},  double-level, double-branch and double-direction, which can fully utilize the multi-scale information. 

\subsection{MRAB}
Let us focus on the physical MRA formula (\ref{eq:mra}), in which the spatial details to be injected, \emph{i.e.,} $\mathbf{G} \odot\mathbf{(P-P_{L})}$, are extracted only from the PAN image with the proper injection coefficient $\mathbf{G}$. Thus, the traditional MRA approaches can equivalently be represented by the following network architecture,
\begin{equation}\label{eq:mra2}
\begin{split}
&\mathbf{D} = \mathcal{H}(\mathbf{P}),\\
&\widehat{\mathbf{MS}} = \widetilde{\mathbf{MS}} + g(\mathbf{D}),
\end{split}
\end{equation}
where $\mathcal{H}(\cdot)$ is represented by the latent convolution layers, aiming to extract the details $\mathbf{D}$ from the PAN image. Besides, $g(\cdot)$ is represented by a spatial attention simulating the rule of the detail injection coefficient in (\ref{eq:mra}). Furthermore, the upsampled MS image $\widetilde{\mathbf{MS}}$ can be realized by a simple PixelShuffle upsampling operation.
The first formula in (\ref{eq:mra2}) can be viewed as the PAN spatial details, \emph{i.e.,} $\mathbf{P-P_{L}}$, and the second formula in (\ref{eq:mra2}) is equivalent to the MRA formula (\ref{eq:mra}), where $g(\cdot)$ represents the nonlinear relationship among the involved images instead of a linear one as in (\ref{eq:mra}).
In summary, the MRA block (MRAB) consists of three parts: 1) the upsampling of the LRMS image, 2) the extraction of feature maps, and 3) the spatial attention module for detail injection. The detailed information for MRAB can be found in Fig.~\ref{fig:mra}(b).

\subsubsection{Upsampling LRMS Image} In Fig. \ref{fig:mra}(b), the first step is to upsample the original LRMS image to the same size as the GT image. In previous researches for pansharpening, the LRMS image is usually upscaled by an interpolation or a deconvolution operation. In \cite{PixelShuffle},  Shi \textit{et al.} propose an efficient sub-pixel convolution operation (referred to as PixelShuffle), which learns a group of filters to upscale the low-resolution features into the high-resolution output. PixelShuffle got high performance when applied to the single image super-resolution problem \cite{PixelShuffle}. Therefore, we introduce PixelShuffle into our model to upscale LRMS images to reach better performance. In particular, the feature map with $c \times {r^{2}}$ channels (where $r$ is the upscaling factor between LRMS and PAN images) is obtained through convolution, then yielding the high-resolution image by the periodic shuffling.

\begin{figure}[!htp]
	\begin{center}
		{\includegraphics[width=0.6\linewidth]{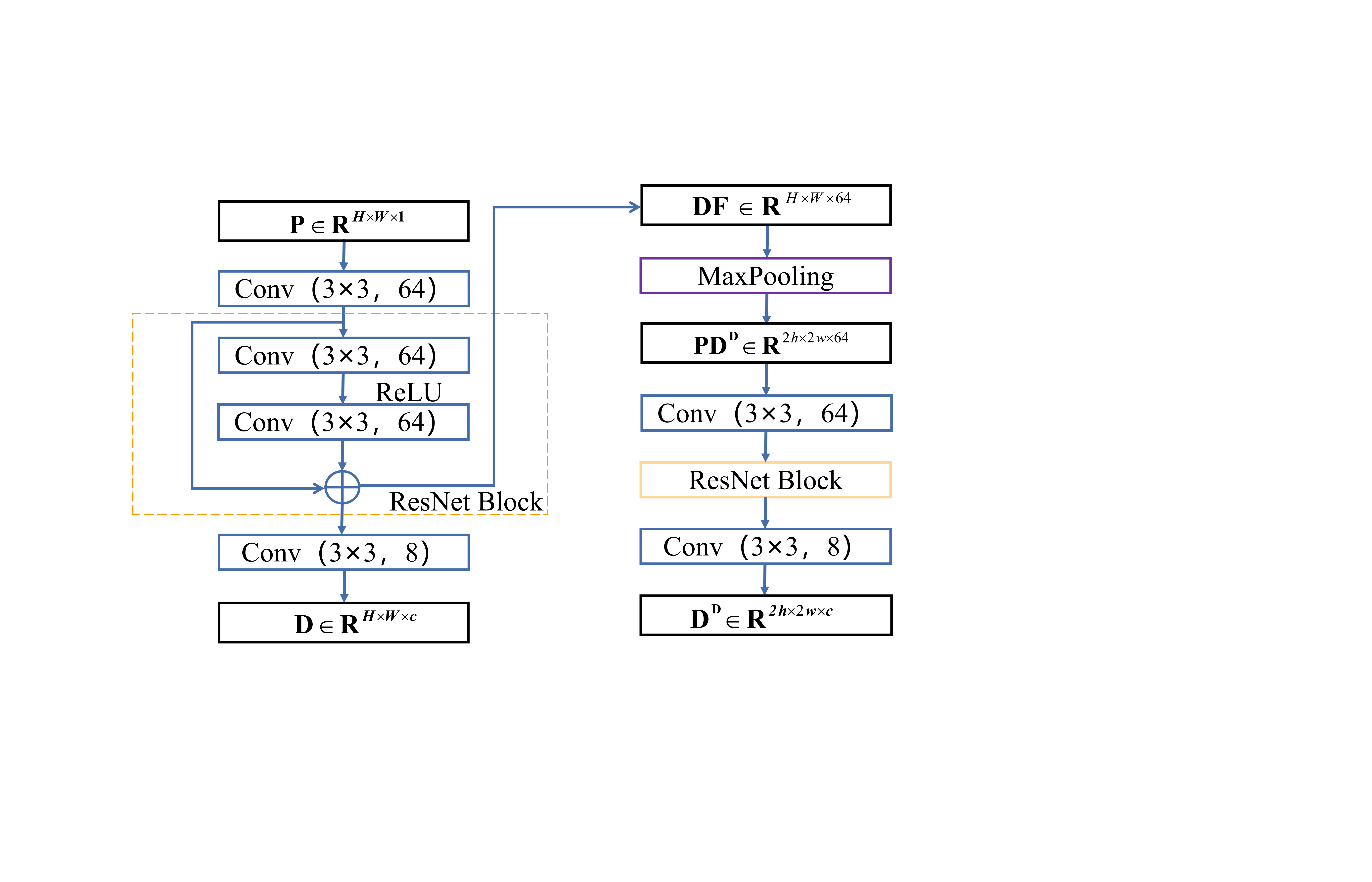}}
	\end{center}
	\caption{An overview of the PAN branch, see the bottom of Fig.~\ref{fig:structure}. $\mathbf{DF}$ is a feature map containing detailed information using 64 channels. $\mathbf{D}$ is a feature map with the same number of channels as $\mathbf{GT}$. The output of the MaxPooling is denoted as $\mathbf{PD^{D}}$, and the feature map with detailed information with reduced size is denoted as $\mathbf{D^{D}}$. Please, note that the number of convolution kernels is related to an exemplary fusion case involving a 8-bands dataset.}\label{fig:panbranch}
\end{figure}

\subsubsection{Extracting Feature Maps}  As mentioned above, the traditional MRA methods extract details calculating the difference between the PAN image and the low-pass filtered PAN image. Thus, the final result depends on the adopted pre-defined filters which may mechanically discard some desired information. Thanks to the use of a convolutional layer, a set of parameters can be learned and dynamically adjusted to thoroughly explore the specific details and select expected features. Besides, to make the model adapt to the different datasets and to heradicate the misfit problem caused by the fixed filters, we extract end-to-end high-frequency information by learning the mapping $\mathcal{H}(\cdot)$ in (\ref{eq:mra2}). In \cite{2019BDPN}, Zhang \textit{et al.} used ResNet blocks as basic structure for feature extraction. However, to retain more information from the original images and to reduce the computational burden, we only adopted one ResNet block to form the PAN branch in TDNet. 

As shown in Fig. \ref{fig:mra}(b), the extracted details from the PAN image are obtained by the PAN branch depicted in Fig. \ref{fig:structure}. The difference with the traditional MRA approaches is that the MS image is upsampled twice, \textit{i.e.}, using a scale factor of 2 (when $r$ is equal to 4). The detailed information for the PAN branch can be found in Fig. \ref{fig:panbranch}.


\subsubsection{Spatial Attention Module for Detail Injection}	
Recalling the original MRA formula (\ref{eq:mra}) and the MRA-inspired formula (\ref{eq:mra2}), we can remark that the detail image $\mathbf{D}$ multiplied by $\mathbf{G}$ in (\ref{eq:mra}) is equivalent to the spatial attention. Since the injection coefficient $\mathbf{G}$ is generally dependent on $\mathbf{MS}$ and $\mathbf{P}$, it motivates us to design spatial attention involving these two components. Specifically, we concatenate $\widetilde{\mathbf{MS}}$ and $\mathbf{D}$ together to carry out the convolution operation as shown in Fig.~\ref{fig:mra}(b), aiming to learn a weight matrix $\mathbf{W}\in\mathbb{R}^{H\times W\times c}$ containing the sufficient features of the $\mathbf{MS}$ and the $\mathbf{P}$ images. The proposed injection strategy is to multiply the learned feature $\mathbf{D}$ obtained by the PAN branch and the weight matrix $\mathbf{W}$, then adding it to the $\widetilde{\mathbf{MS}}$ generated by PixelShuffle to yield the MRAB output.


\subsection{MSCB}\label{MSCB}
\begin{figure}[!htp]
	\begin{center}
		{\includegraphics[width=0.6\linewidth]{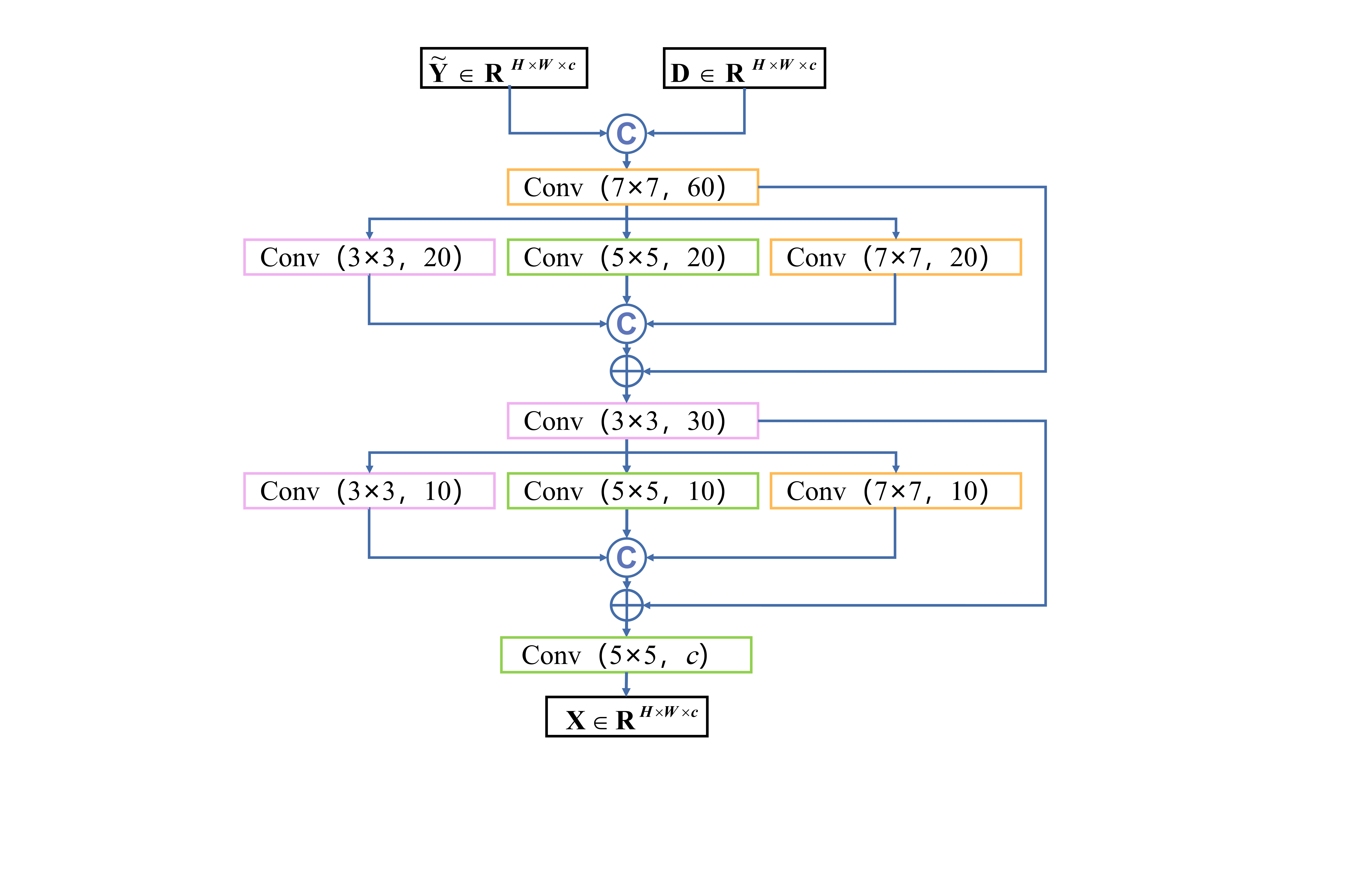}}
	\end{center}
	\caption{An overview of the MSCB. Please, note that since TDNet has a double-level structure, the diagram here refers to only the MSCB at size $H \times W$. The inputs $\mathbf{\tilde{Y}}$ and $\mathbf{D}$ are from MRAB and the feature map of the PAN branch, respectively, and the output $\mathbf{X}$ is the final pansharpened HRMS image, \textit{i.e.}, $\widehat{\mathbf{MS}}$.}\label{fig:mscb}
\end{figure}

Although the MRAB could lead to competitive outcomes with a physical interpretability, the obtained network architecture does not have deep layers, limiting the feature extraction and its nonlinear fitting abilities. Thus, we introduce a multi-scale convolutional block (denoted as MSCB) inspired by Yuan \textit{et al.} \cite{2017MSDCNN} into our model to deepen the network. Fig. \ref{fig:mscb} shows the details of MSCB and its corresponding parameters.


\subsection{Overall Structure of Triple-Double Network}\label{mutilevel}	 
To solve the problem of a different size between LRMS and GT images, conventional methods directly upsample the LRMS image to the GT image size (usually with an upsampling by a factor 4). However, such operation can lead to spatial loss, even causing image distortion. By considering the issues of the size difference and by fully utilizing the multi-scale information, we design the triple-double network (TDNet), \emph{i.e.,} double-level, double-branch, and double-direction. In what follows, we will present the overall structure of the proposed TDNet. An overview of the proposed TDNet is depicted in Fig. \ref{fig:structure}. 


\subsubsection{Double-branch}
From Fig. \ref{fig:structure}, it is clear that the network is divided into two branches, \emph{i.e.,} the PAN branch and the fusion branch. The PAN branch takes the PAN image as the only input. It extracts and represents the multi-scale spatial features, which will be injected into the fusion branch for providing sufficient spatial details. The goal of the fusion branch is instead to fuse the input LRMS image and the multi-scale spatial features from the PAN branch to obtain the final HRMS image. The fusion branch contains some essential strategies mentioned before, such as, MRAB and MSCB.


\subsubsection{Double-level} 
In this work, we upsample the MS image using a two levels strategy, in which the MS image is upscaled to its double size (\emph{i.e.,} with an upscaling factor of 2) for each level, thus exploiting the multi-scale features for pansharpening. In particular, both the PAN branch and fusion branch have the double-level structure for a better ability in resolution enhancement.



\subsubsection{Double-direction} 
Due to the use of double-level, a promising strategy for fully employing multi-scale information of PAN and MS images is to design a network architecture with two directions (called double-direction). As shown in the flowchart of TDNet in
Fig.~\ref{fig:structure}, both the PAN branch and the fusion branch involve the double-level structure. The former downsamples the PAN image to a smaller size, and the latter one upsamples the LRMS image to a larger size. The information flows of the two branches are opposite and correspond to each other, in order to achieve the fusion of information between the branches. A similar strategy has been proven to be effective in a previous benchmark work \cite{2019BDPN}.

In summary, the final architecture of the proposed TDNet has been formulated by the above-mentioned three aspects, \emph{i.e.,} double-branch, double-level, and double-direction. Especially, double-branch takes the known LRMS image and PAN image as input to achieve the distinguishing feature representation. The double-level enables the network to exploit the multi-scale features, and the double-direction reinforces the mutual interaction between the two branches improving the performance.

\subsection{Loss Fuction}\label{lossfuction}
As mentioned before, our TDNet architecture contains a double-level structure, which results in two loss functions. Let $\mathbf{\widehat{{MS}}^{D}}\in \mathbb{R}^{2h\times2w}$ and $\mathbf{\widehat{MS}}\in \mathbb{R}^{H\times W}$ stand for the output of the first and second levels, respectively, and let $\mathbf{GT^{D}}\in \mathbb{R}^{2h\times2w}$ and  $\mathbf{GT}\in \mathbb{R}^{H\times W}$ represent the GT images of the first and second levels, respectively. We define the following loss function for supervised learning for both the levels,
\begin{equation}\label{eq:loss}
\begin{aligned}
\min_{\mathbf{\Theta}} ~\mathcal{L}oss=\gamma\mathcal{L}oss_{1} + (1-\gamma)\mathcal{L}oss_{2},
\end{aligned}
\end{equation}
where $\gamma\in[0, 1]$ is a constant during the training phase, and the magnitude of $\gamma$ is deeply discussed in Sect.~\ref{sec:result}. Specifically, $\mathcal{L}oss_{1}$ and $\mathcal{L}oss_{2}$ are defined as follows,
\begin{equation}\label{eq:loss1}
\mathcal{L}oss_{1}=\frac{1}{N}\sum_{i=1}^{N} {\|\mathbf{GT^{D}}_{i}-\mathbf{\widehat{MS}^{D}}_{i}\|_{1}},
\end{equation}
\begin{equation}\label{eq:loss2}
\mathcal{L}oss_{2}=\frac{1}{N}\sum_{i=1}^{N}\| \mathbf{GT}_i-\mathbf{\widehat{MS}}_i\|_{1},
\end{equation}
where $ \|\cdot \|_{1}$ indicates the $\ell_1$ norm and $N$ is the number of training samples.


\section{Experimental Results}\label{sec:result} 
This section is devoted to the demonstration of the superiority of our TDNet by comparing it with some state-of-the-art pansharpening methods on various datasets. In addition, we comprehensively discuss the structure of our TDNet to explore the inherent potential of the proposed network.

\subsection{Datasets}\label{data}
Three different datasets captured by three different sensors (\textit{i.e.}, WorldView-3, GaoFen-2 and the QuickBird) are considered in this paper. The WorldView-3 works in the visible and near-infrared spectrum range, which provide eight MS-band (coastal, blue, green, yellow, red, red edge, near-infrared 1, and near-infrared 2) and a PAN channel with a spatial resolution of 1.2 m and 0.3 m, respectively. The radiometric resolution of WorldView-3 is 11 bits. Both GaoFen-2 and QuickBird provide four MS-band (red, green, blue, and near-infrared) and a PAN channel. For GaoFen-2, the spatial resolution is about 3.2 m for the MS bands and 0.8 m for the PAN channel, and the radiometric resolution is 10 bits. For QuickBird, the spatial resolution is about 2.44 m for the MS bands and 0.61 m for the PAN channel, and the radiometric resolution is 11 bits.

\subsection{Implementation Details}\label{Sec-Train}
This section is devoted to the presentation of some implementation details related to the proposed approach.

\subsubsection{Dataset Simulation} ~\label{dataset}
In this work, we mainly conduct the network training on WorldView-3 datasets, which are available to download on the website\footnote{\url{http://www.digitalglobe.com/samples?search=Imagery}}. In particular, there are no GT images used as reference. Thus, the original LRMS and PAN images are simultaneously blurred and downsampled according to Wald's protocol \cite{waldprocol1997}. The original LRMS image is used as reference image, \textit{i.e.}, the GT image. We simulate the WorldView-3 dataset with 12,580 samples (also called patch pairs), each sample including PAN (with size $64\times 64$), LRMS (with size $16\times 16\times 8$), and GT (with size $64\times 64\times 8$) patches. For the 12,580 samples, we divided them into 8806/2516/1258 (70\%/20\%/10\%) as for the training dataset, validation dataset, and testing dataset, respectively. 
The simulation process for the datasets is the same as in \cite{fusionnet}. Interesting readers can refer to \cite{fusionnet} for further details. Moreover, we also assess the performance on two 4-bands datasets, \emph{i.e.,}  QuickBird and GaoFen-2. More details about the simulation of these two datasets can be found in Sect. \ref{sec:4-band}.

\subsubsection{Training Platform and Parameter Setting}
The proposed network is coded with Python 3.8.0 and Pytorch 1.7.0, and is trained with NVIDIA GPU GeForce GTX 3080. We use Adam optimizer, in which the betas and weight decay are set as (0.9, 0.999) and 0, respectively, to minimize the loss function (\ref{eq:loss}) by 300 epochs, and the batch size is set as 32. To achieve better performance, we set the initial learning rate as 0.01, then dynamically adjusting it to 0.001 after 220 epochs. About the hyper-parameter $\gamma$ in (\ref{eq:loss}), more discussions can be found in Sect. \ref{sec:gamma}.


\subsection{Benchmark}
Several competitive methods belonging to different pansharpening categories are employed.

\begin{itemize}
	\item EXP: MS image interpolated by a polynomial kernel with 23 coefficients~\cite{glp}
	\item CS methods: 
	\begin{enumerate}
		\item GS: Gram-Schmidt sharpening approach~\cite{laben2000process}.
		\item PRACS: partial replacement adaptive component substitution approach~\cite{pracs}.
		\item BDSD-PC: robust band-dependent spatial-detail approach~\cite{bdsdpc2019}.
	\end{enumerate}
	\item MRA methods: 
	\begin{enumerate}
		\item SFIM: smoothing filter-based intensity modulation~\cite{sfim1}.
		\item GLP-HPM: GLP with MTF-matched filter~\cite{aiazzi2006mtf} with multiplicative injection model~\cite{atwt}.
		\item GLP-CBD: the GLP with MTF-matched filter~\cite{aiazzi2006mtf} and regression-based injection model~\cite{rbim,glp}.
		\item GLP-Reg: the GLP with MTF-matched filter~\cite{aiazzi2006mtf} and full-scale regression~\cite{GlpReg2018}\footnote{\url{http://openremotesensing.net/kb/codes/pansharpening/}}.
	\end{enumerate}
	\item DL-based methods: 
	\begin{enumerate}
		\item PNN: pansharpening via convolutional neural networks~\cite{pnn}\footnote{Note that the given source code in Open Remote Sensing does not contain the trained models for WorldView-2 and WorldView-3, thus we re-implemented the network with default parameters in Python using Tensorflow for fair comparisons.}.
		\item PanNet: convolutional neural networks for residual learning on the high-frequency domain for pansharpening~\cite{2017PanNet}\footnote{Code link: \url{https://xueyangfu.github.io/}}.
		\item DiCNN1: detail injection based convolutional neural networks~\cite{dicnn1}\footnote{DiCNN1 has been implemented by ourselves with default parameters.}.
		\item BDPN: efficient bidirectional pyramid network for pansharpening~\cite{2019BDPN}\footnote{BDPN has been implemented by ourselves with default parameters.}
		\item DMDNet: deep multi-scale detail convolutional neural networks for pansharpening~\cite{fu2020}\footnote{DMDNet has been implemented by ourselves with default parameters.}.
		\item FusionNet: deep convolutional neural network inspired by traditional CS and MRA methods~\cite{fusionnet}.
	\end{enumerate}
	
\end{itemize}

\begin{figure*}[hp]
	\scriptsize
	\setlength{\tabcolsep}{0.9pt}
	\centering
	\begin{tabular}{cccccccc}
		\includegraphics[width=0.85in,height=0.85in]{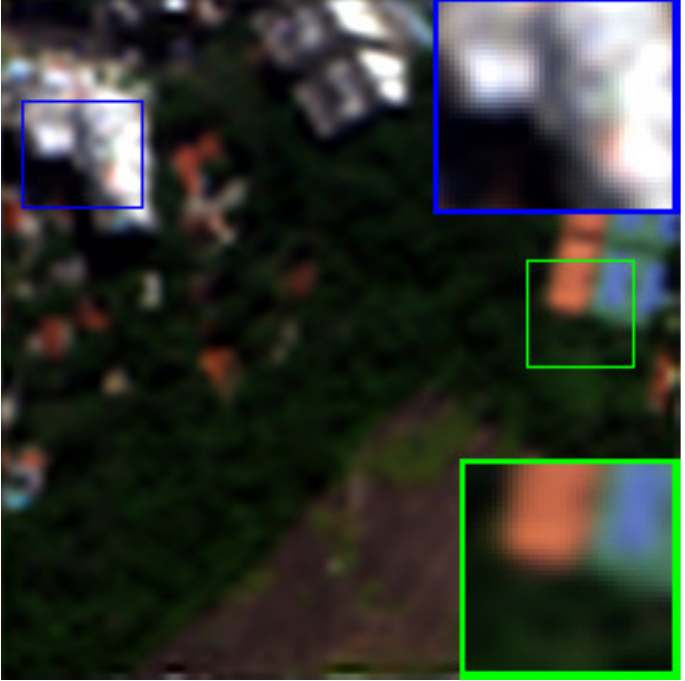}&
		\includegraphics[width=0.85in,height=0.85in]{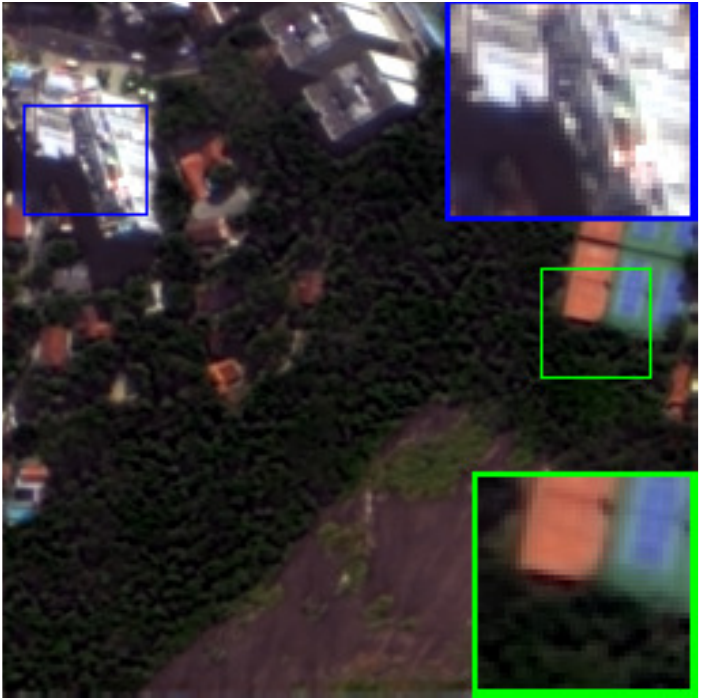}&
		\includegraphics[width=0.85in,height=0.85in]{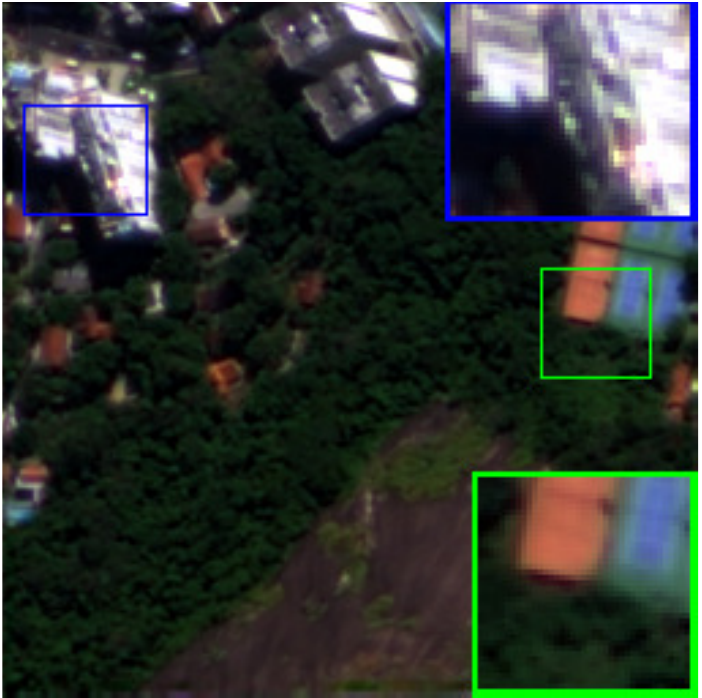}&
		
		\includegraphics[width=0.85in,height=0.85in]{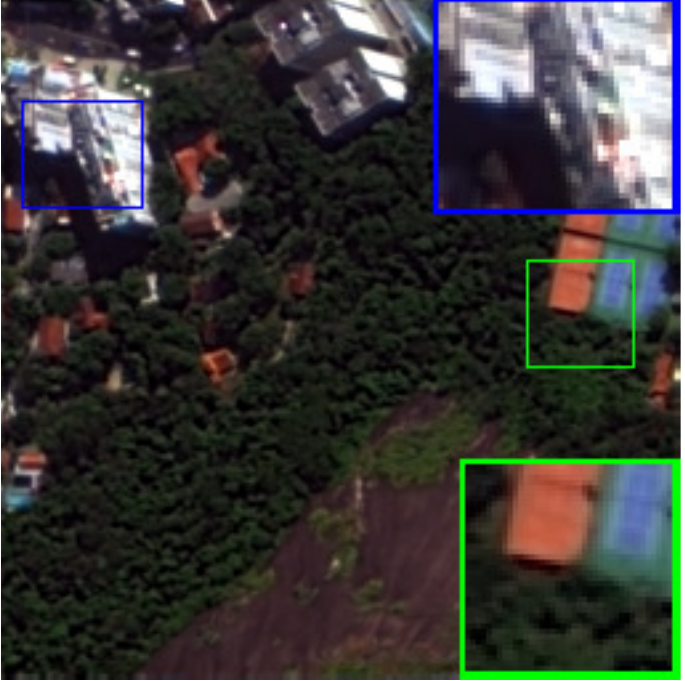}&
		\includegraphics[width=0.85in,height=0.85in]{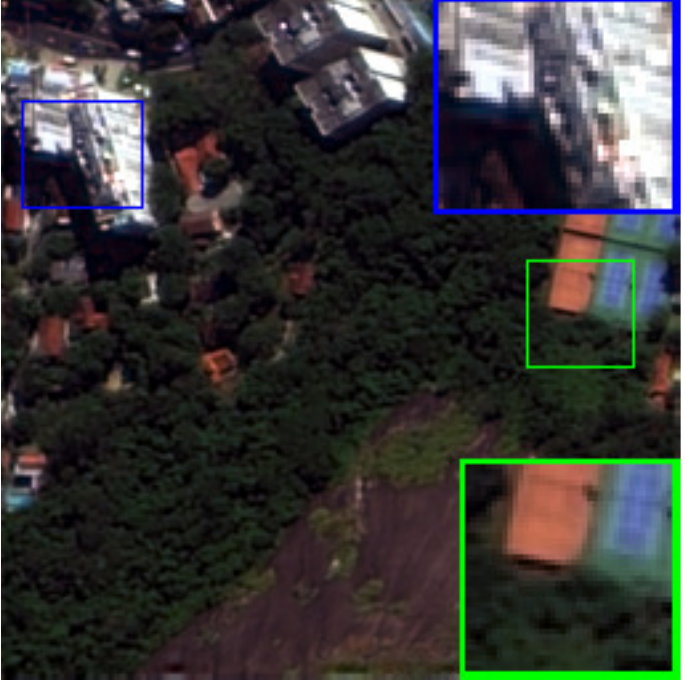}&
		\includegraphics[width=0.85in,height=0.85in]{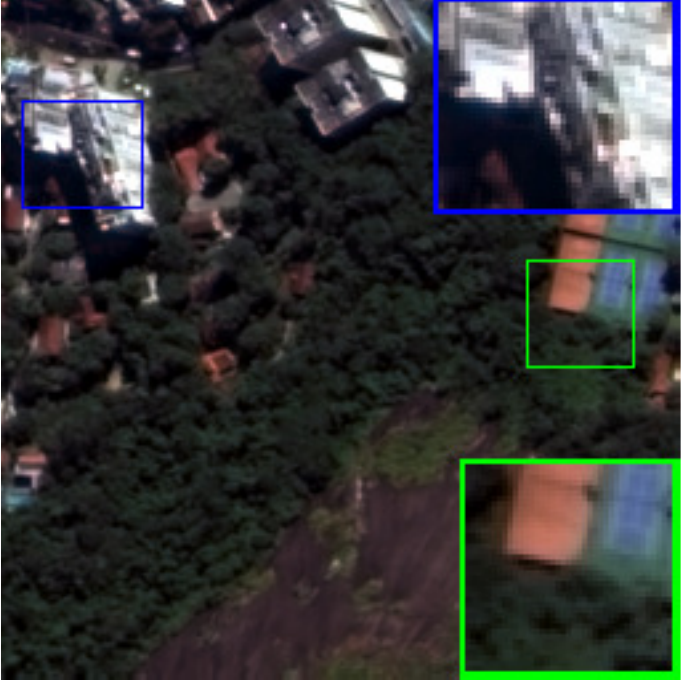}&
		\includegraphics[width=0.85in,height=0.85in]{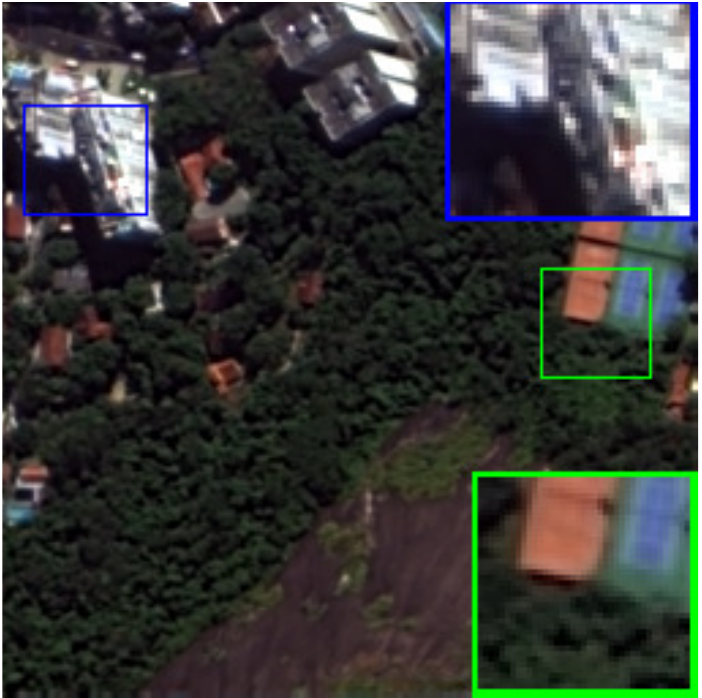}&
		\includegraphics[width=0.85in,height=0.85in]{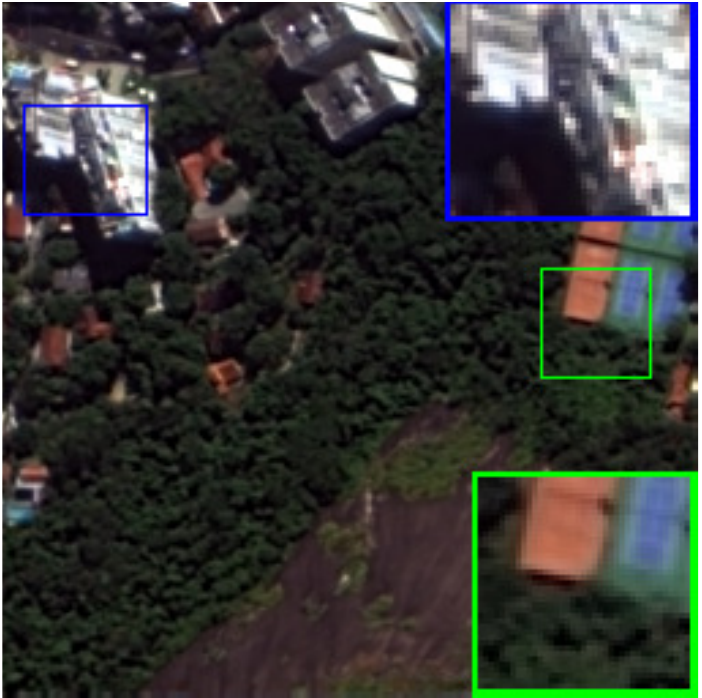}
		\\
		EXP& GS & PRACS& BDSD-PC& SFIM& GLP-HPM& GLP-CBD&GLP-Reg
	\end{tabular}
	\begin{tabular}{cccccccc}
		
		\includegraphics[width=0.85in,height=0.85in]{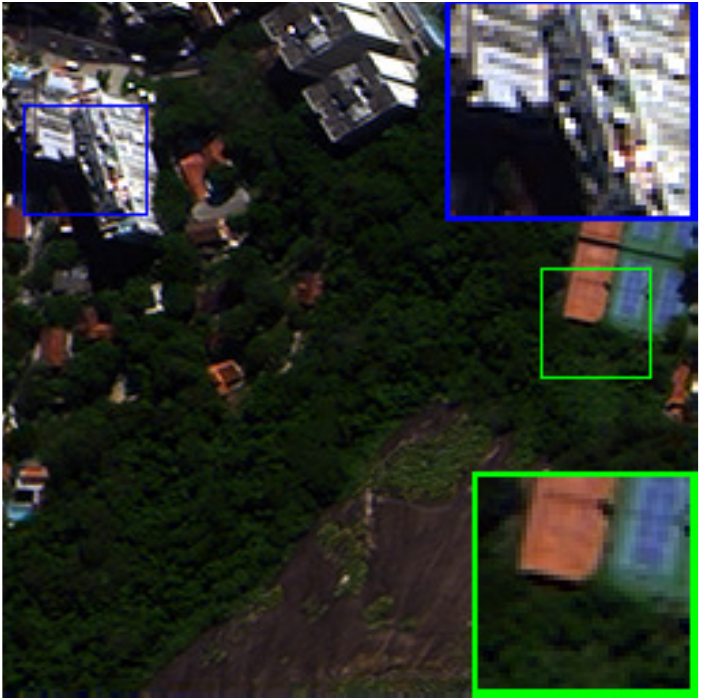}&
		\includegraphics[width=0.85in,height=0.85in]{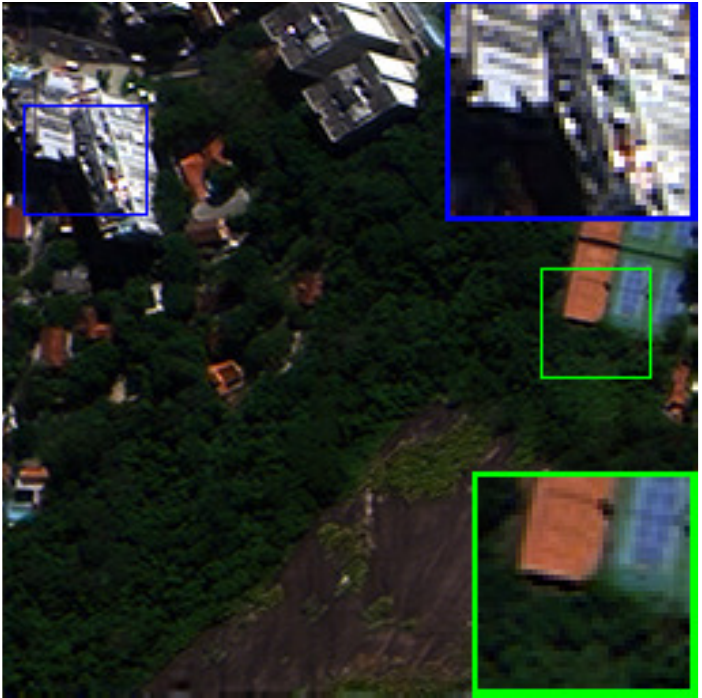}&
		\includegraphics[width=0.85in,height=0.85in]{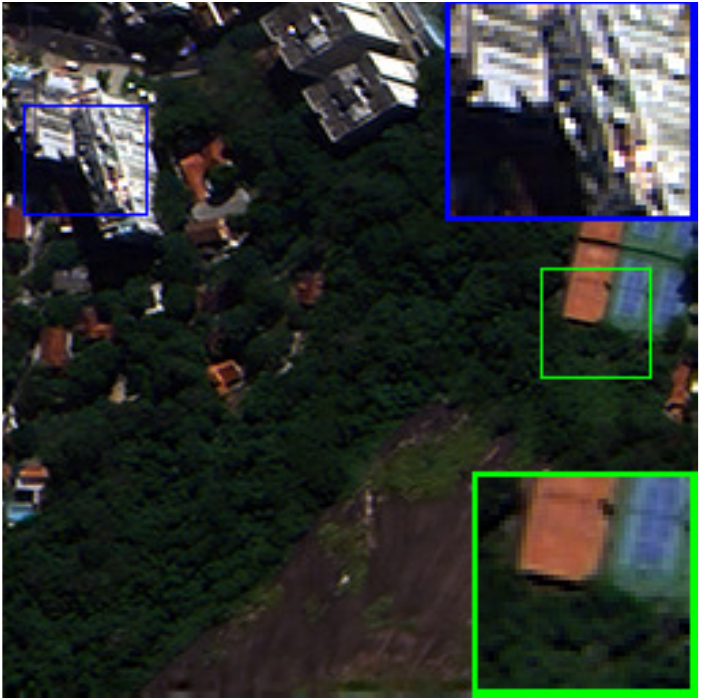}&
		\includegraphics[width=0.85in,height=0.85in]{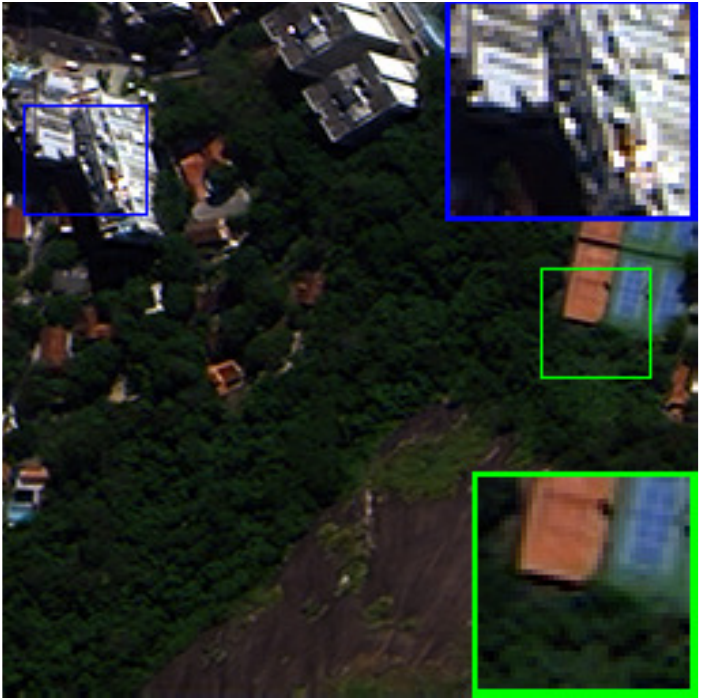}&
		\includegraphics[width=0.85in,height=0.85in]{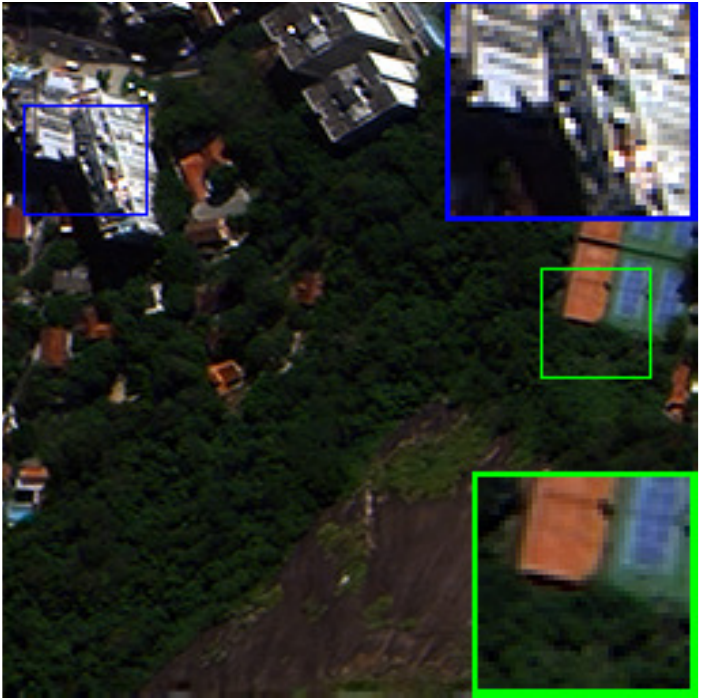}&
		
		\includegraphics[width=0.85in,height=0.85in]{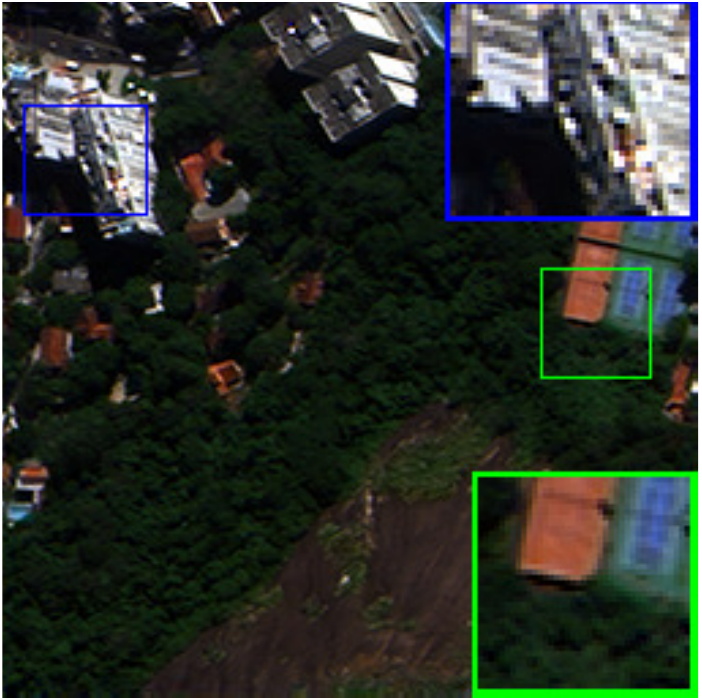}&
		\includegraphics[width=0.85in,height=0.85in]{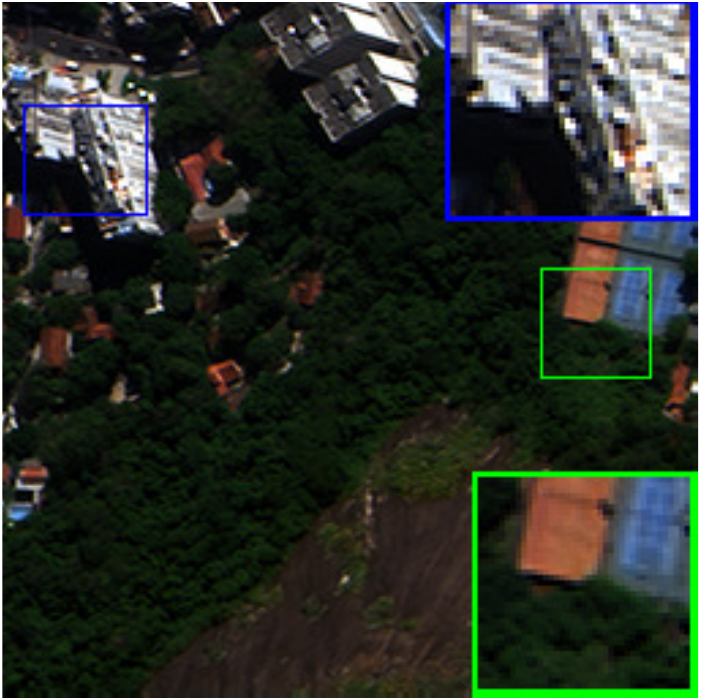}&
		\includegraphics[width=0.85in,height=0.85in]{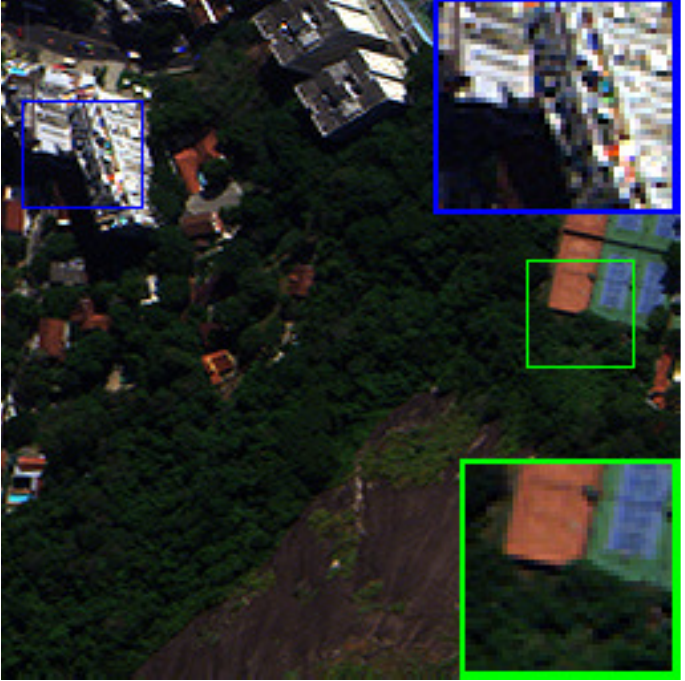}\\
		PNN&DiCNN1& PanNet&BDPN& DMDNet& FusionNet& TDNet& GT
	\end{tabular}
	\caption{Visual comparisons of all the compared approaches on the reduced resolution Rio dataset (sensor: WorldView-3). }
	\label{fig:newdata8}
\end{figure*}
\begin{figure*}[hp]
	\scriptsize
	\setlength{\tabcolsep}{0.9pt}
	\centering
	\begin{tabular}{cccccccc}
		\includegraphics[width=0.85in,height=0.85in]{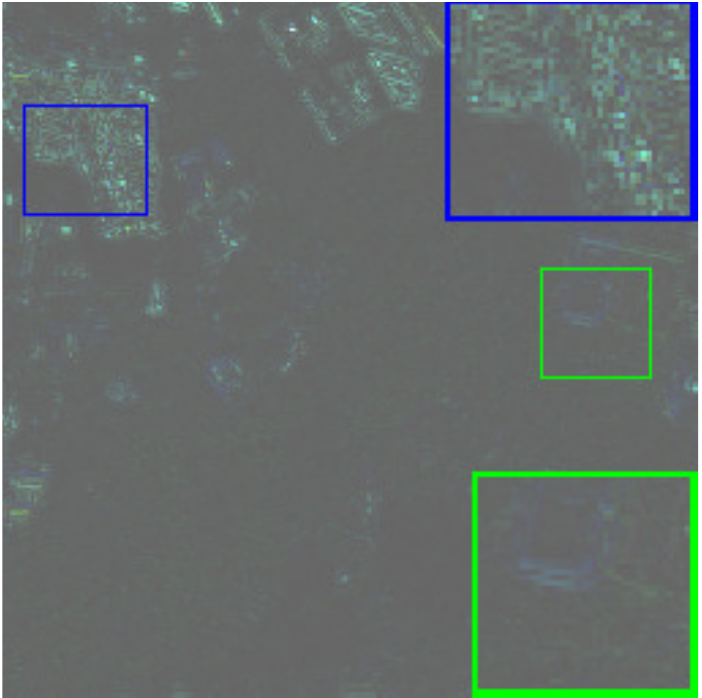}&
		\includegraphics[width=0.85in,height=0.85in]{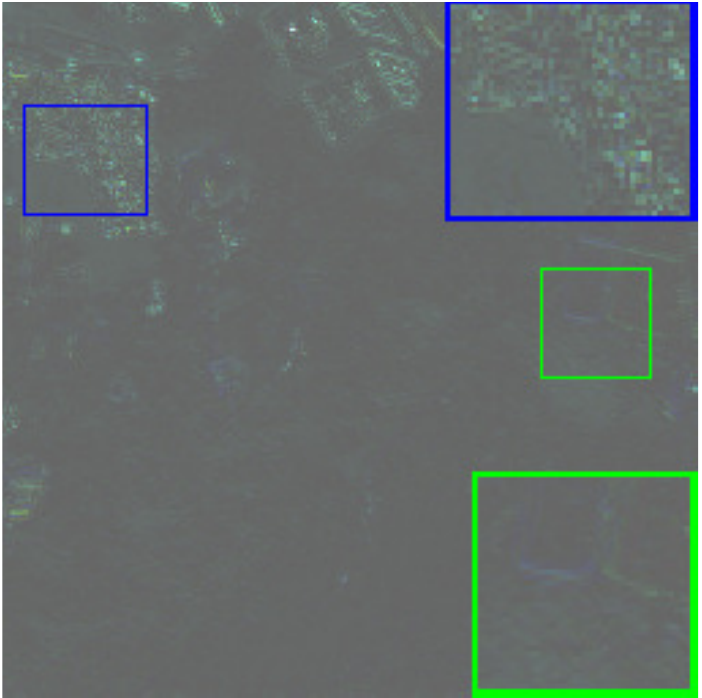}&
		\includegraphics[width=0.85in,height=0.85in]{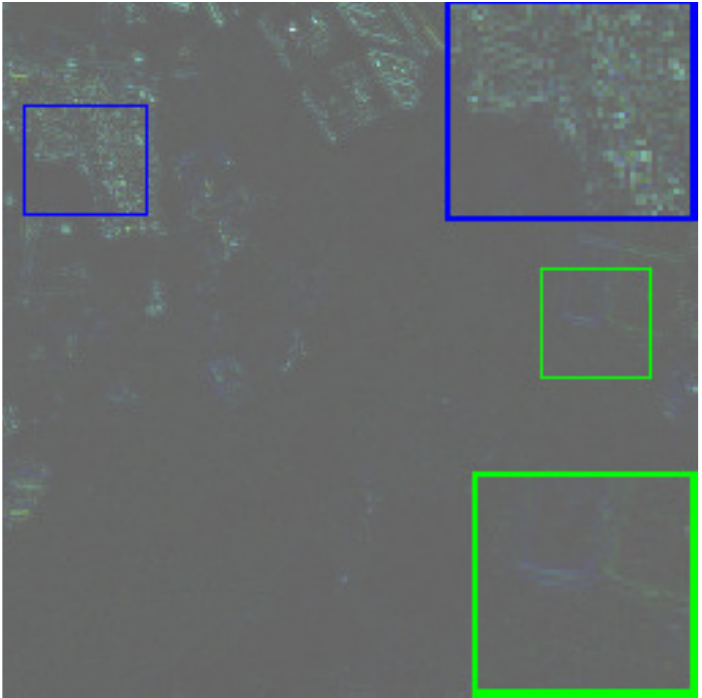}&
		\includegraphics[width=0.85in,height=0.85in]{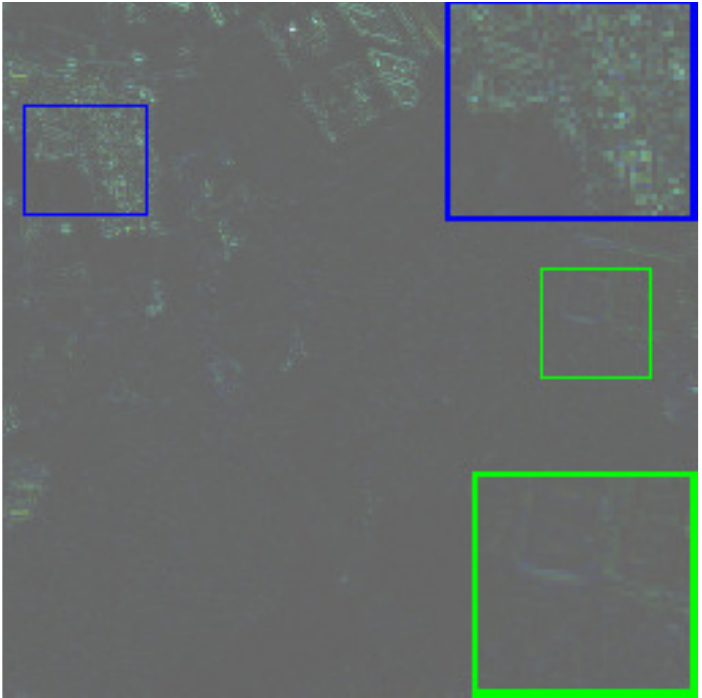}&
		\includegraphics[width=0.85in,height=0.85in]{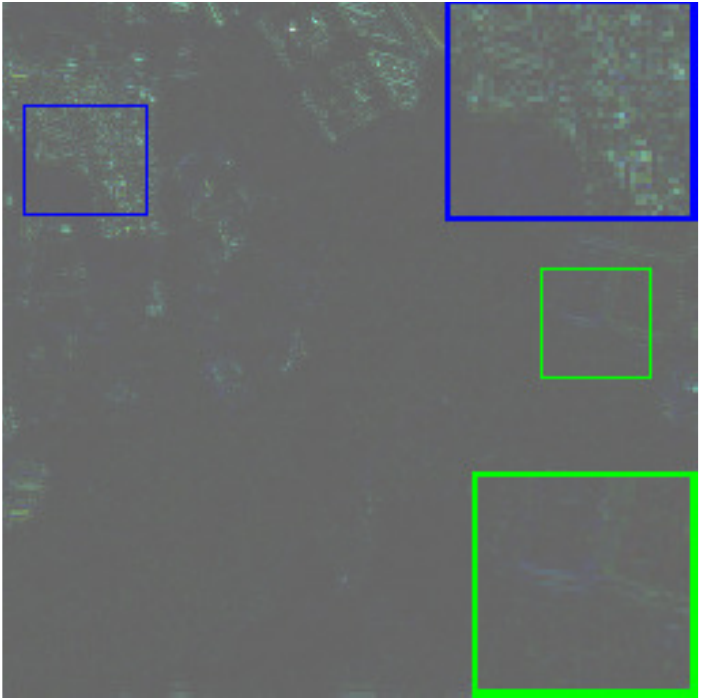}&
		\includegraphics[width=0.85in,height=0.85in]{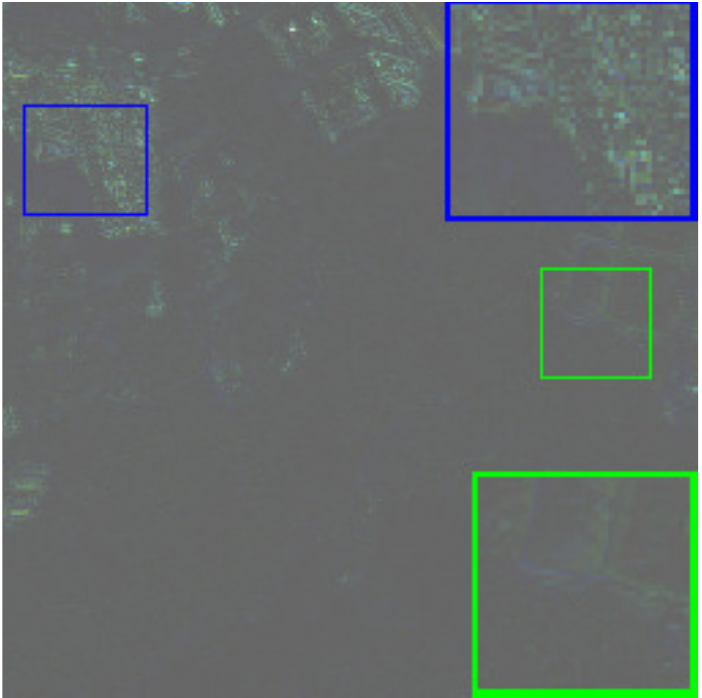}&
		
		\includegraphics[width=0.85in,height=0.85in]{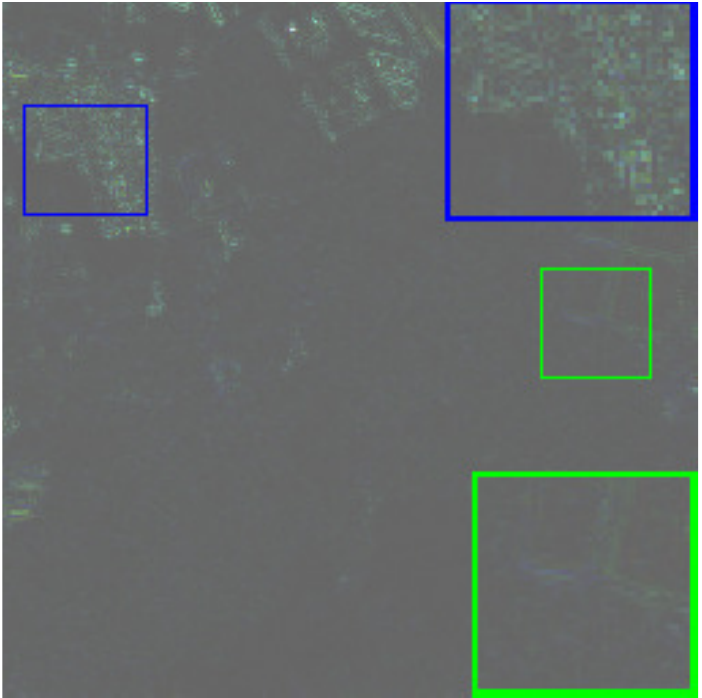}&
		\includegraphics[width=0.85in,height=0.85in]{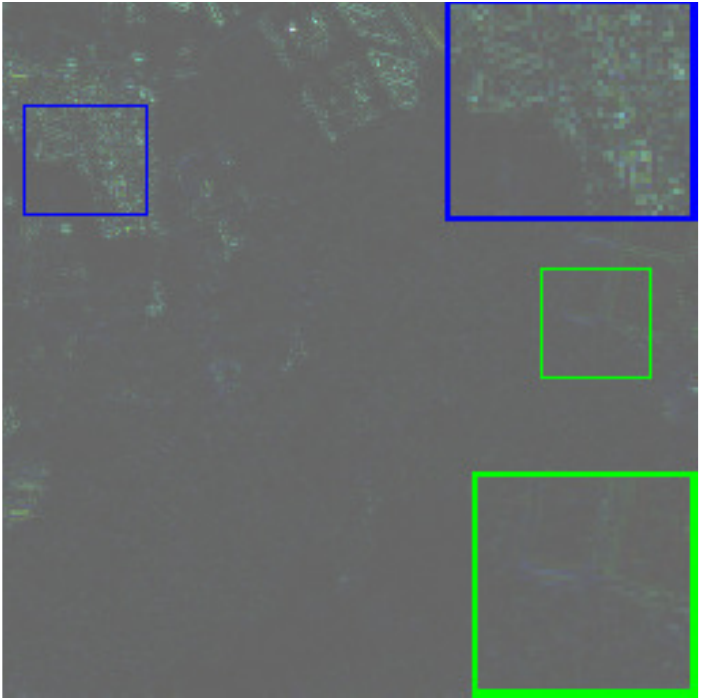}
		\\
		EXP& GS & PRACS& BDSD-PC& SFIM& GLP-HPM& GLP-CBD&GLP-Reg
	\end{tabular}
	\begin{tabular}{cccccccc}
		\includegraphics[width=0.85in,height=0.85in]{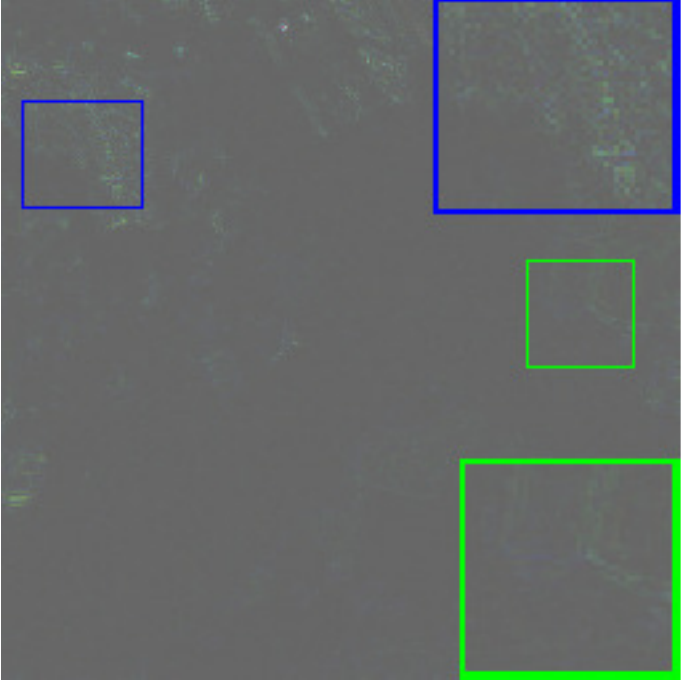}&
		\includegraphics[width=0.85in,height=0.85in]{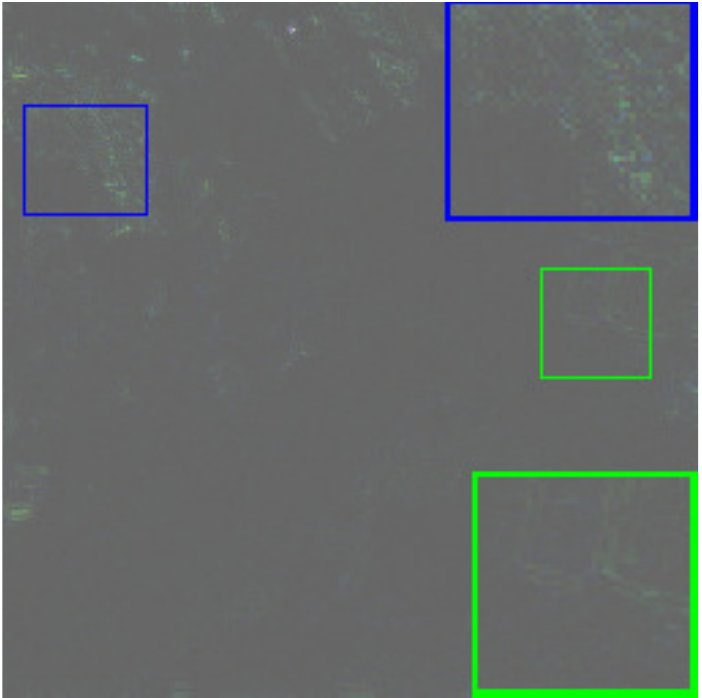}&
		\includegraphics[width=0.85in,height=0.85in]{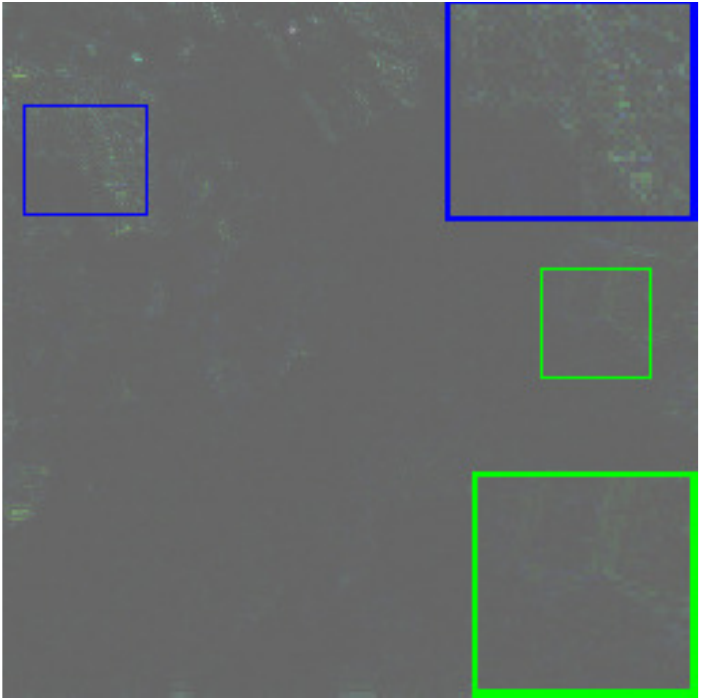}&
		\includegraphics[width=0.85in,height=0.85in]{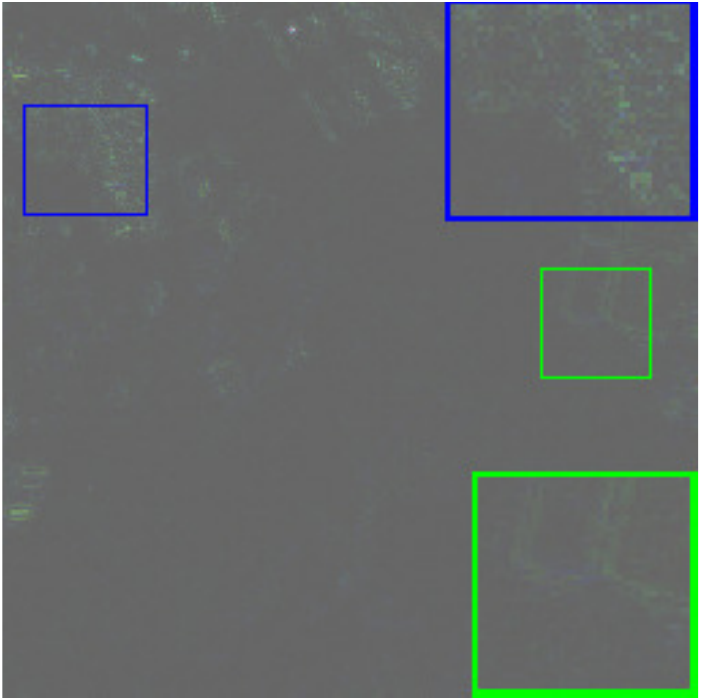}&
		\includegraphics[width=0.85in,height=0.85in]{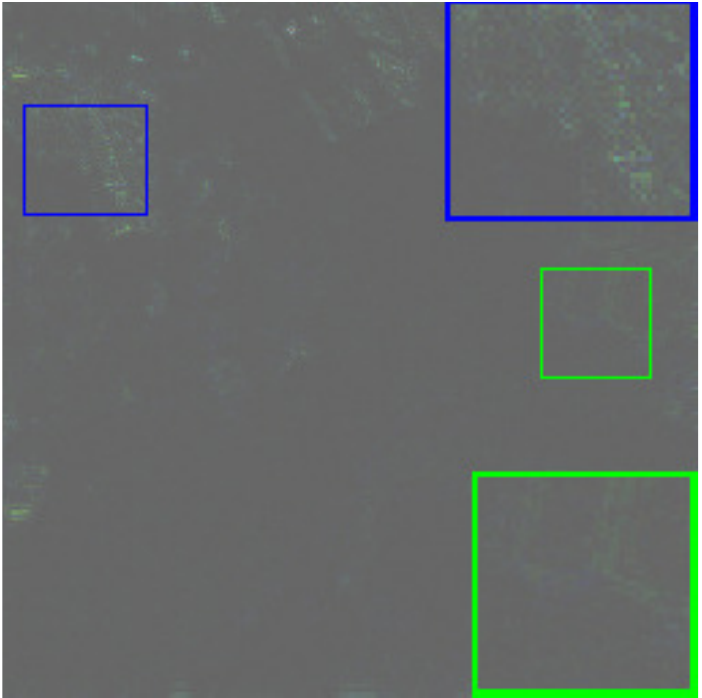}&
		\includegraphics[width=0.85in,height=0.85in]{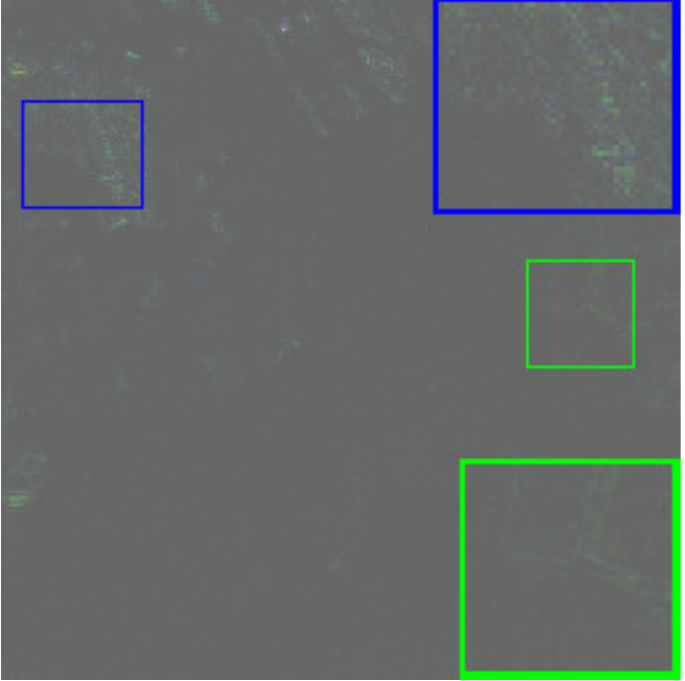}&
		\includegraphics[width=0.85in,height=0.85in]{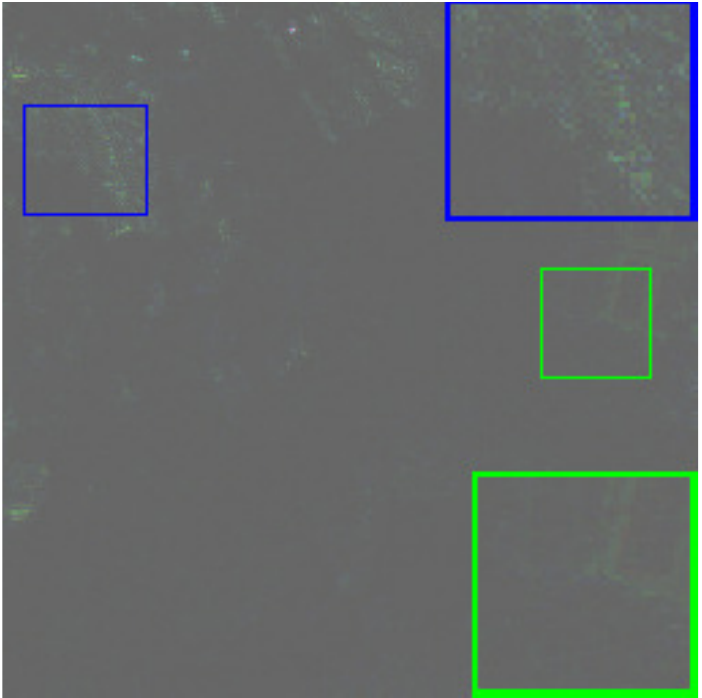}&
		\includegraphics[width=0.85in,height=0.85in]{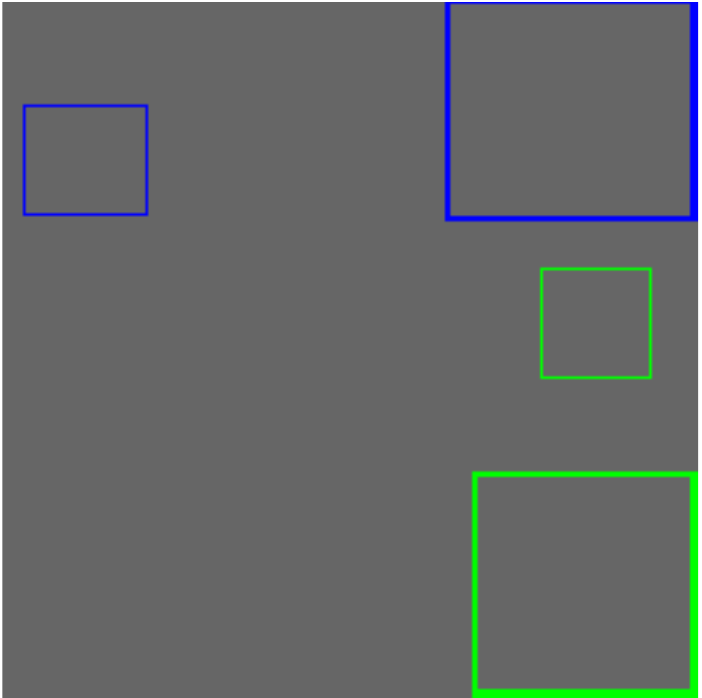}\\
		PNN&DiCNN1& PanNet&BDPN& DMDNet& FusionNet& TDNet& GT
	\end{tabular}
	\caption{The corresponding absolute error maps (AEMs) using the reference (GT) image on the reduced resolution Rio dataset (sensor: WorldView-3). For a better visualization, we doubled the intensities of the AEMs and added 0.3.}
	\label{fig:newdata8err}
\end{figure*}

\begin{table*}[hp]
	\footnotesize
	\setlength{\tabcolsep}{4.5pt}
	\renewcommand\arraystretch{1.2}
	\centering
	\caption{\footnotesize{Average metrics for all the compared DL-based approaches on 1258 reduced resolution samples. (Bold: best; Underline: second best)} \label{tab:1258}}
	\begin{tabular}{c|ccccc}
		\Xhline{1.5pt}
		\textbf{Method}&\emph{SAM} ($\pm$ std)&\emph{ERGAS} ($\pm$ std)& \emph{Q8} ($\pm$ std)& \emph{SCC} ($\pm$ std)\\
		\Xhline{1pt}
		\textbf{PNN~\cite{pnn}}  &4.4015 $\pm$ 1.3292  & 3.2283 $\pm$ 1.0042 & 0.8883 $\pm$ 0.1122  & 0.9215 $\pm$ 0.0464\\
		\textbf{DiCNN1~\cite{dicnn1}}  &3.9805  $\pm$ 1.3181  & 2.7367 $\pm$ 1.0156& 0.9096 $\pm$ 0.1117  & 0.9517 $\pm$ 0.0471\\
		\textbf{PanNet~\cite{2017PanNet}}  &4.0921  $\pm$ 1.2733 & 2.9524  $\pm$ 0.9778 & 0.8941  $\pm$ 0.1170  & 0.9494 $\pm$ 0.0460  \\
		\textbf{BDPN~\cite{2019BDPN}}  &3.9952  $\pm$ 1.3869 & 2.7234 $\pm$ 1.0394 & 0.9123 $\pm$ 0.1128 & 0.9515 $\pm$ 0.0457 \\
		\textbf{DMDNet~\cite{fu2020}}  &3.9714  $\pm$ 1.2482 & 2.8572 $\pm$ 0.9663   & 0.9000 $\pm$ 0.1141  & 0.9527 $\pm$ 0.0446   \\	
		\textbf{FusionNet~\cite{fusionnet}}  &\underline{3.7435 $\pm$ 1.2259} & \underline{2.5679 $\pm$ 0.9442} &\underline{0.9135 $\pm$ 0.1122} & \underline{0.9580 $\pm$ 0.0450}\\
		\textbf{TDNet}  &\bf{3.5036 $\pm$ 1.2411} & \bf{2.4439 $\pm$ 0.9587} &\bf{0.9212 $\pm$  0.1117} & \bf{0.9621 $\pm$ 0.0440}\\
		\Xhline{1pt}
		\textbf{Ideal value}&\bf{0}&\bf{0}&\bf{1}&\bf{1}\\ 
		\Xhline{1pt}
	\end{tabular}
\end{table*}

\begin{figure*}[tp]
	\scriptsize
	\setlength{\tabcolsep}{0.9pt}
	\centering
	\begin{tabular}{cccccccc}
		\includegraphics[width=0.85in,height=0.85in]{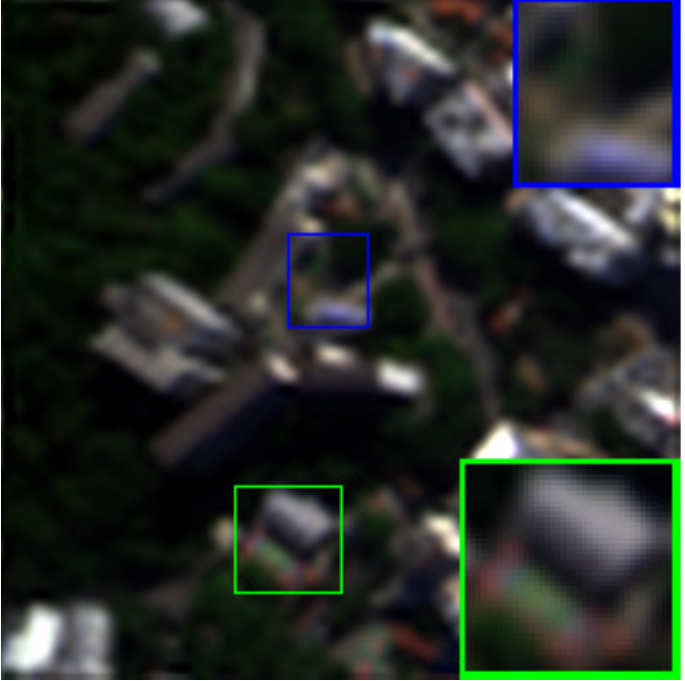}&
		\includegraphics[width=0.85in,height=0.85in]{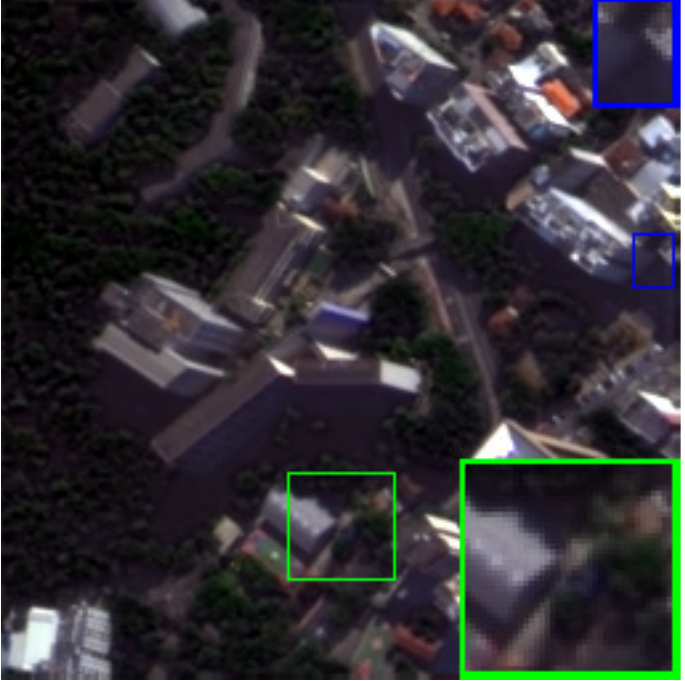}&
		\includegraphics[width=0.85in,height=0.85in]{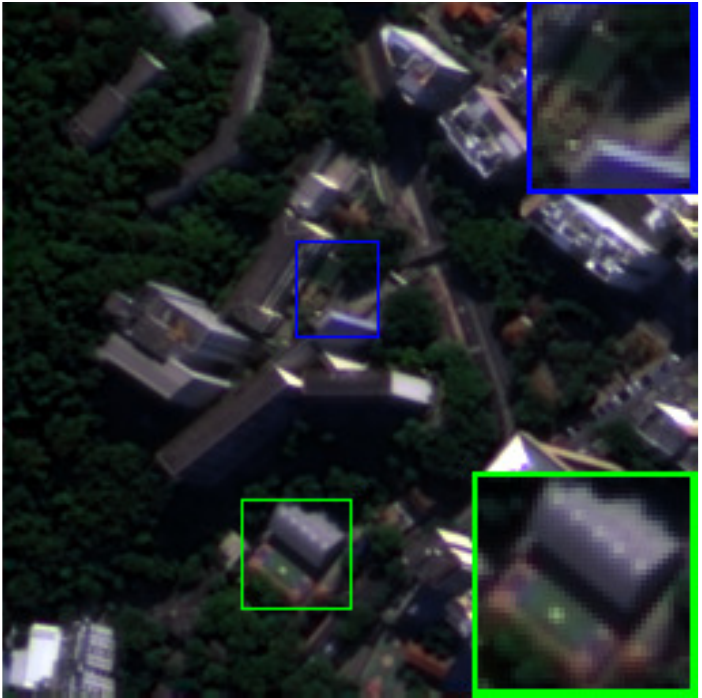}&
		
		\includegraphics[width=0.85in,height=0.85in]{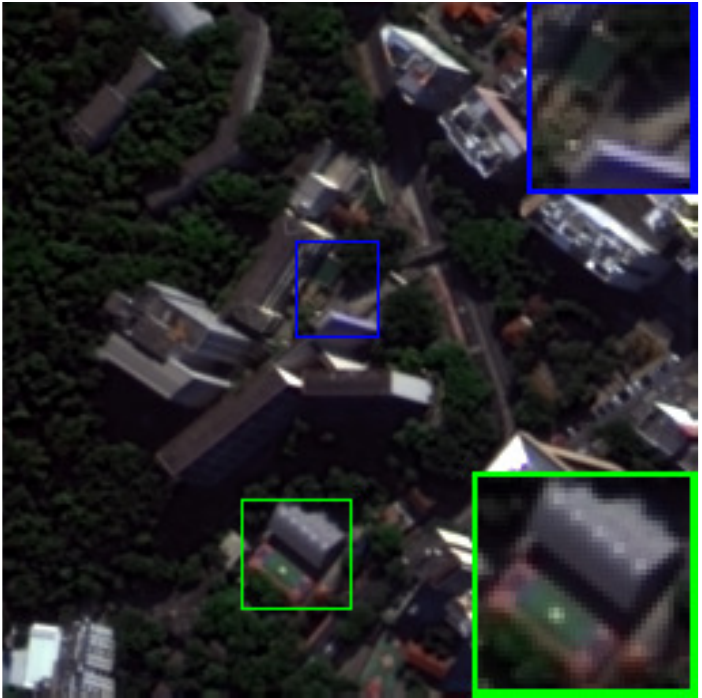}&
		\includegraphics[width=0.85in,height=0.85in]{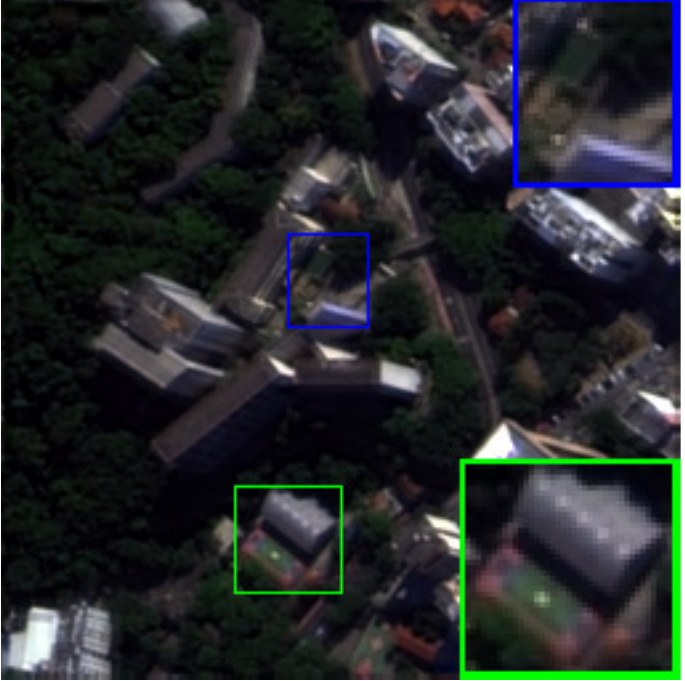}&
		\includegraphics[width=0.85in,height=0.85in]{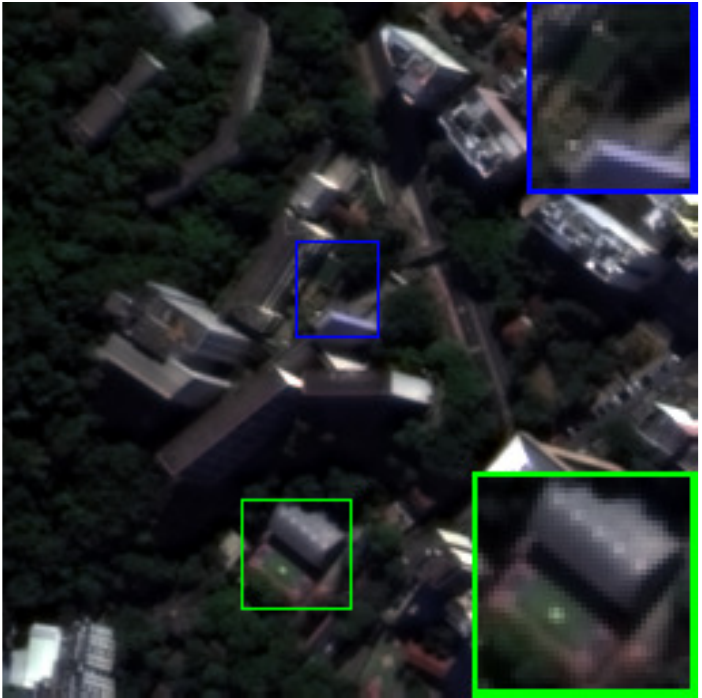}&
		\includegraphics[width=0.85in,height=0.85in]{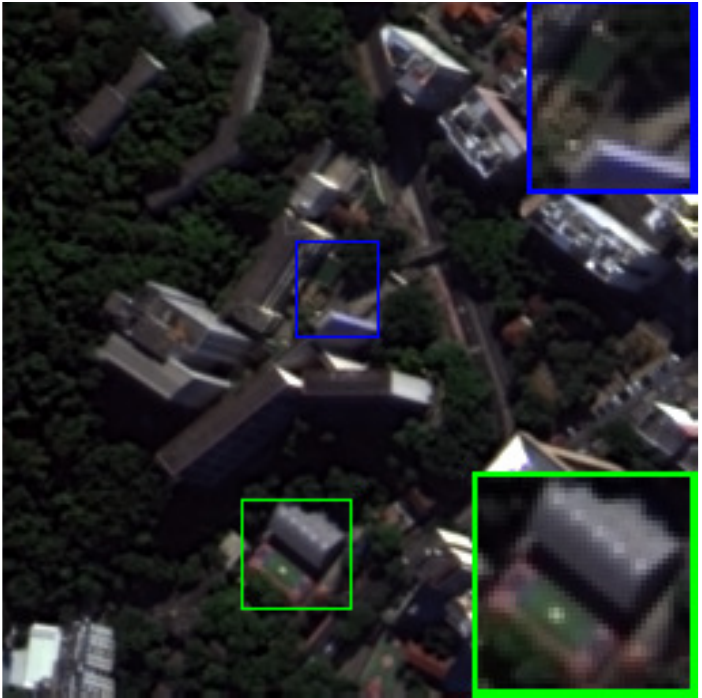}&
		\includegraphics[width=0.85in,height=0.85in]{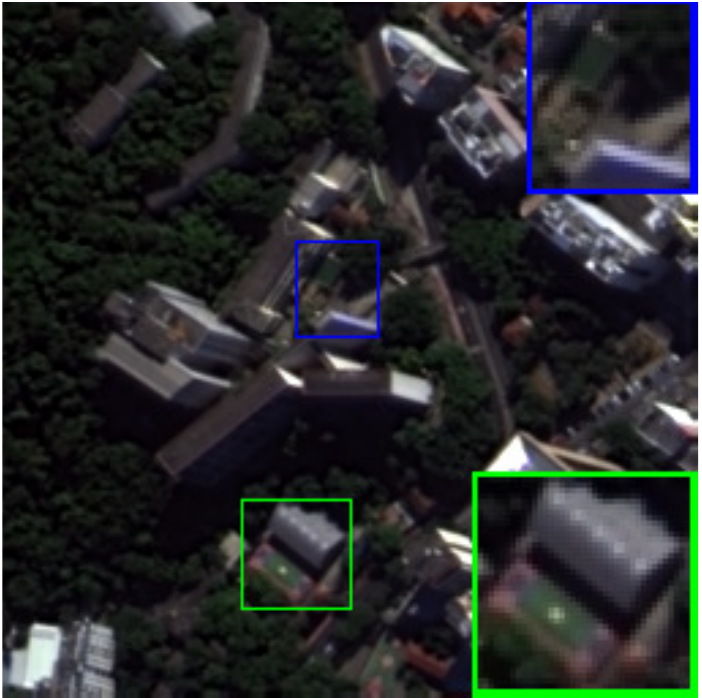}\\
		EXP& GS & PRACS& BDSD-PC& SFIM& GLP-HPM& GLP-CBD&GLP-Reg
	\end{tabular}
	\begin{tabular}{cccccccc}
		
		\includegraphics[width=0.85in,height=0.85in]{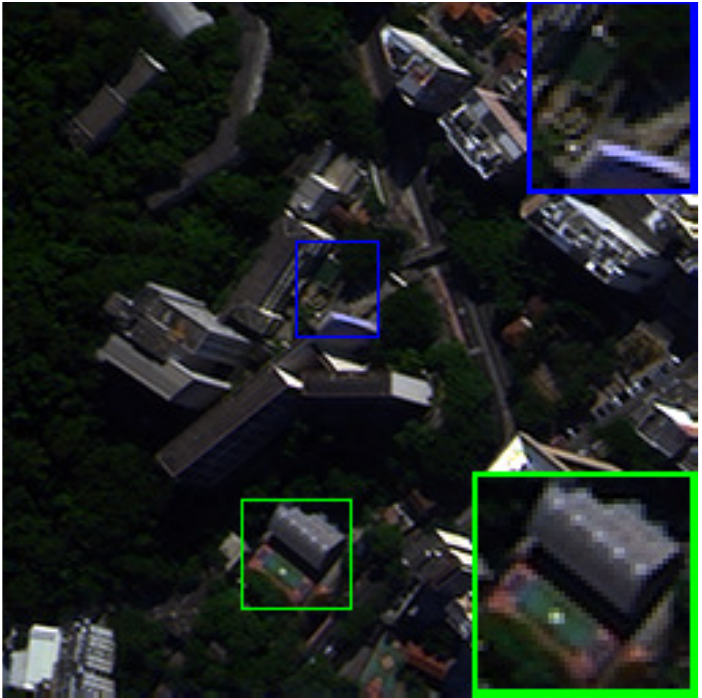}&
		\includegraphics[width=0.85in,height=0.85in]{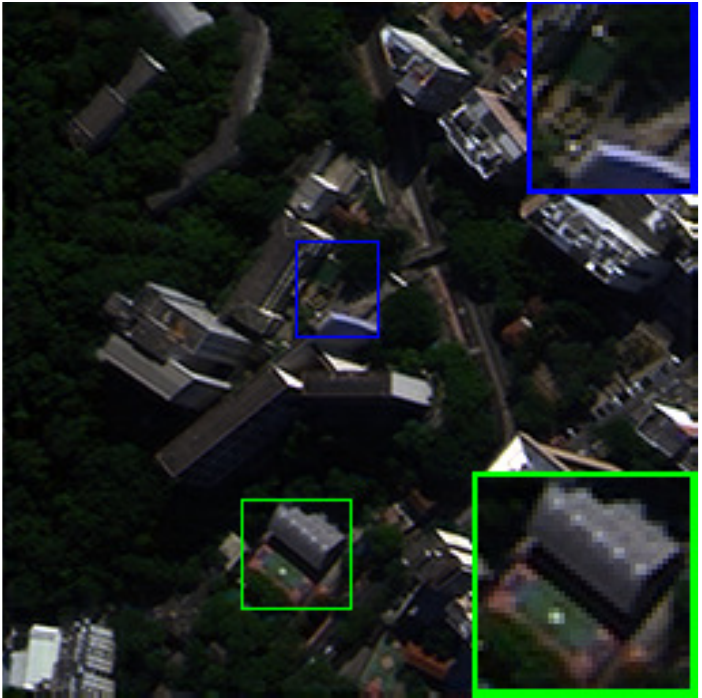}&
		\includegraphics[width=0.85in,height=0.85in]{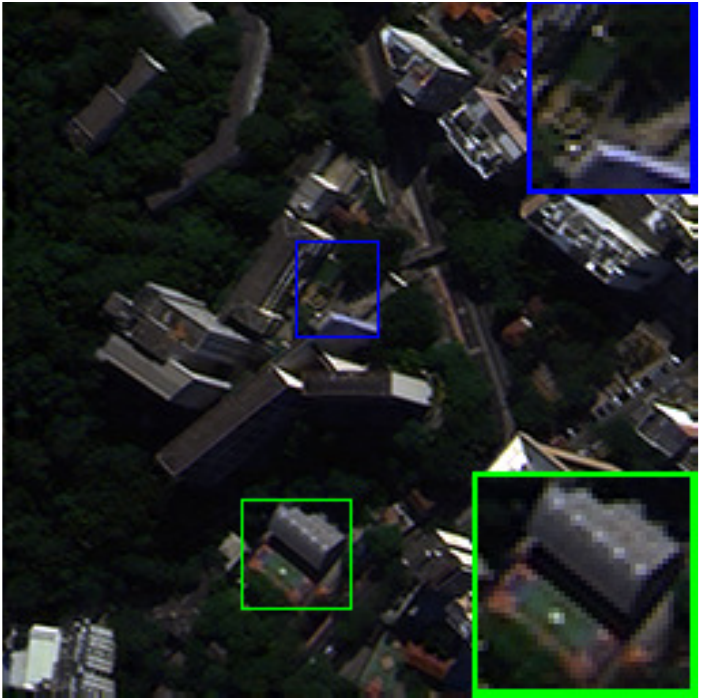}&
		\includegraphics[width=0.85in,height=0.85in]{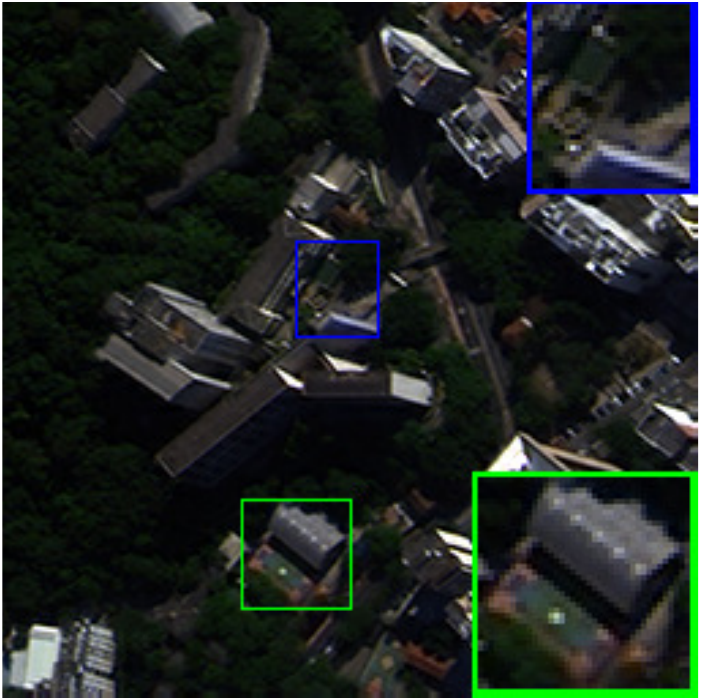}&
		\includegraphics[width=0.85in,height=0.85in]{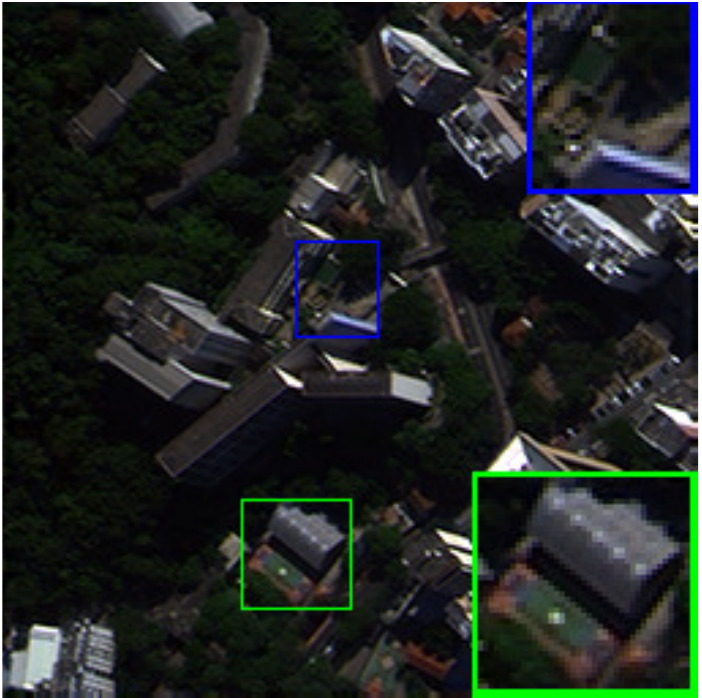}&
		\includegraphics[width=0.85in,height=0.85in]{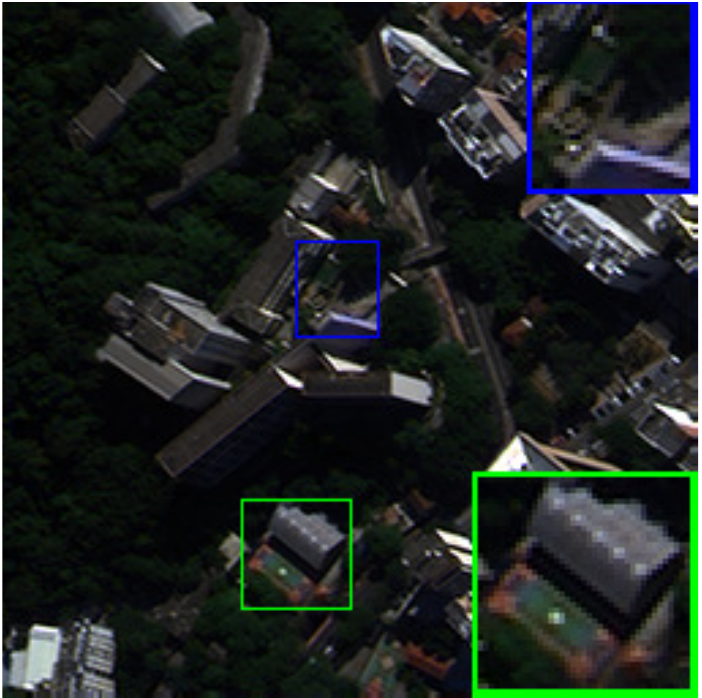}&
		\includegraphics[width=0.85in,height=0.85in]{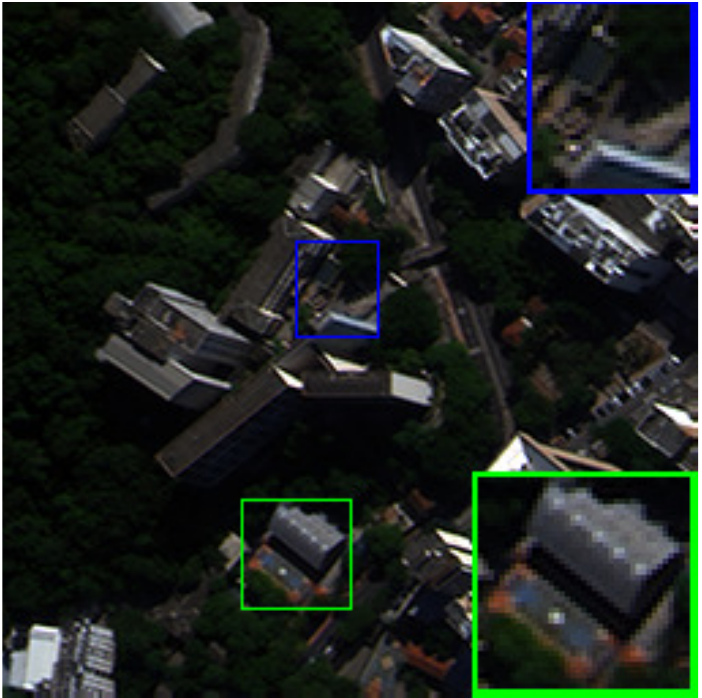}&
		\includegraphics[width=0.85in,height=0.85in]{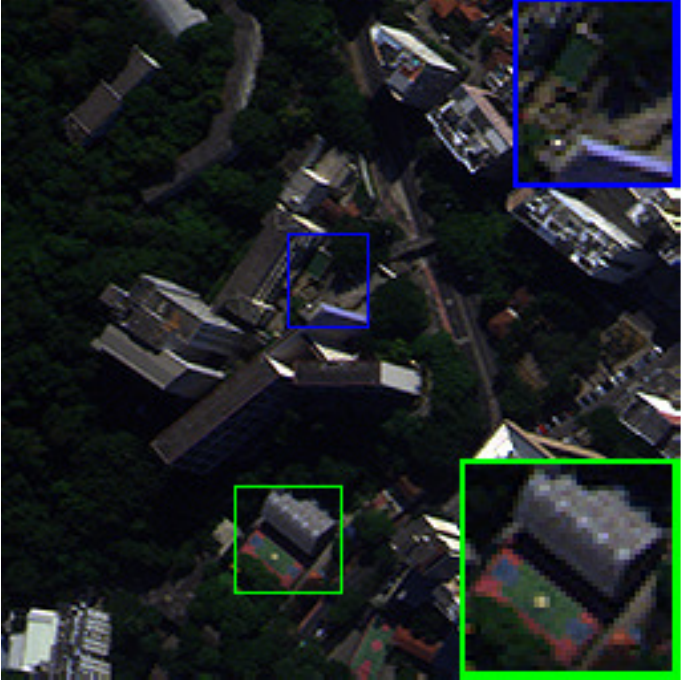}\\
		PNN&DiCNN1& PanNet&BDPN& DMDNet& FusionNet& TDNet& GT
	\end{tabular}
	\caption{Visual comparisons of all the compared approaches on the reduced resolution Tripoli dataset (sensor: WorldView-3).}
	\label{fig:newdata10}
\end{figure*}

\begin{figure*}[h]
	\scriptsize
	\setlength{\tabcolsep}{0.9pt}
	\centering
	\begin{tabular}{cccccccc}
		\includegraphics[width=0.85in,height=0.85in]{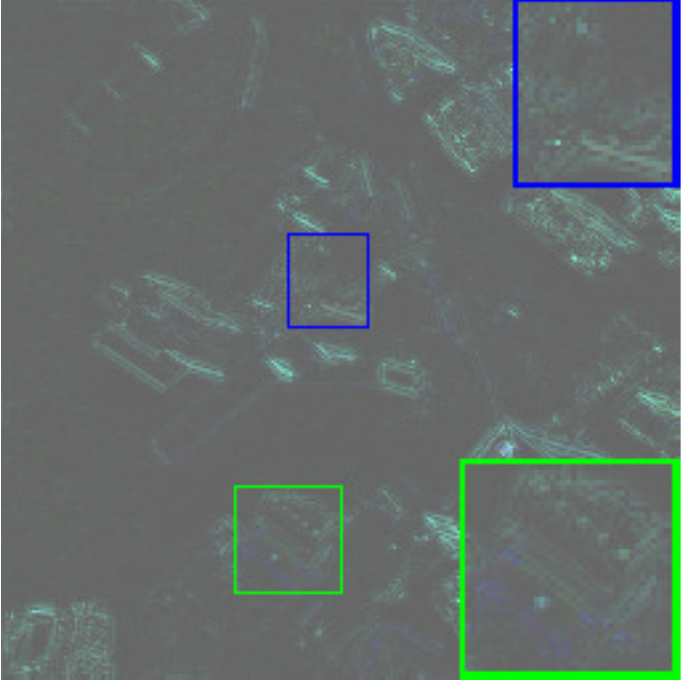}&
		\includegraphics[width=0.85in,height=0.85in]{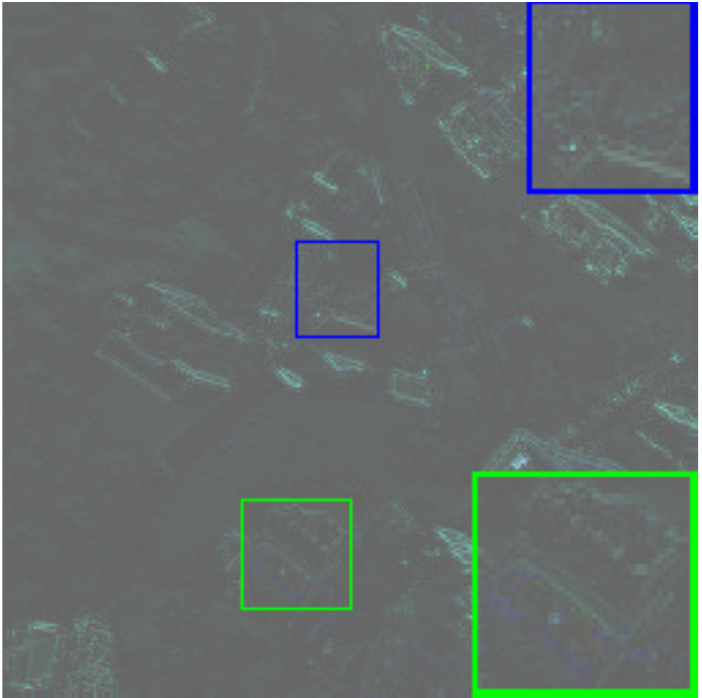}&
		\includegraphics[width=0.85in,height=0.85in]{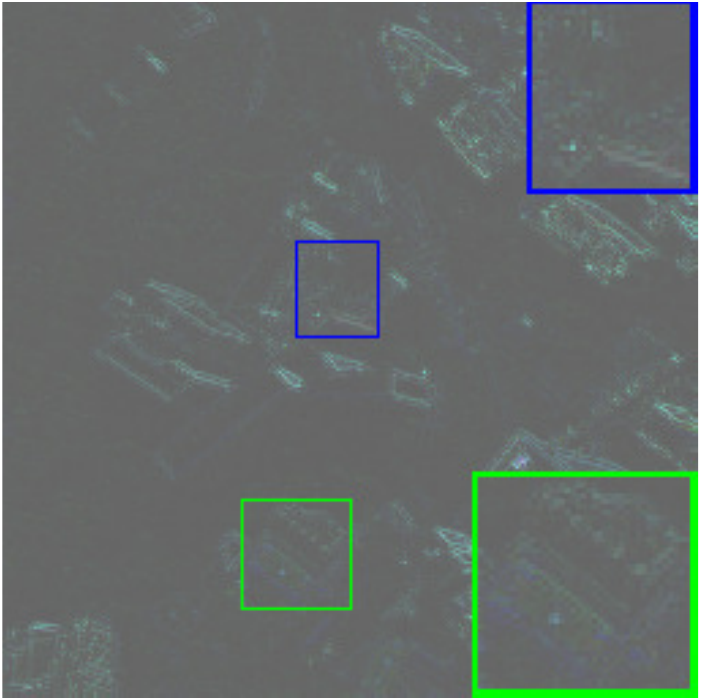}&
		\includegraphics[width=0.85in,height=0.85in]{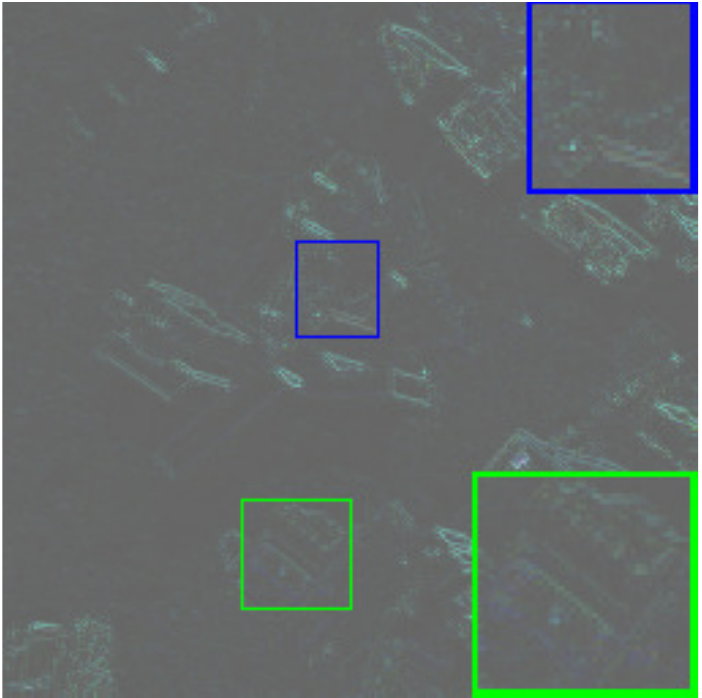}&
		\includegraphics[width=0.85in,height=0.85in]{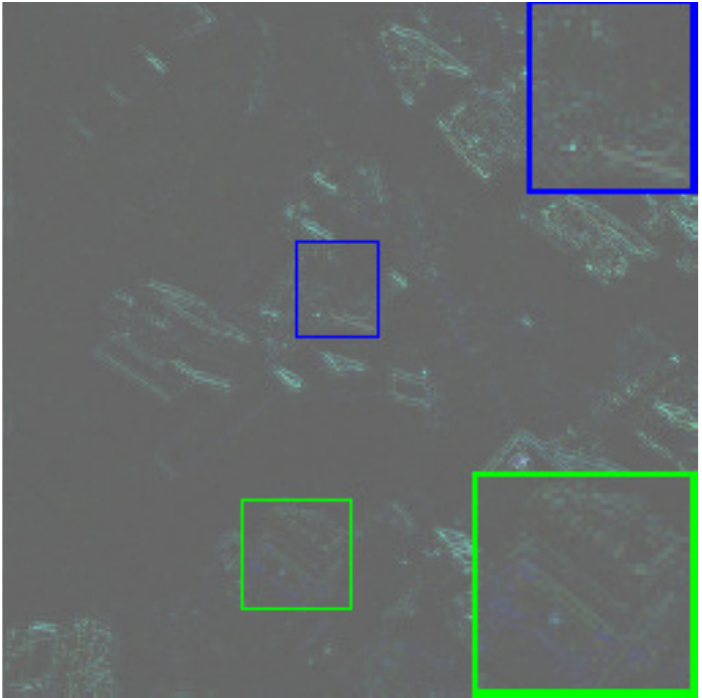}&
		\includegraphics[width=0.85in,height=0.85in]{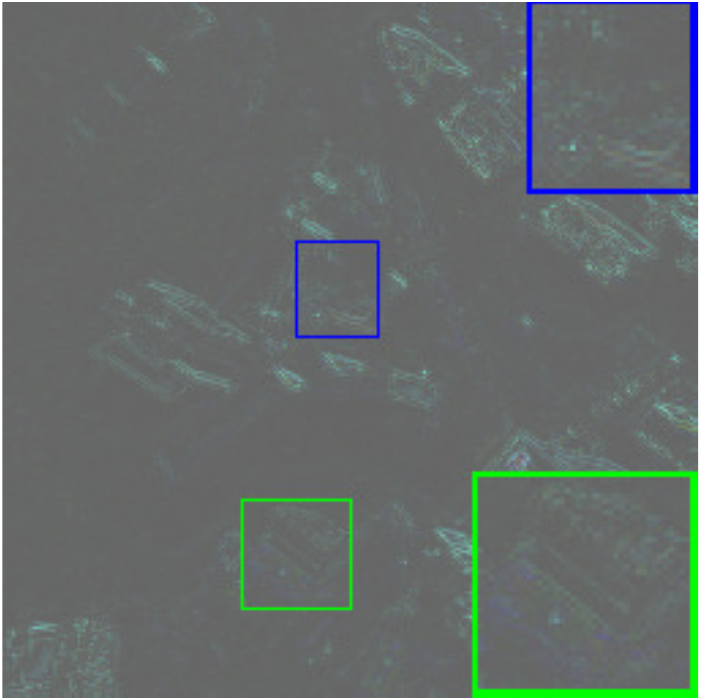}&
		\includegraphics[width=0.85in,height=0.85in]{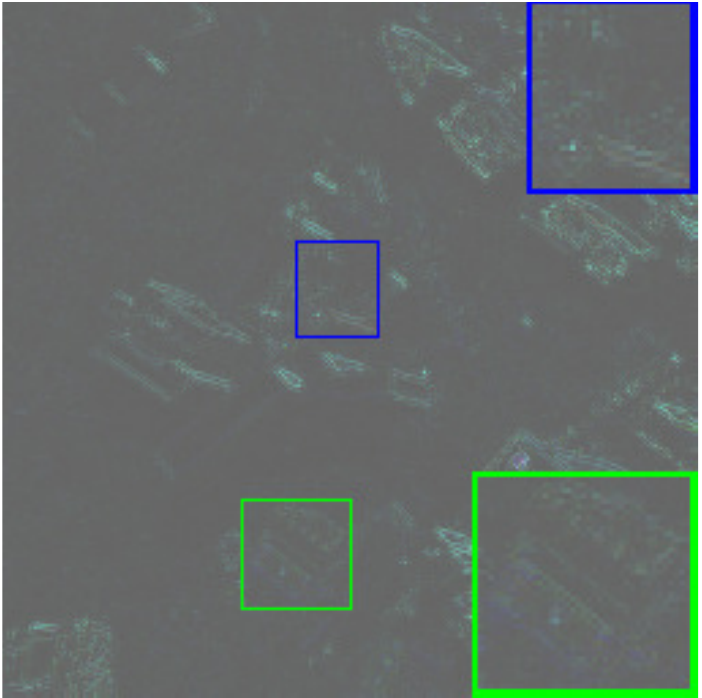}&
		\includegraphics[width=0.85in,height=0.85in]{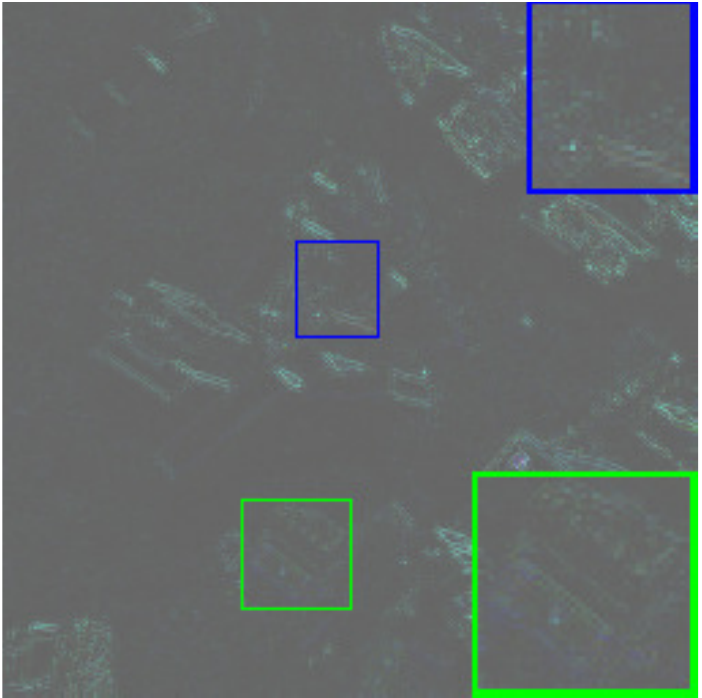}
		\\
		EXP& GS & PRACS& BDSD-PC& SFIM& GLP-HPM& GLP-CBD&GLP-Reg
	\end{tabular}
	\begin{tabular}{cccccccc}
		
		\includegraphics[width=0.85in,height=0.85in]{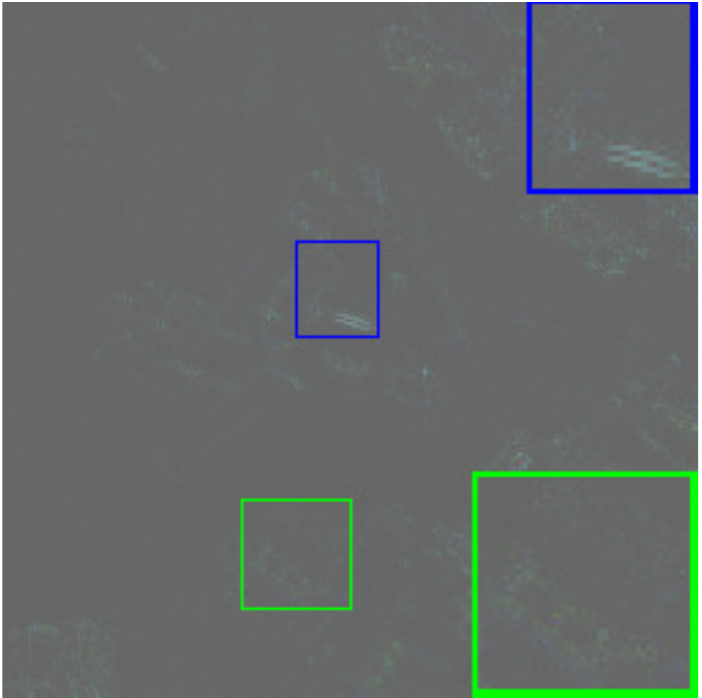}&
		\includegraphics[width=0.85in,height=0.85in]{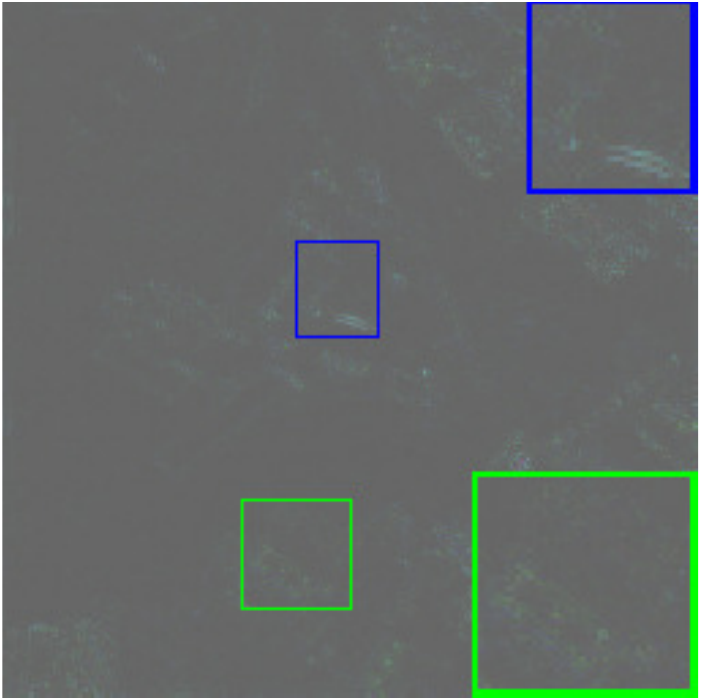}&
		\includegraphics[width=0.85in,height=0.85in]{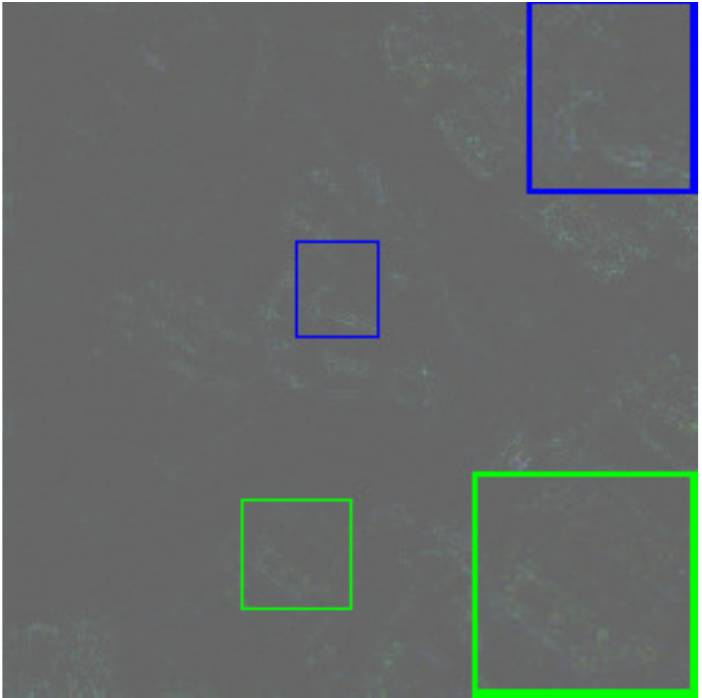}&
		\includegraphics[width=0.85in,height=0.85in]{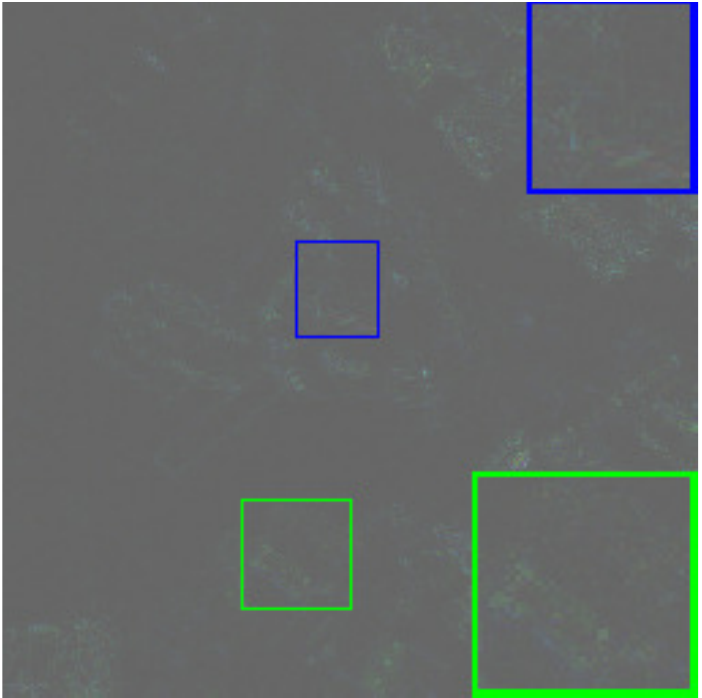}&
		\includegraphics[width=0.85in,height=0.85in]{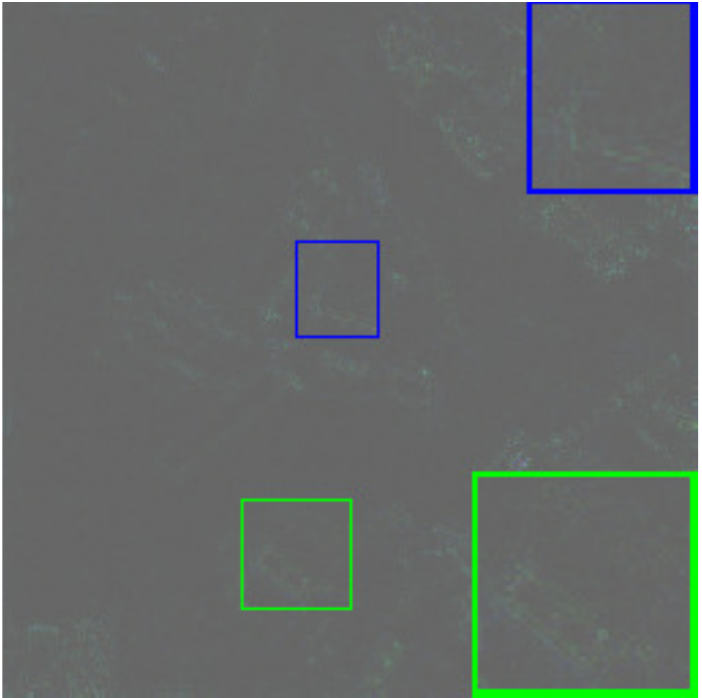}&
		\includegraphics[width=0.85in,height=0.85in]{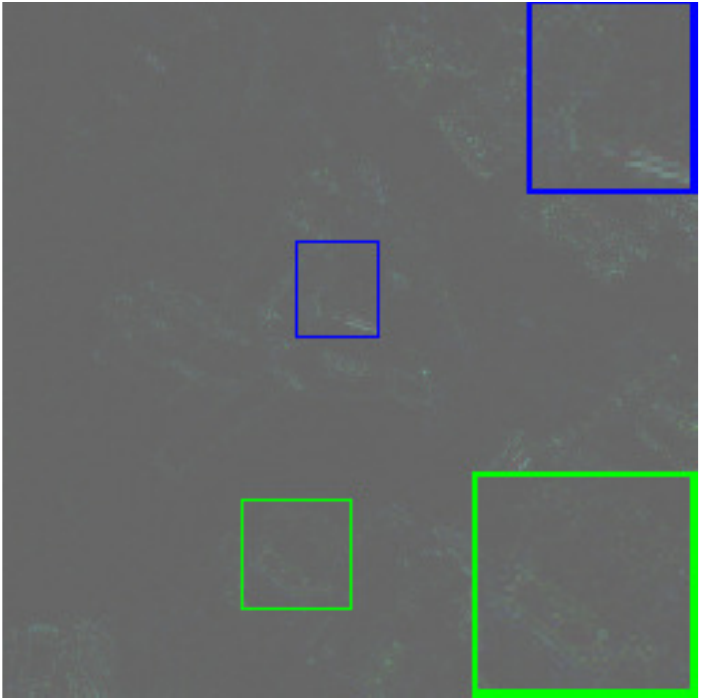}&
		\includegraphics[width=0.85in,height=0.85in]{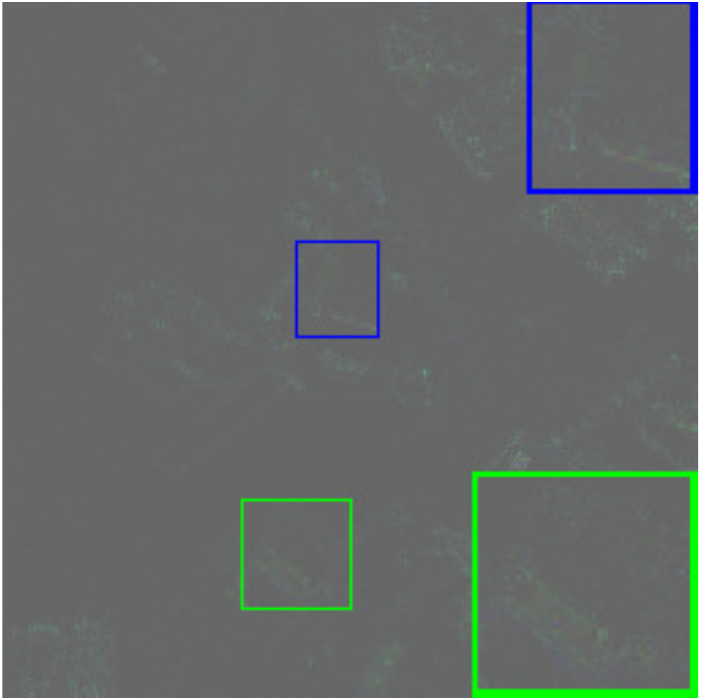}&
		\includegraphics[width=0.85in,height=0.85in]{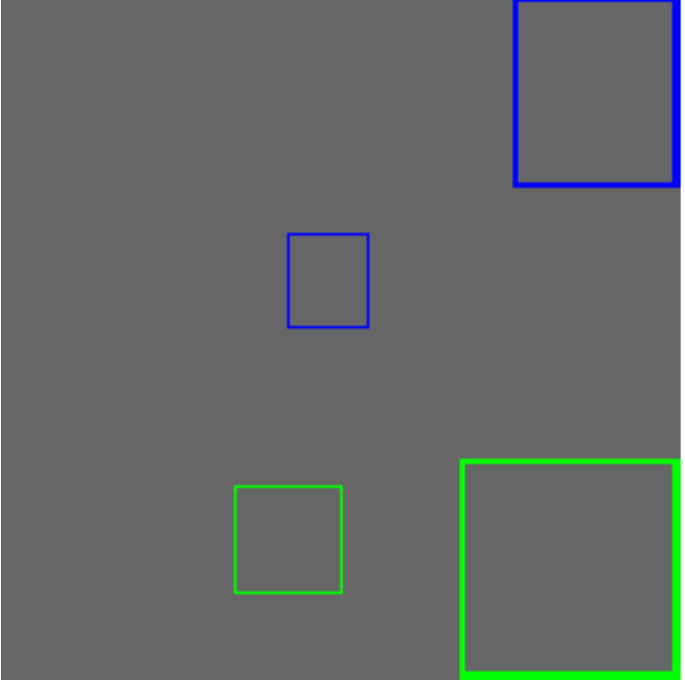}\\
		PNN&DiCNN1& PanNet&BDPN& DMDNet& FusionNet& TDNet& GT
	\end{tabular}
	\caption{The corresponding AEMs on the reduced resolution Tripoli dataset (sensor: WorldView-3). For a better visualization, we doubled the intensities of the AEMs and added 0.3.}
	\label{fig:newdata10err}
\end{figure*}
\begin{table*}[h]
	\footnotesize
	\setlength{\tabcolsep}{4.5pt}
	\renewcommand\arraystretch{1.2}
	\centering
	\caption{\footnotesize{Quality metrics for all the compared approaches on the reduced resolution Rio and Tripoli datasets, respectively. (Bold: best; Underline: second best)} 
		\label{tab:810}}
	\begin{tabular}{l|cccc|cccc}
		\Xhline{1pt}
		\multirow{2}{*}\textbf{Method}&\multicolumn{4}{c|}{\emph{(a) Rio dataset}}&\multicolumn{4}{c}{\emph{(b) Tripoli dataset}}\\\cline{2-9}
		&\emph{SAM}&\emph{ERGAS}&\emph{SCC}&\emph{Q8}&\emph{SAM}&\emph{ERGAS}&\emph{SCC}&\emph{Q8}\\\hline
		\textbf{EXP~\cite{glp}}&7.0033&6.5368&0.4797&0.5929&7.8530&9.0903& 0.5137&0.6608\\
		\textbf{GS~\cite{laben2000process}}&8.9481&6.3662&0.6775&0.8666&8.3772&7.8041&0.7240&0.7291\\
		\textbf{PRACS~\cite{pracs}}&7.1176 &5.5392&0.6370&0.7380&7.9705&7.4826& 0.7111&0.7951\\
		\textbf{BDSD-PC~\cite{bdsdpc2019}}&7.0721&4.9515&0.7164&0.7853&7.7347&7.0345&0.7304&0.8202\\
		\textbf{SFIM~\cite{sfim1}}&6.8501 &4.9571&0.7558&0.7882&7.4338&7.0695& 0.7514&0.8020\\
		\textbf{GLP-HPM~\cite{atwt}}&7.2994&5.1185&0.7369&0.7849&7.9390&7.1489&0.7415&0.8238\\
		\textbf{GLP-CBD~\cite{rbim}}&7.4053&5.0372&0.6738&0.7880&7.8051&6.9162& 0.7312& 0.8300\\
		\textbf{GLP-Reg~\cite{GlpReg2018}}&7.3275 &5.0154&0.6822&0.7889&7.7680&6.9100& 0.7327&0.8298\\
		\textbf{PNN~\cite{pnn}}&4.0659&2.7144& 0.9487& 0.8888&5.6714&3.5657&0.9435&0.9166\\
		\textbf{DiCNN~\cite{dicnn1}}&3.8289&2.5819&0.9544&0.8895&5.3622&3.3104&0.9523&0.9348\\
		\textbf{PanNet~\cite{2017PanNet}}&3.9062&2.6583&0.9522&0.8814&4.8500&3.1744& 0.9642&0.9190\\
		\textbf{BDPN~\cite{2019BDPN}}  &4.0788  &2.6897 & 0.8969 &0.9439&5.5728&3.3880&0.9520&0.9379 \\
		\textbf{DMDNet~\cite{fu2020}}&3.6917&2.4968&0.9594&0.8990&4.5649&2.9795&0.9706& 0.9243\\
		\textbf{FusionNet~\cite{fusionnet}}&\underline{3.5700}&\underline{2.4346}&\underline{0.9607}&\underline{0.9044}&\underline{4.3850}&\underline{2.8533}&\underline{0.9718}&\underline{0.9422}\\
		\textbf{TDNet}&\textbf{3.3801}&\textbf{2.3522}&\textbf{0.9648}&\textbf{0.9155}&\textbf{4.1445}&\textbf{2.7076}&\textbf{0.9761}&\bf{0.9526}\\
		\hline
		\textbf{Ideal value}&\bf{0}&\bf{0}&\bf{1}&\bf{1}&\bf{0}&\bf{0}&\bf{1}&\bf{1}\\ 
		\Xhline{1pt}
	\end{tabular}
\end{table*}

\subsection{Reduced Resolution Assessment}\label{ra}
The reduced resolution assessment measures the similarity between the fused image and the ideal reference image (the original MS image). The similarity can be determined by the calculation of several evaluation indexes. For the reduced resolution experiments, the spectral angle mapper (SAM) \cite{sam1992}, the dimensionless global error in synthesis (ERGAS) \cite{ergas2002}, the spatial correlation coefficient (SCC) \cite{SCC}, and the Q$2^n$ (Q8 for 8-band datasets and Q4 for 4-band datasets)~\cite{q2n} are employed as evaluation indexes. The ideal values for SAM and ERGAS are 0, whereas 1 for Q$2^n$ and SCC.

As mentioned in Sect.~\ref{Sec-Train}, we have 1258 testing samples from WorldView-3 images. We first compare the proposed TDNet with the five state-of-the-art CNN-based pansharpening approaches on the 1258 samples. From Tab.~\ref{tab:1258}, it is clear that the TDNet obtains the best average quantitative performance on all the metrics demonstrating the superiority of the proposed method. This can be justified because, comparing our approach with conventional CNNs for pansharpening, it utilizes multi-scale convolution kernels for a better feature extraction. Besides, comparing it with PanNet and DMDNet, which directly send the high-frequency information of PAN and MS images into the network, our TDNet adopts two branches for better exploiting the multi-scale structures of PAN and MS images. Moreover, our MRA-inspired TDNet could hold better physical meanings with respect to DiCNN. Furthermore, due to the triple-double network structure, \emph{i.e.,} double-level, double-branch, and double-direction, the TDNet can fully utilize the latent multi-scale information of PAN and MS images, thus obtaining better results than FusionNet.
\begin{table*}[h]
	\caption{\small{Average values of QNR, $\text{D}_{\lambda}$ and $\text{D}_{s}$ with the related standard deviations (std) for 50 full resolution WorldView-3 samples. (Bold: best; Underline: second best)} \label{tab:os}}
	\footnotesize
	\renewcommand\arraystretch{1.2}
	\centering
	\begin{center}
		\begin{tabular}{l|ccc}
			\hline
			\textbf{Method}&\emph{QNR} ($\pm$ std)&\emph{$\text{D}_{\lambda}$} ($\pm$ std)&\emph{$\text{D}_{s}$} ($\pm$ std)\\
			\hline
			\textbf{EXP~\cite{glp}} &0.8078  $\pm$ 0.0673  & 0.0582 $\pm$ 0.0274&  0.1010  $\pm$ 0.0428 \\
			\textbf{GS~\cite{laben2000process}} &0.8806  $\pm$ 0.0351  & 0.0343 $\pm$ 0.0239&  0.0882  $\pm$ 0.0238\\
			\textbf{PRACS~\cite{pracs}} &0.9204 $\pm$ 0.0172 & 0.0335 $\pm$ 0.0055 & 0.0668   $\pm$ 0.0159\\
			\textbf{BDSD-PC~\cite{bdsdpc2019}} &0.9063 $\pm$ 0.0231  & 0.0322 $\pm$ 0.0108 & 0.0731 $\pm$  0.0175\\
			\textbf{SFIM~\cite{sfim1}} &0.8999 $\pm$ 0.0423 & 0.0371 $\pm$ 0.0207 & 0.0657 $\pm$ 0.0244 \\ 
			\textbf{GLP-HPM~\cite{atwt}} & 0.8384 $\pm$ 0.0757 & 0.0332 $\pm$ 0.0162  &  0.0647 $\pm$ 0.0234 \\
			\textbf{GLP-CBD~\cite{rbim}}   &0.8795 $\pm$ 0.0510  & 0.0418 $\pm$ 0.0210   &0.0827 $\pm$ 0.0337 \\	
			\textbf{GLP-Reg~\cite{GlpReg2018}}   & 0.8812 $\pm$ 0.0498  & 0.0408 $\pm$ 0.0205   &0.0818 $\pm$ 0.0328 \\
			\textbf{PNN~\cite{pnn}}  &0.9446 $\pm$  0.0233 &  0.0255 $\pm$ 0.0138  & 0.0306 $\pm$ 0.0117 \\
			\textbf{DiCNN1~\cite{dicnn1}}  &\underline{0.9564  $\pm$ 0.0124 }&\underline{0.0231  $\pm$ 0.0113}   & \bf{0.0208 $\pm$ 0.0072} \\ 
			\textbf{PanNet~\cite{2017PanNet}}  &0.9421 $\pm$ 0.0227  & 0.0345 $\pm$ 0.0146 & 0.0242 $\pm$ 0.0107 \\
			\textbf{BDPN~\cite{2019BDPN}}  &0.9206 $\pm$ 0.0399   &0.0365 $\pm$ 0.0252   & 0.0350 $\pm$ 0.0089\\ 
			\textbf{DMDNet~\cite{fu2020}}  &0.9383  $\pm$ 0.0329 &  0.0309  $\pm$ 0.0162 & 0.0320 $\pm$ 0.0192 \\
			\textbf{FusionNet~\cite{fusionnet}}  &0.9435 $\pm$ 0.0259  & 0.0303 $\pm$ 0.0096  & 0.0255 $\pm$ 0.0076  \\
			\textbf{TDNet}  &\bf{0.9575 $\pm$ 0.0051}  & \bf{0.0209 $\pm$ 0.0079}   & \underline{0.0219 $\pm$ 0.0052}  \\
			\Xhline{1pt}
			\textbf{Ideal value}&\bf{1}&\bf{0}&\bf{0}\\
			\Xhline{1pt}
		\end{tabular}
	\end{center}
\end{table*}
\begin{figure*}[h]
	\scriptsize
	\setlength{\tabcolsep}{0.9pt}
	\centering
	\begin{tabular}{cccccccc}
		\includegraphics[width=0.85in,height=0.85in]{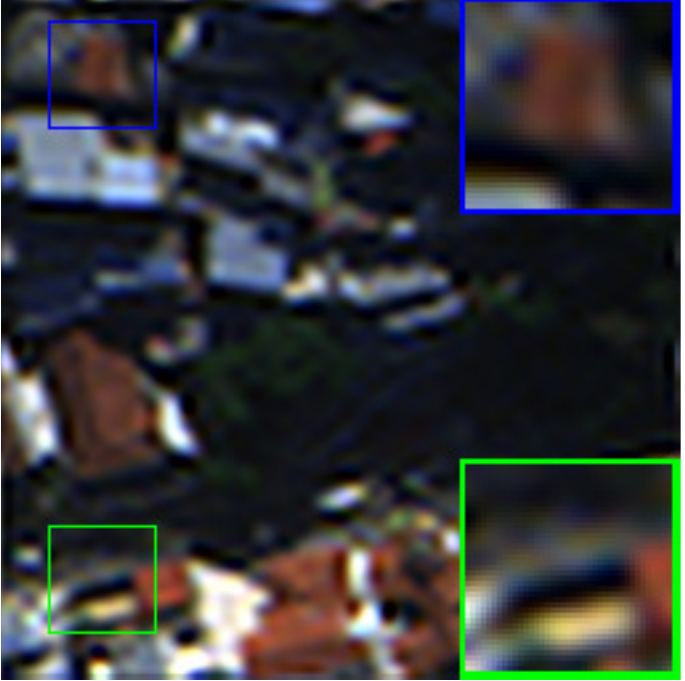}&
		\includegraphics[width=0.85in,height=0.85in]{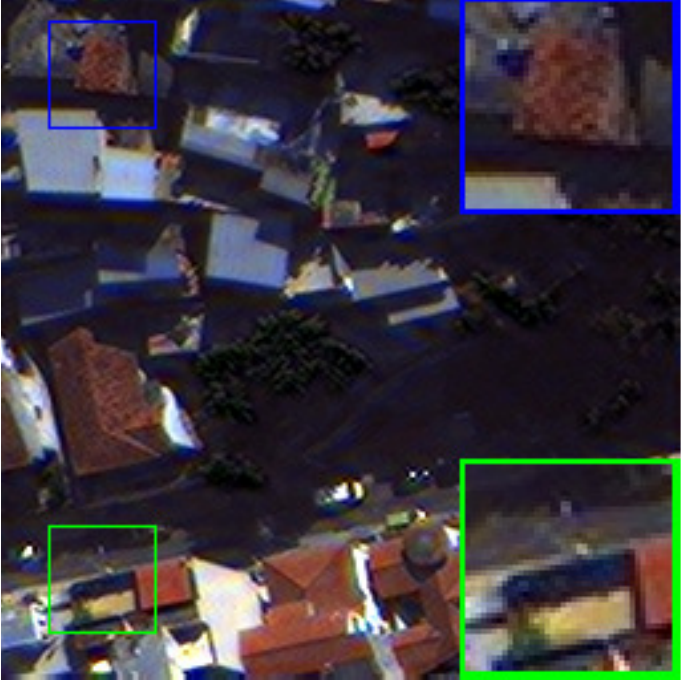}&
		\includegraphics[width=0.85in,height=0.85in]{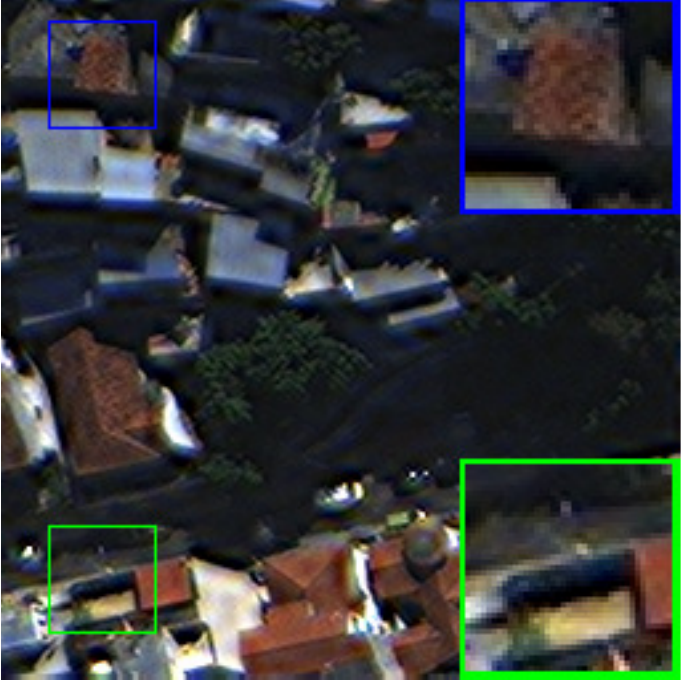}&
		\includegraphics[width=0.85in,height=0.85in]{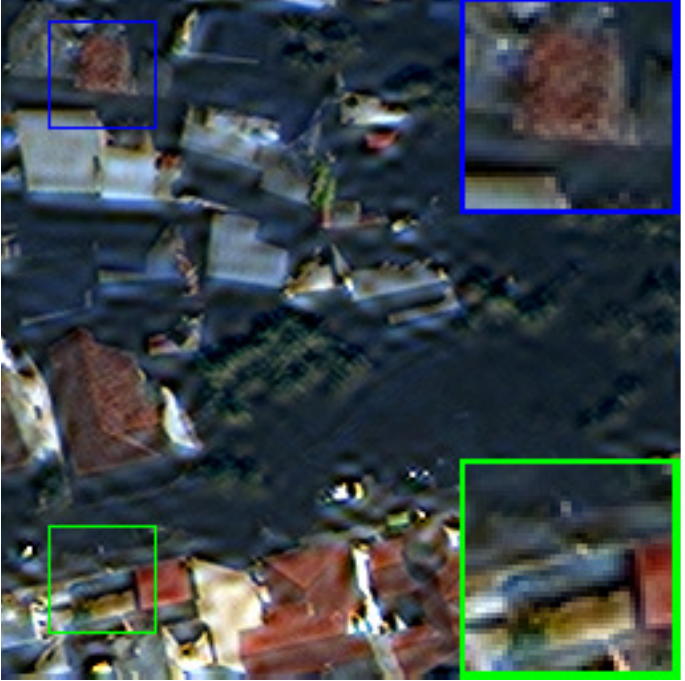}&
		\includegraphics[width=0.85in,height=0.85in]{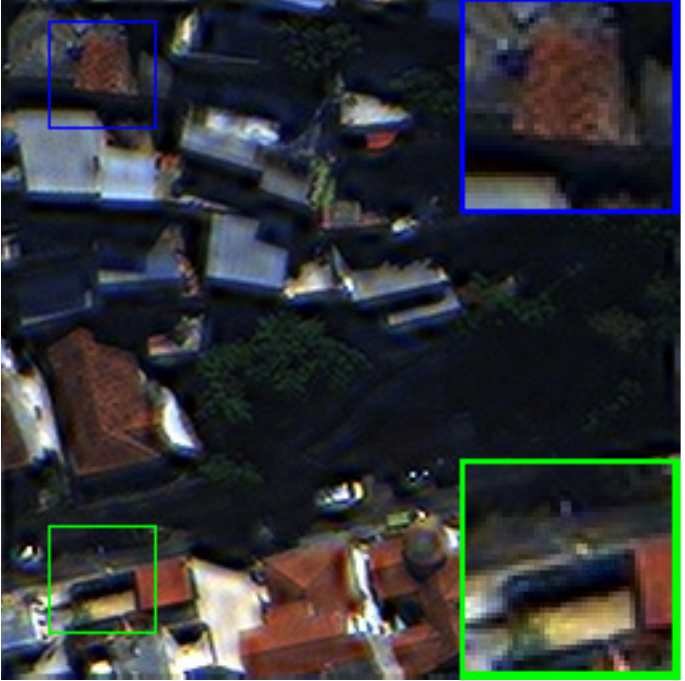}&
		\includegraphics[width=0.85in,height=0.85in]{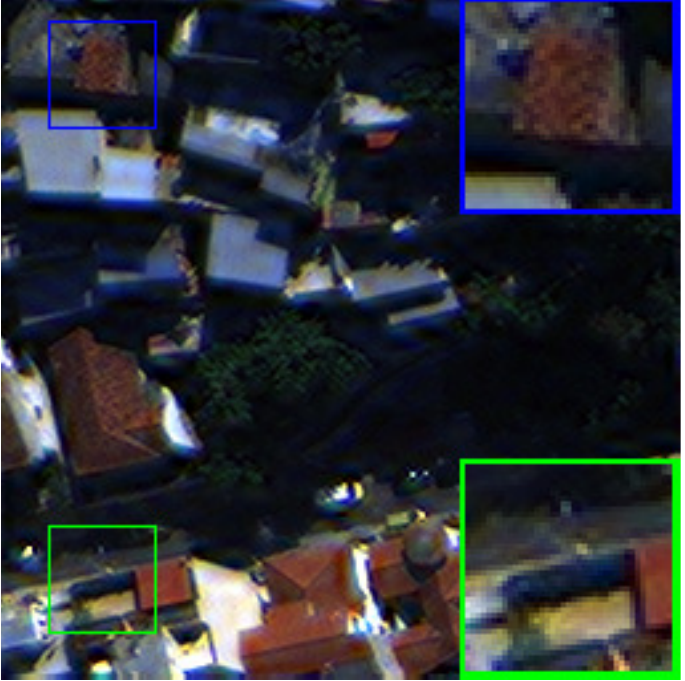}&
		\includegraphics[width=0.85in,height=0.85in]{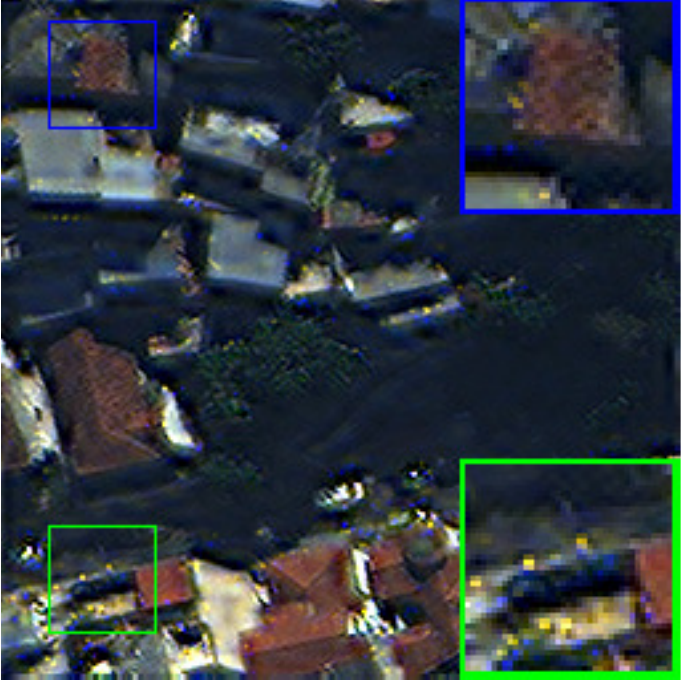}&
		\includegraphics[width=0.85in,height=0.85in]{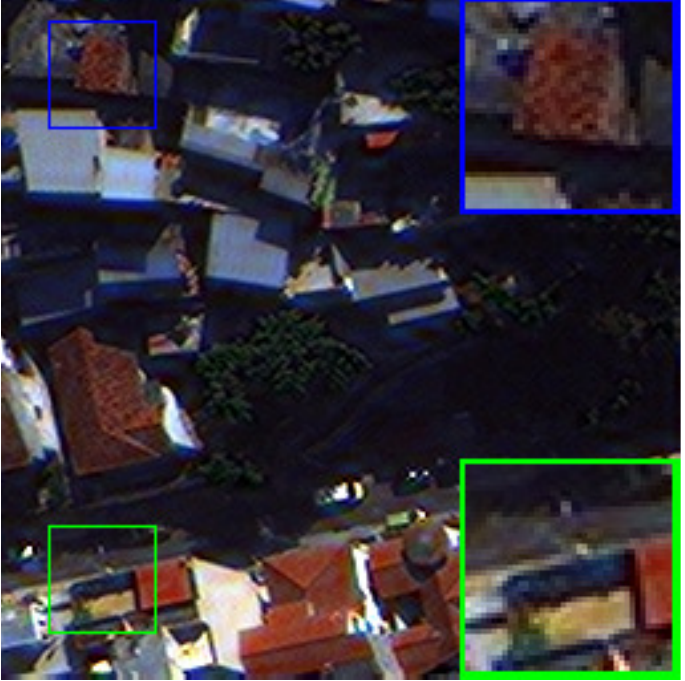}\\
		EXP&GS & PRACS& BDSD-PC& SFIM& GLP-HPM& GLP-CBD&GLP-Reg
	\end{tabular}
	\begin{tabular}{cccccccc}
		
		\includegraphics[width=0.85in,height=0.85in]{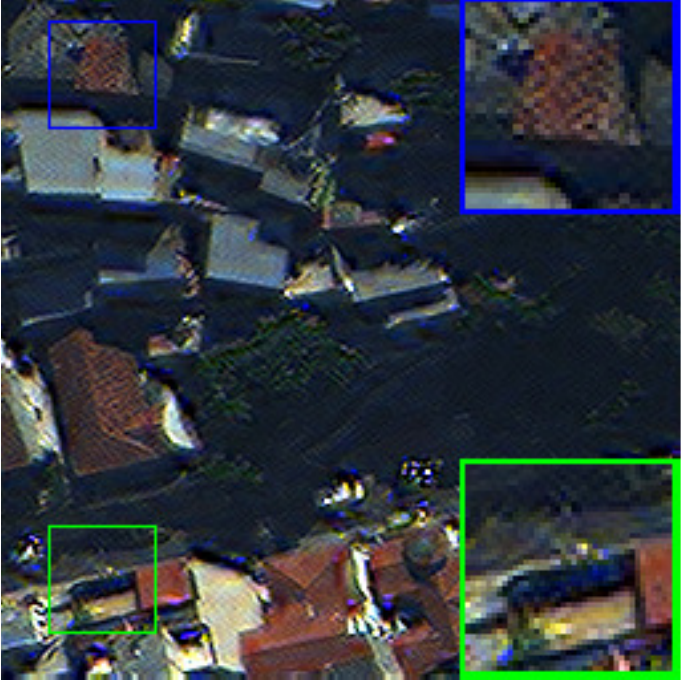}&
		\includegraphics[width=0.85in,height=0.85in]{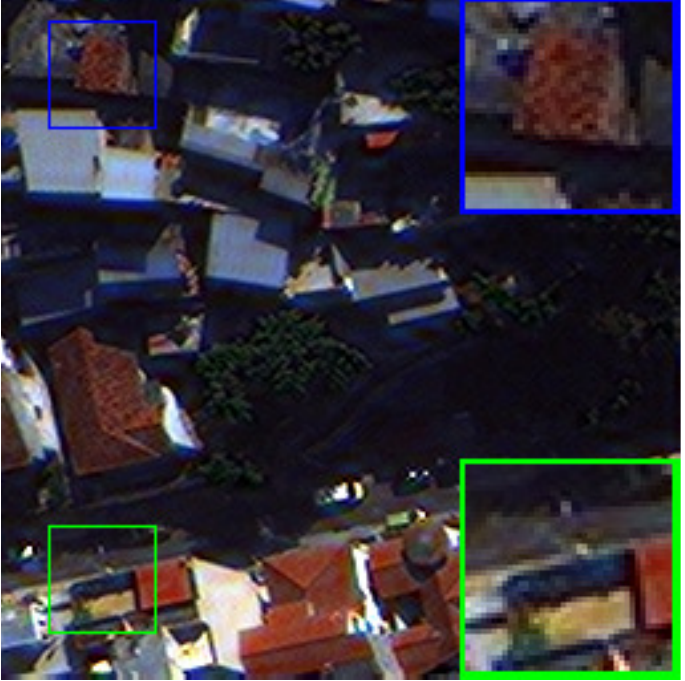}&
		\includegraphics[width=0.85in,height=0.85in]{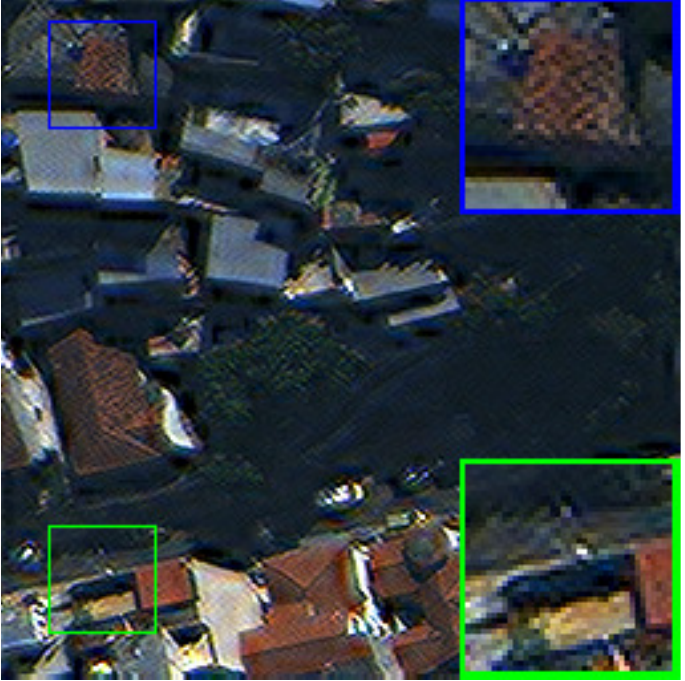}&
		\includegraphics[width=0.85in,height=0.85in]{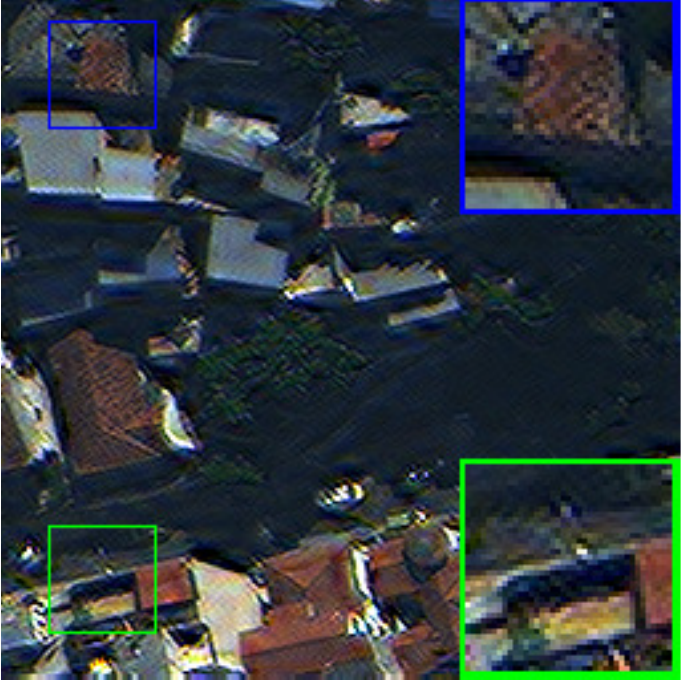}&
		\includegraphics[width=0.85in,height=0.85in]{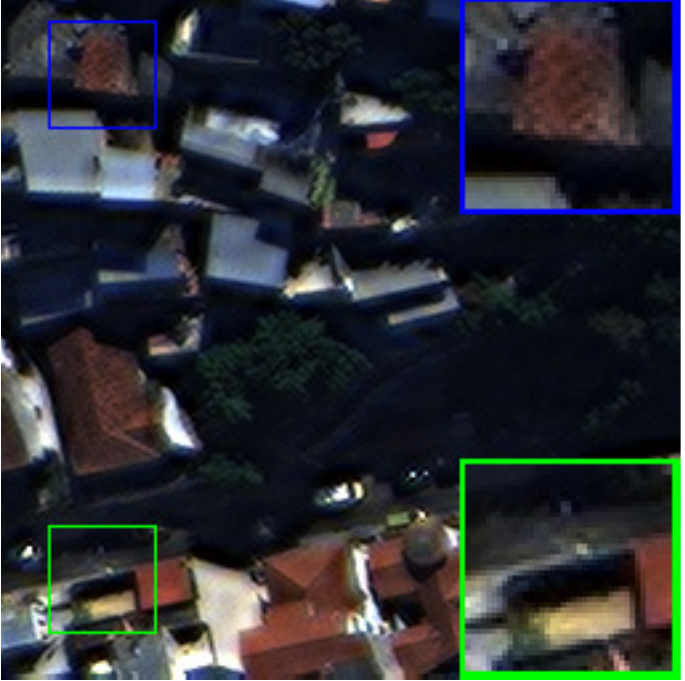}&
		\includegraphics[width=0.85in,height=0.85in]{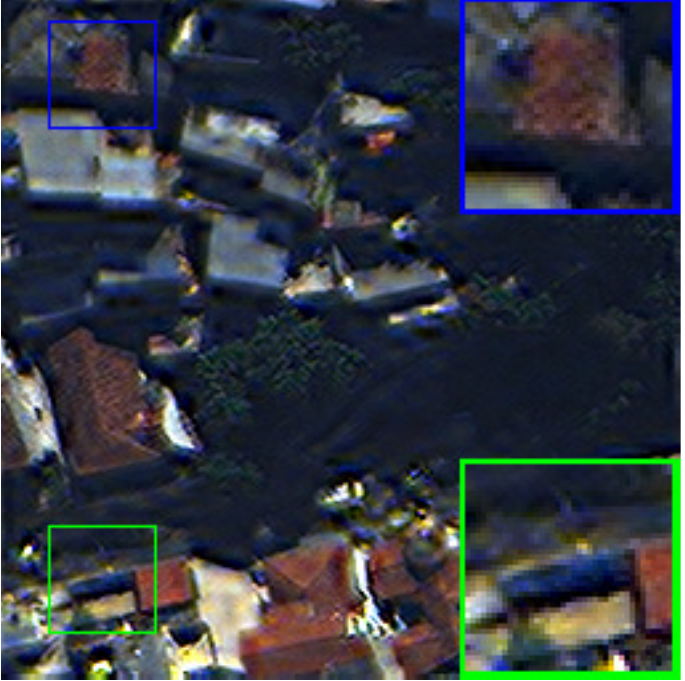}&
		\includegraphics[width=0.85in,height=0.85in]{wv3_OriginalScale737_bdsd_pc-eps-converted-to.pdf}&
		\includegraphics[width=0.85in,height=0.85in]{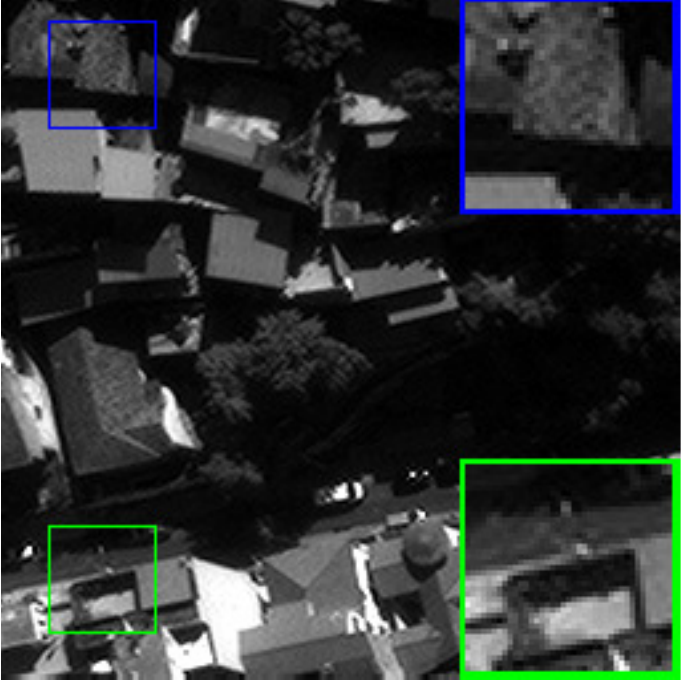}\\
		PNN&DiCNN1&PanNet& BDPN& DMDNet& FusionNet& TDNet& PAN
	\end{tabular}
	\caption{Visual comparisons between the TDNet and the benchmark on the full resolution Rio dataset (sensor: WorldView-3).}
	\label{fig:os}
\end{figure*}

Furthermore, we generated two WorldView-3 test cases (Rio dataset and Tripoli dataset) at reduced resolution by applying Wald's protocol (please, refer to Sect.~\ref{Sec-Train} for details about Wald's protocol implementation). The GT image has a size $256\times 256\times 8$, as well as the LRMS and PAN images have a size $64\times 64\times 8$ and $256\times 256$, respectively. Figs. \ref{fig:newdata8}-\ref{fig:newdata10err} show the visual comparisons among all the 15 compared pansharpening approaches. From these figures, it is easy to note that the TDNet yields results very close to the GT image. The traditional CS and MRA techniques generate products with some obvious spatial blur, especially at near the boundaries of the buildings and/or spectral distortion. The other DL-based approaches perform better than the traditional methods. However, TDNet shows less image residuals compared with the other CNN-based methods, see Fig.~\ref{fig:newdata8err} and Fig.~\ref{fig:newdata10err}. In Tab.~\ref{tab:810}, the quantitative metrics demonstrate that TDNet still gets the best performance, assessing the superiority of our TDNet approach, which is able to reduce both spatial and spectral distortions in the fusion outcome.


\subsection{Full Resolution Assessment}
To corroborate the results at reduced resolution, a full resolution analysis is also needed involving the original MS and PAN products. Unlike the reduced resolution test cases, we have no reference (GT) image. Thus, three widely used metrics, that do not exploit the GT image, are employed, \emph{i.e.,} the quality with no reference (QNR) index, the spectral distortion $\text{D}_{\lambda}$ index, and the spatial distortion $\text{D}_{s}$ index~\cite{vivone2015critical}.

Fig.~\ref{fig:os} presents the visual comparison of all the compared pansharpening approaches on a full resolution example. In this case, the spectral quality of the image should refer to the LRMS image, and the spatial details of a high-quality fusion image should be close to the PAN image. It can be observed that result by the TDNet is the most qualified one, with sharper and clearer edges and without ghosting, blurring, \textit{etc.}
Furthermore, Tab.~\ref{tab:os} reports the average performance on 50 full resolution examples. Again, TDNet obtains the best average results and the minimum standard deviations demonstrating the superiority and stability of our method on full resolution test cases.

\begin{table*}[h]
	\footnotesize
	\setlength{\tabcolsep}{4.5pt}
	\renewcommand\arraystretch{1.2}
	\centering
	\caption{\footnotesize{ Quality metrics for different network structures on the reduced resolution 1258 datasets. (Bold: best; Underline: second best)} 
		\label{tab:ablation}}
	\begin{tabular}{c|ccccc}
		\hline
		\textbf{Method}&\emph{SAM} ($\pm$ std)&\emph{ERGAS} ($\pm$ std)& \emph{Q8} ($\pm$ std)& \emph{SCC} ($\pm$ std)\\
		\hline
		\textbf{w/o MRAB}  &3.7731 $\pm$ 1.2394  & 2.5608 $\pm$ 0.9443 & 0.9139 $\pm$ 0.1156  & 0.9587 $\pm$ 0.0445\\
		\textbf{ SSCB}  &3.5739  $\pm$ 1.3035  & \underline{2.4501 $\pm$ 0.9812} & 0.9199 $\pm$ 0.1241  & 0.9601 $\pm$ 0.0517 \\
		\textbf{w/o PAN branch}  &3.8701 $\pm$ 1.3509 & 2.7572 $\pm$ 0.9910 &0.9099 $\pm$  0.1127 & 0.9577 $\pm$  0.0443\\
		\textbf{Single-stage}  &3.9706 $\pm$ 1.2493 & 2.8432 $\pm$ 0.9677 &0.9098 $\pm$  0.1134 & 0.9523 $\pm$ 0.0430\\
		\textbf{TDNet(bilinear)}  &{\underline{3.5197 $\pm$ 1.2567}} & {2.5207 $\pm$ 0.9908} &{0.9198 $\pm$  0.1232} & {0.9607 $\pm$ 0.0510}\\
		\textbf{{TDNet(Deconv)}}  &{3.5276 $\pm$ 1.2721} &{ 2.5103 $\pm$ 0.9601} &{\underline{0.9207 $\pm$  0.1219}} & {\underline0.9610 $\pm$ 0.0456}\\
		\textbf{ {TDNet(-)}}  & {3.6987 $\pm$ 1.3107} &  {2.5479 $\pm$ 0.9511} & { 0.9187 $\pm$  0.1201} &  {0.9607 $\pm$ 0.0457}\\
		\textbf{ TDNet-TMRA}  & 3.9942 $\pm$ 1.8703 & 2.7812 $\pm$ 1.0123 &0.8997 $\pm$  0.1257 &  0.9465 $\pm$ 0.0529\\
		\Xhline{1pt}
		\textbf{TDNet}  &\textbf{3.5036 $\pm$ 1.2411} & \textbf{2.4439 $\pm$ 0.9587} & \textbf{0.9212 $\pm$  0.1117} & \textbf{0.9621 $\pm$ 0.0440}\\
		\Xhline{1pt}
	\end{tabular}
\end{table*}

\subsection{The Ablation Study}\label{sec:ablation}
\begin{figure}[!t]
	\begin{center}
		\begin{minipage}{ 0.98\linewidth}
			\begin{minipage}{ 0.48\linewidth}
				{\includegraphics[width=1\linewidth]{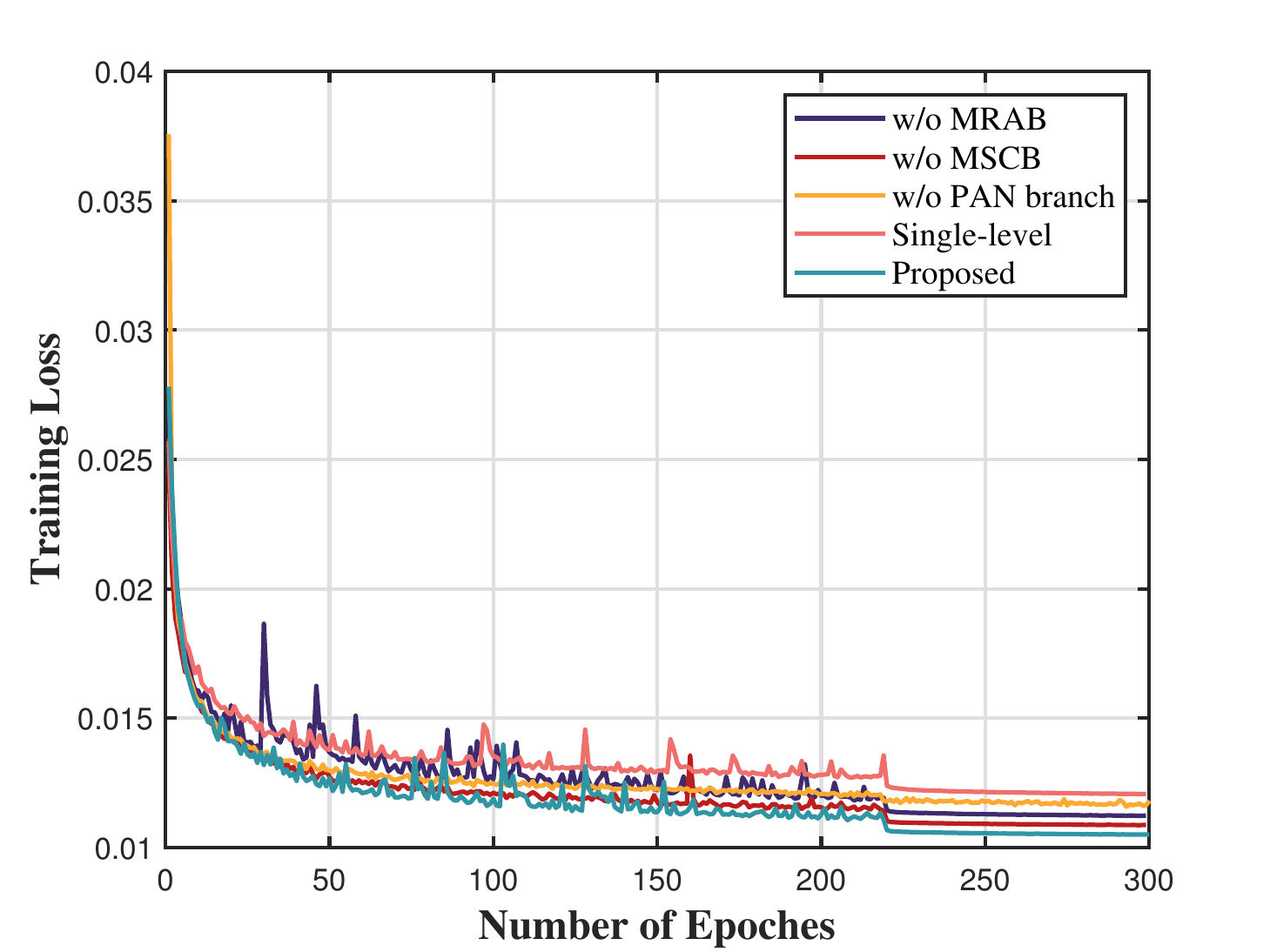}}
				\centering
				{(a) }
			\end{minipage}
			\begin{minipage}{ 0.48\linewidth}
				{\includegraphics[width=1\linewidth]{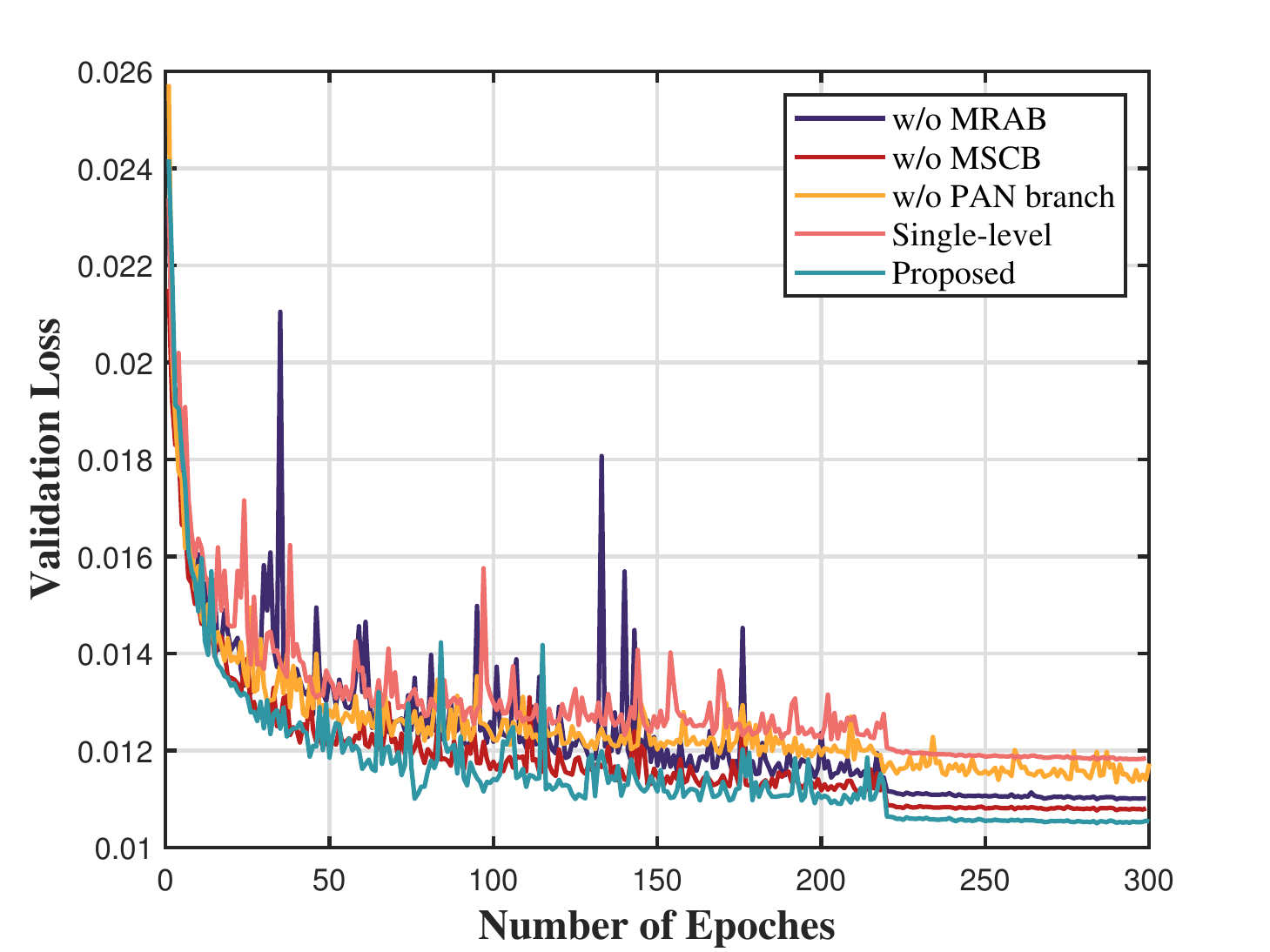}}
				\centering
				{(b) }
			\end{minipage}
			\centering
		\end{minipage}
	\end{center}
	\caption{Convergence curves for different network structures. (a) Training loss curves. (b) Validation loss curves. Please note that the single-level loss is calculated by (\ref{eq:loss2}), while the loss of the other structures is calculated by (\ref{eq:loss}) (with $\gamma$ = 0.4). Please note that the learning rate for 0-220 epochs is 0.001, and for 220-300 epochs is adjusted to 0.0001.}\label{loss}
\end{figure}

This section is devoted to ablation studies to investigate the effect of each component of the TDNet. For simplicity, we take a WorldView-3 dataset as reference. An overall performance calculated on the training/validation loss function can be found in Fig. \ref{loss} for various network structures. It is easy to show that the proposed TDNet shows the smallest loss compared with the other test cases.

\subsubsection{The Effect of MRAB}
To explore whether the MRAB contributes to the final result, we remove the MRAB from TDNet running the training with the same data and parameters. Tab.~\ref{tab:ablation} presents the average outcomes and the corresponding standard deviations for the TDNet with and without MRAB (w/o MRAB). It can be observed that the fusion results w/o MRAB have inferior performance on all the metrics compared with the original TDNet. This indicates that the MRAB can help the network in learning more details and features. 

\begin{figure}[h]
	\begin{center}
		{\includegraphics[width=0.6\linewidth]{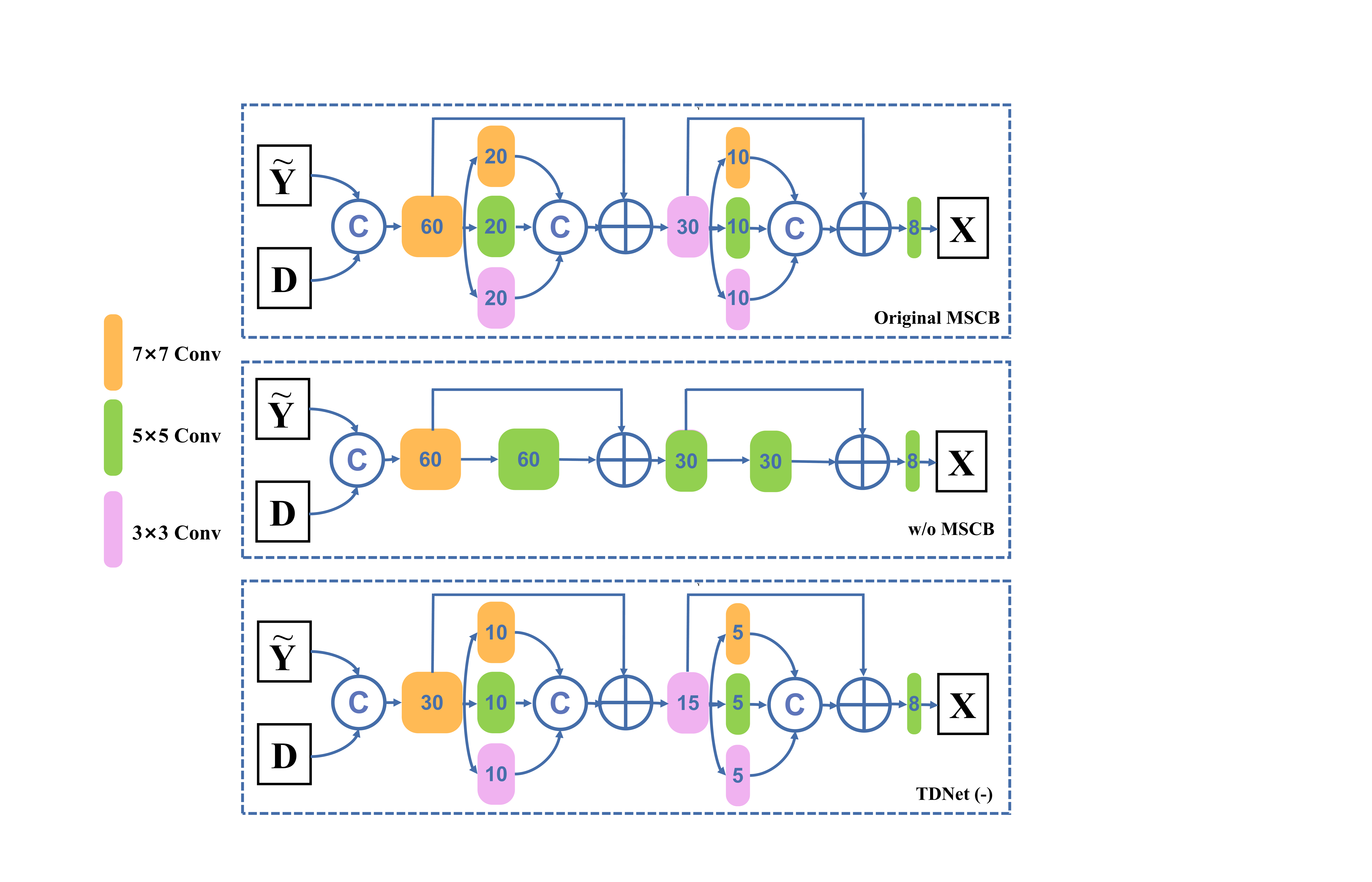}}
	\end{center}
	\caption{TDNet with MSCB (top row) and without MSCB (bottom row, also called SSCB).  Note that the numbers in color boxes mean the number of convolution kernels.}\label{fig:abla_mscb}
\end{figure}

\subsubsection{The Effect of MSCB}
The role of MSCB in TDNet is to increase the depth and width of the network to improve the ability of feature extraction. Its innovation lies in the use of convolution kernels with different scales to learn various scales in real scenes. As shown in Fig.~\ref{fig:abla_mscb}, to explore whether such MSCB can favor the fusion task, we change the original multi-scale convolution kernels (\textit{i.e.}, $3\times 3$, $5\times 5$ and $7\times 7$) to a single-scale convolution kernel (\textit{i.e.}, $5\times 5$). The reduced structure is called single-scale convolution block (SSCB). Tab.~\ref{tab:ablation} reports that the TDNet with SSCB has lower performance with respect to the original TDNet with MSCB, which verifies the benefits of using the MSCB module.

\subsubsection{The Effect of PAN Branch}
In the original network structure, we designed the PAN branch to extract the spatial features from the PAN image. These spatial features are then injected into the fusion branch. Hence, we alternatively feed the fusion branch directly using the PAN image and its downsampled version to explore whether such PAN branch can affect the final outcomes. We denoted the TDNet without PAN branch as w/o PAN branch. The quantitative results shown in Tab.~\ref{tab:ablation} demonstrate that the original TDNet with the PAN branch yields the best performance.

\subsubsection{The Effect of Double-level Structure}
In the work, the proposed TDNet has two levels. In each level, the MS image is upsampled to its double size. The fusion performance can benefit from the use of the double-level structure. To corroborate this point, we implemented a single-level (directly upsampling by a factor of 4) strategy using our TDNet approach. Tab.~\ref{tab:ablation} clearly shows that the single-level structure will significantly reduce the fusion performance demonstrating the advantages of the double-level structure.

\subsubsection{The Effect of PixelShuffle}
In the fusion branch, we introduce PixelShuffle to upscale LRMS images, instead of the more common interpolation or deconvolution operations. To demonstrate the superiority of PixelShuffle, we replaced PixelShuffle with linear interpolation and deconvolution upsampling, denoted as TDNet(bilinear) and TDNet(Deconv), respectively. The results of the variants are reported in Tab.~\ref{tab:ablation}. It can be seen that TDNet performs better with the assistance of PixelShuffle.

\subsubsection{The Effectiveness of the Overall strcture}
Compared with other DL-based methods, the results of TDNet are significantly improved. To prove the effectiveness of the TDNet structure more fairly, we reduce the number of channels in the original MSCB as showed in Fig.~\ref{fig:abla_mscb}, aiming at bring the model complexity to the level of FusionNet and DMDNet. The degenerate model is denoted as TDNet(-). The results of TDNet(-) are shown in Tab.~\ref{tab:ablation}. Compared with the results obtained by other comparative methods shown in Tab.~\ref{tab:1258}. It is clear that although the performance of TDNet(-) has degraded, it still outperforms all the compared DL-based methods. This is also a proof about the superiority of the triple-double structure.
 \subsection{Comparison of MRAB and Traditional MRA Scheme}
	
	An important module in our model, MRAB, is a derivative of the traditional MRA scheme. The traditional MRA scheme requires manual estimation of its injection coefficient. These handcrafted coefficient with subjective and relatively simple forms, however, always could not sufficiently
	and adaptively reflect complex spatial and spectral relationships among PAN image, LRMS image, and HRMS image. In addition, in the framework of deep learning, traditional schemes need to be improved to learn more discriminate representations from big data. In order to verify the rationality of our motivation, we conducted experiments to compare MRAB and traditional MRA scheme.
	
	Specifically, one competing MRA method, GLP-HPM~\cite{atwt}, is exploited here. Its generalized form is the same as equation (\ref{eq:mra}). GLP-HPM adopts the injection model as follows:
	\begin{equation}\label{eq:g_hpm}
	\mathbf{G_{k}} = \mathbf{\frac{\widetilde{MS}_{k}}{P_{L}}}, k=1,...,c.
	\end{equation}
	where $\mathbf{G_{k}}$ represents the injection gain in the $\mathbf{k}$-th band, and the division is intended as pixel-wise. The main difference between this model and our proposed MRAB~(\ref{eq:mra2}) is
	that the latter uses the injection gain learned by designed convolutional layers. By replacing the proposed MRAB as following, we can obtain the TDNet-TMRA method:
	
	\begin{equation}\label{eq:g_hpm2}
	\begin{split}
	&\mathbf{D} = \mathcal{H}(\mathbf{P}),\\
	&\widehat{\mathbf{MS}}_\mathbf{{k}} = \widetilde{\mathbf{MS}}_\mathbf{{k}} + \mathbf{\frac{\widetilde{MS}_{k}}{P_{L}}}\odot \mathbf{D_{k}}, k=1,...,c.
	\end{split}
	\end{equation}
	
	Table~\ref{tab:ablation} shows the average results produced by TDNet-TMRA and the proposed TDNet over
	1258 testing samples. The numerical results on four metrics show that the MRAB is indeed effective for TDNet, while it is hard to achieve good performance by embedding a traditional MRA scheme into our network. Therefore, when combining traditional methods with DL-based methods, how to overcome the uncertainty brought by deep learning deserves more in-depth research. In this work, we provide a feasible method, \textit{i.e.}, MRAB, uses unique network structures such as the attention mechanism to make the parameters adaptive to the data.

\subsection{Discussion}\label{sec:discussion}

\subsubsection{Evaluation on 4-band Datasets} \label{sec:4-band}
\begin{figure*}[t]
	\scriptsize
	\setlength{\tabcolsep}{0.9pt}
	\centering
	
	\begin{tabular}{cccccccc}
		{\includegraphics[width=0.85in,height=0.85in]{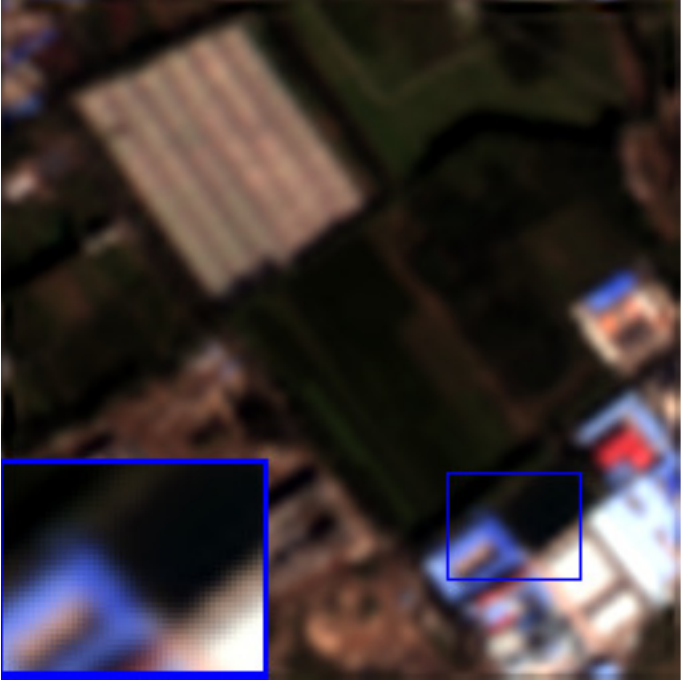}}&
		{\includegraphics[width=0.85in,height=0.85in]{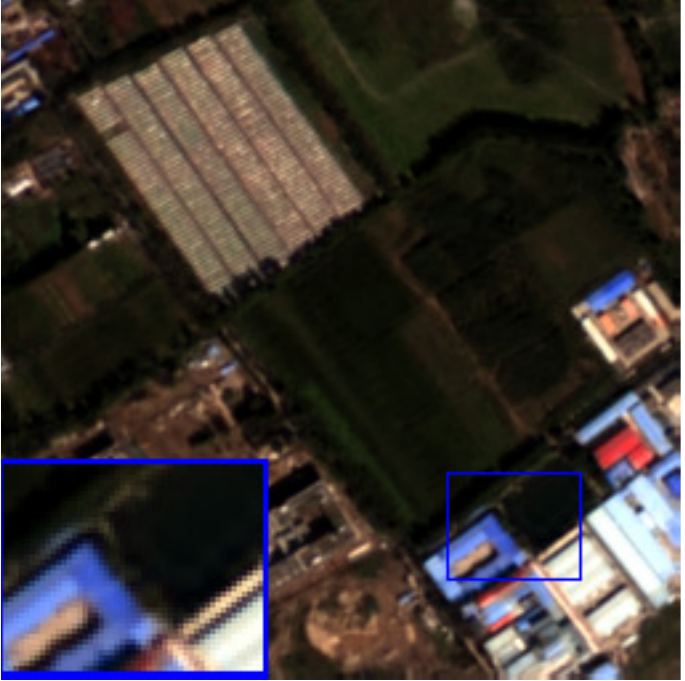}}&
		{\includegraphics[width=0.85in,height=0.85in]{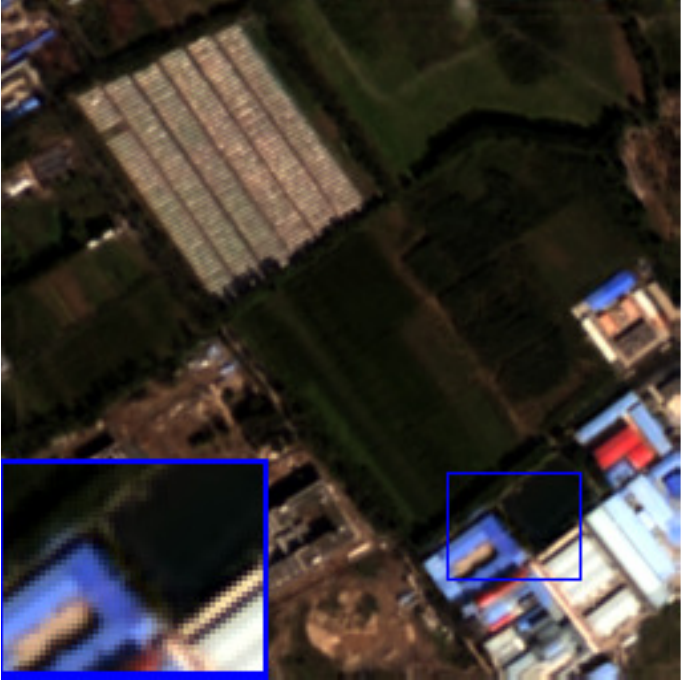}}&
		{\includegraphics[width=0.85in,height=0.85in]{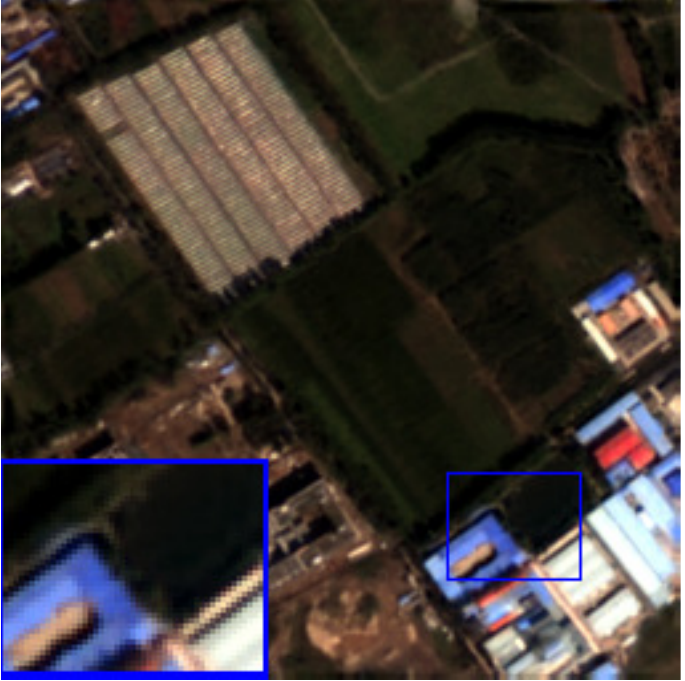}}&
		{\includegraphics[width=0.85in,height=0.85in]{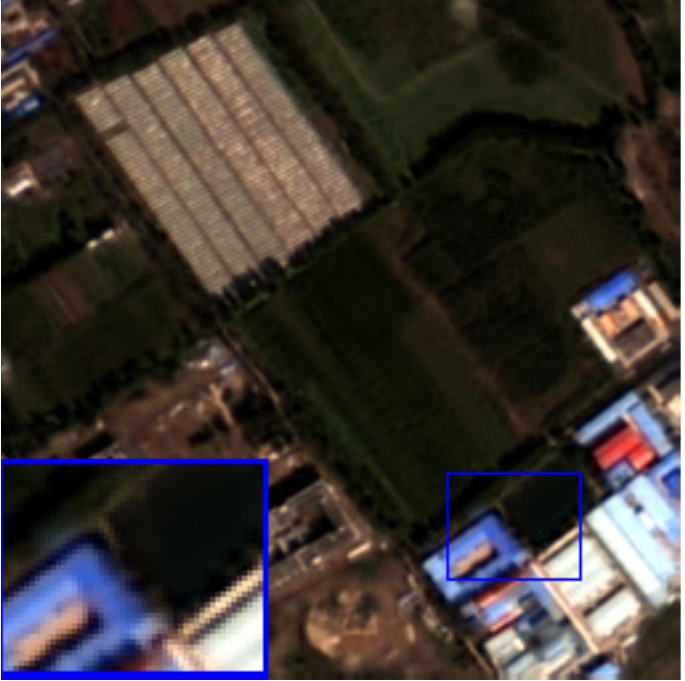}}&
		{\includegraphics[width=0.85in,height=0.85in]{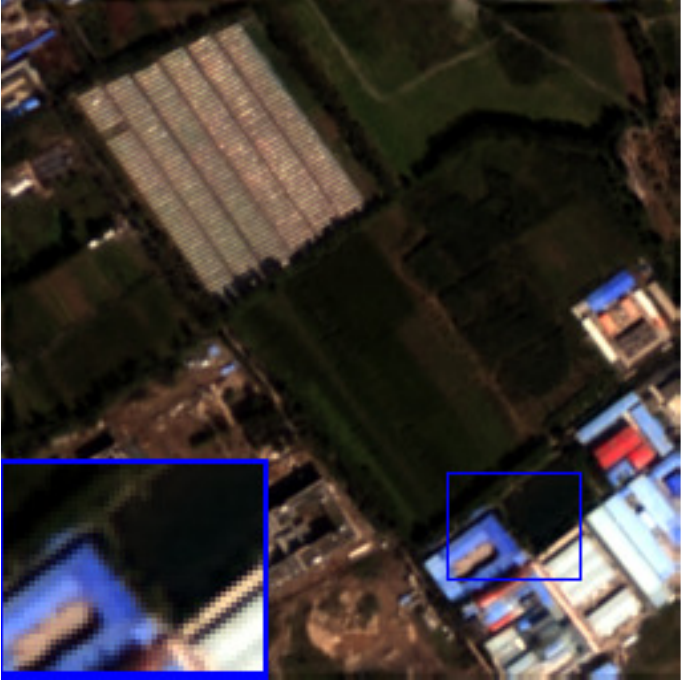}}&
		{\includegraphics[width=0.85in,height=0.85in]{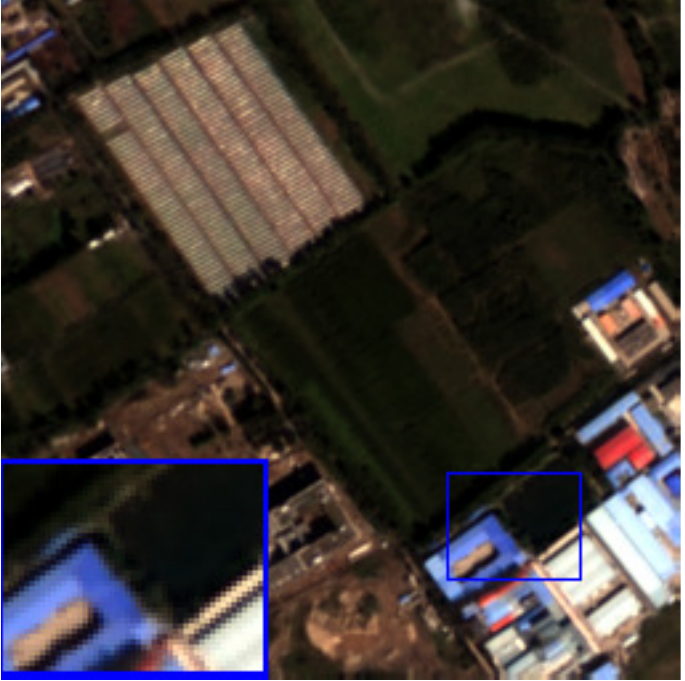}}&
		{\includegraphics[width=0.85in,height=0.85in]{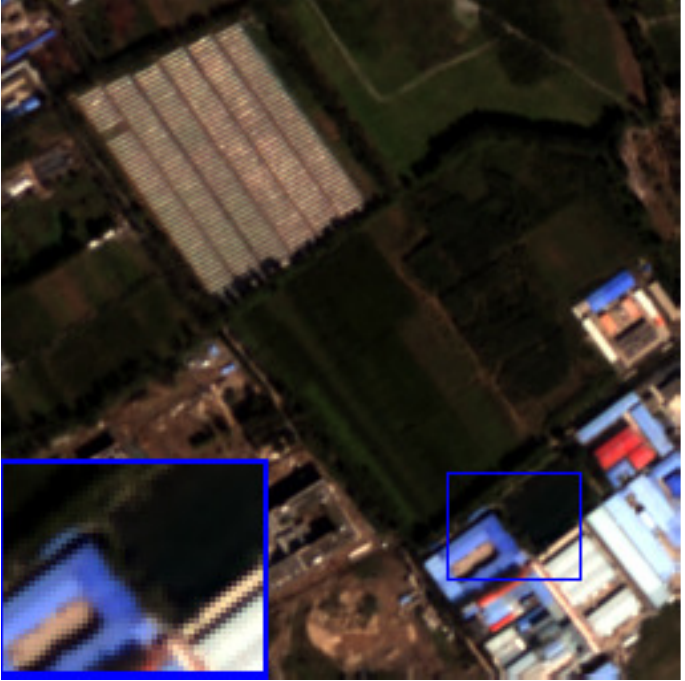}}\\
		EXP&PNN&DiCNN1& PanNet& BDPN& DMDNet& FusionNet& TDNet\\
		
		{\includegraphics[width=0.85in,height=0.85in]{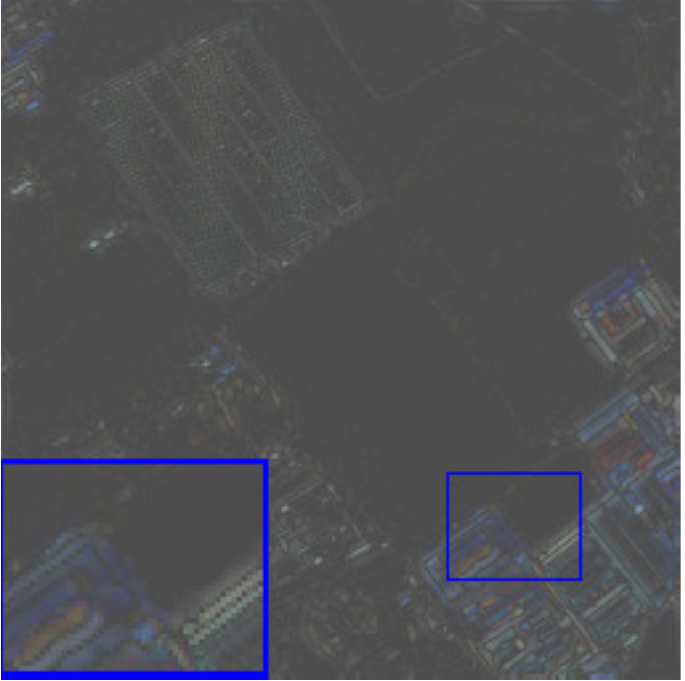}}&
		{\includegraphics[width=0.85in,height=0.85in]{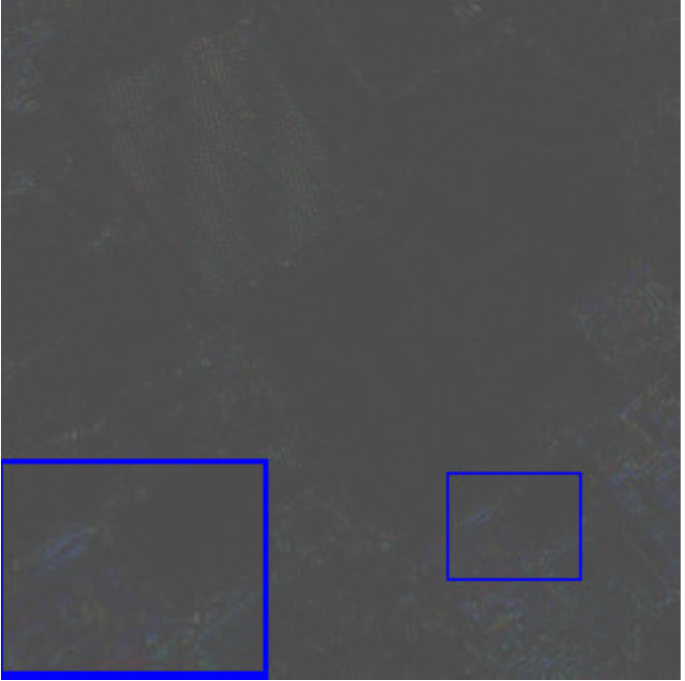}}&
		{\includegraphics[width=0.85in,height=0.85in]{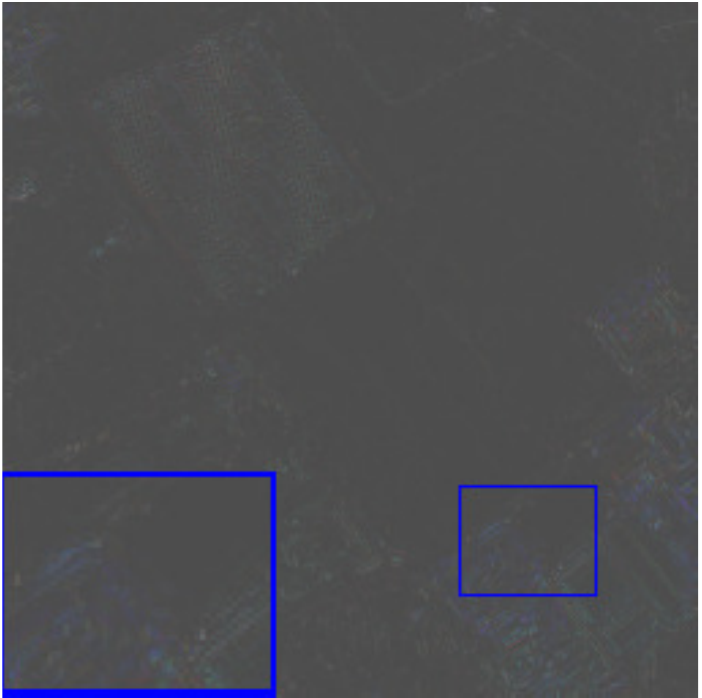}}&
		{\includegraphics[width=0.85in,height=0.85in]{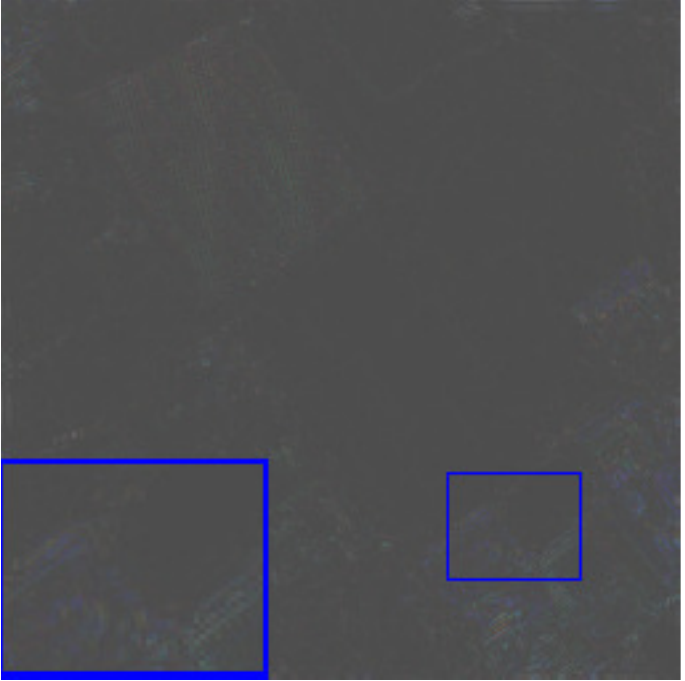}}&
		{\includegraphics[width=0.85in,height=0.85in]{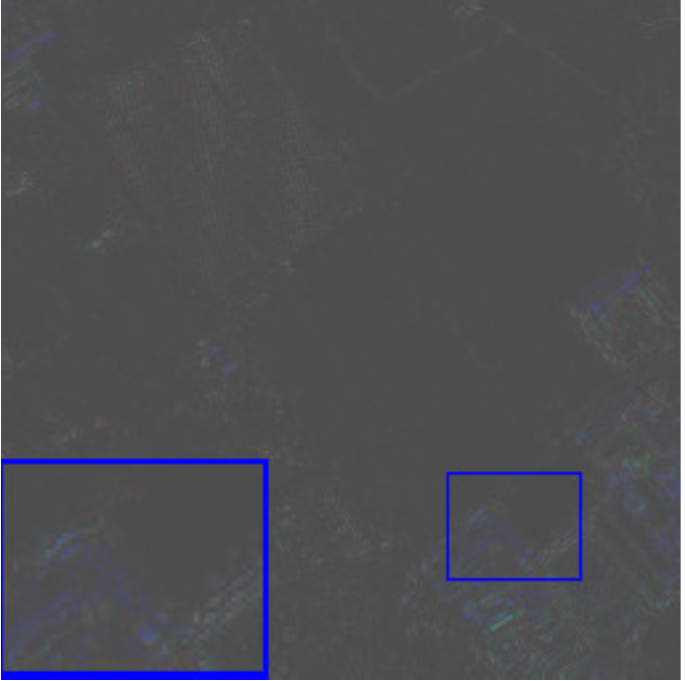}}&
		{\includegraphics[width=0.85in,height=0.85in]{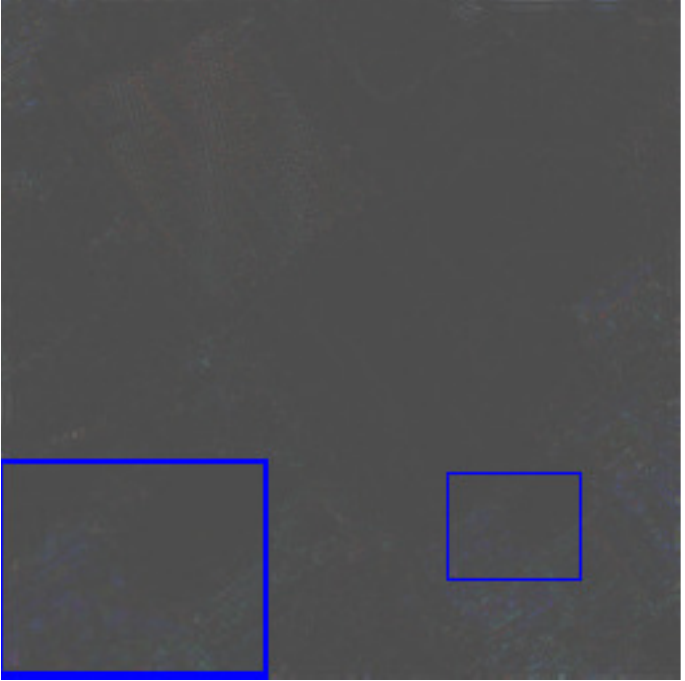}}&		
		{\includegraphics[width=0.85in,height=0.85in]{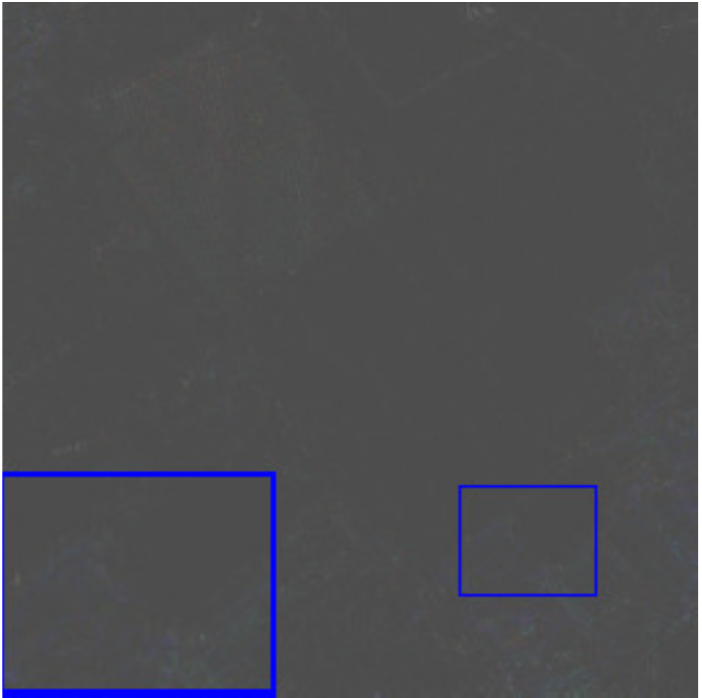}}&
		{\includegraphics[width=0.85in,height=0.85in]{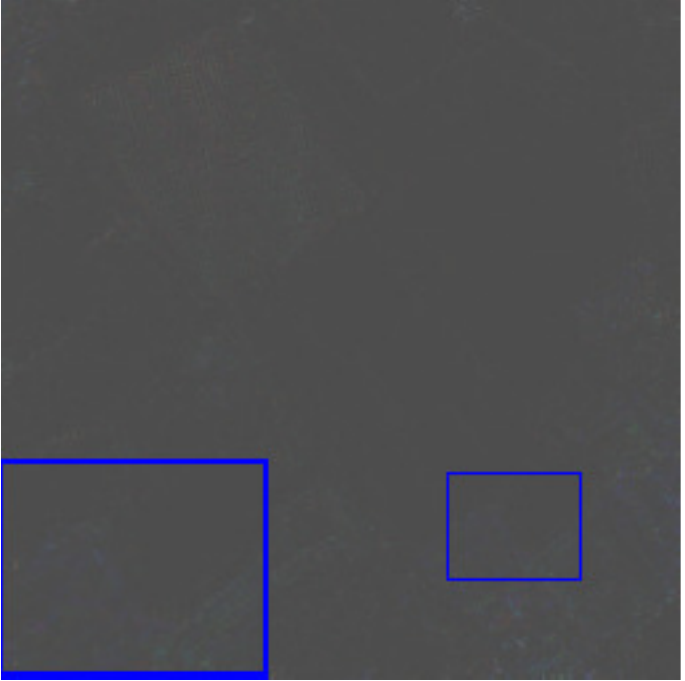}}\\
		EXP&PNN&DiCNN1& PanNet& BDPN& DMDNet& FusionNet& TDNet
	\end{tabular}
	
	\caption{Visual comparisons between the TDNet and the seven DL-based methods on the Guangzhou datasets (sensor: GaoFen-2). For orderly display, we show the GT image in Fig.~\ref{ref}.}
	\label{fig:gf_data2_new}
\end{figure*}

\begin{figure*}[t]
	\scriptsize
	\setlength{\tabcolsep}{0.9pt}
	\centering
	\begin{tabular}{cccccccc}
		
		{\includegraphics[width=0.85in,height=0.85in]{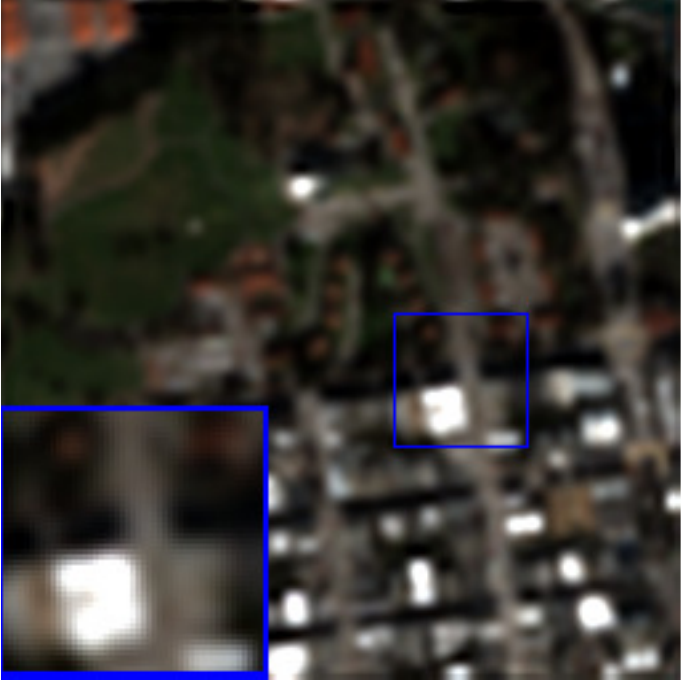}}&
		\includegraphics[width=0.85in,height=0.85in]{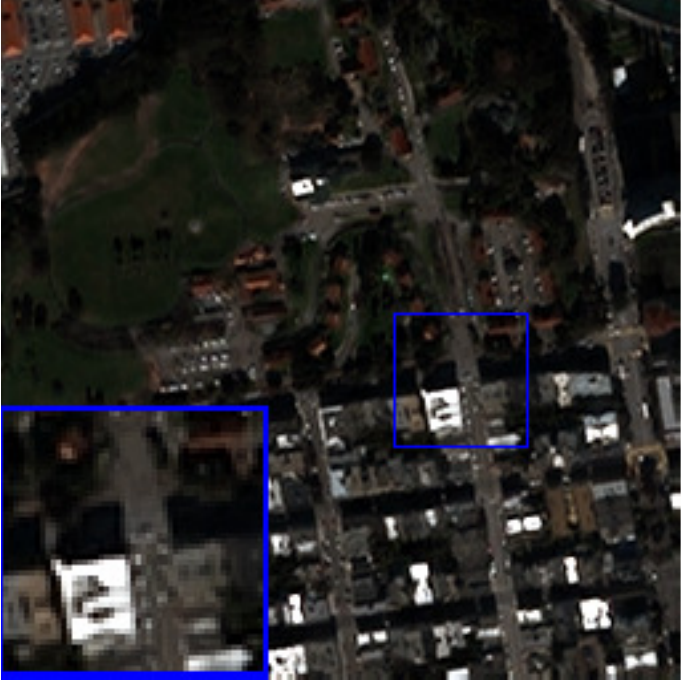}&
		\includegraphics[width=0.85in,height=0.85in]{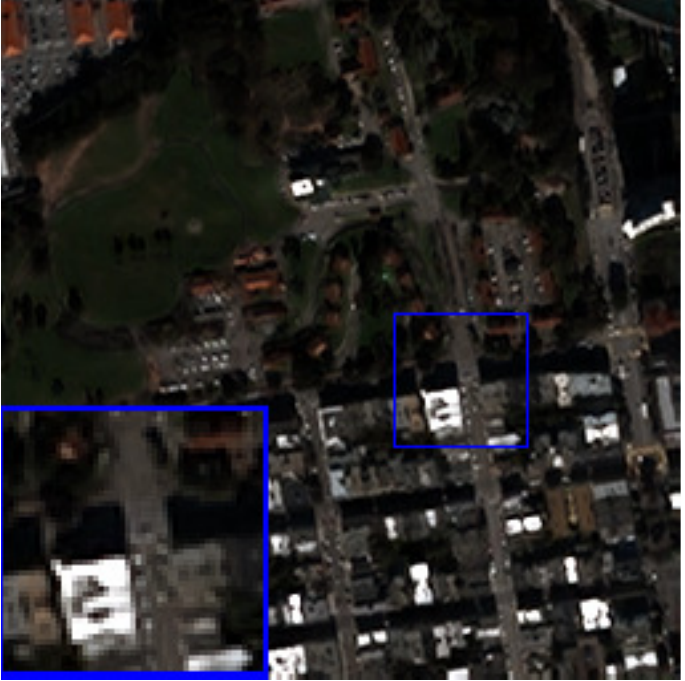}&
		\includegraphics[width=0.85in,height=0.85in]{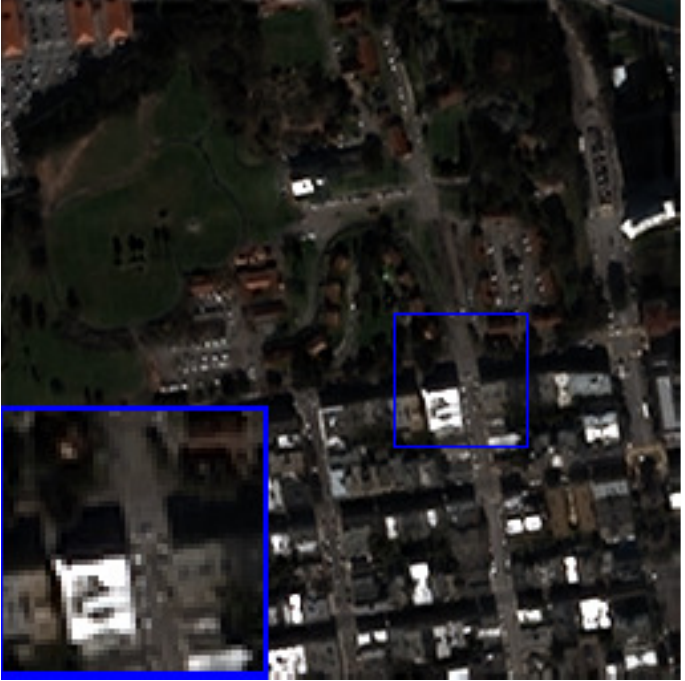}&
		\includegraphics[width=0.85in,height=0.85in]{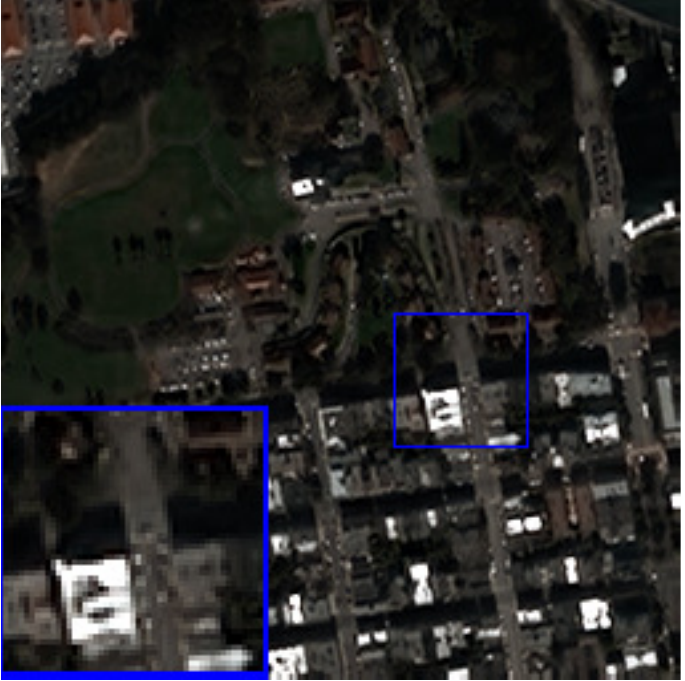}&
		\includegraphics[width=0.85in,height=0.85in]{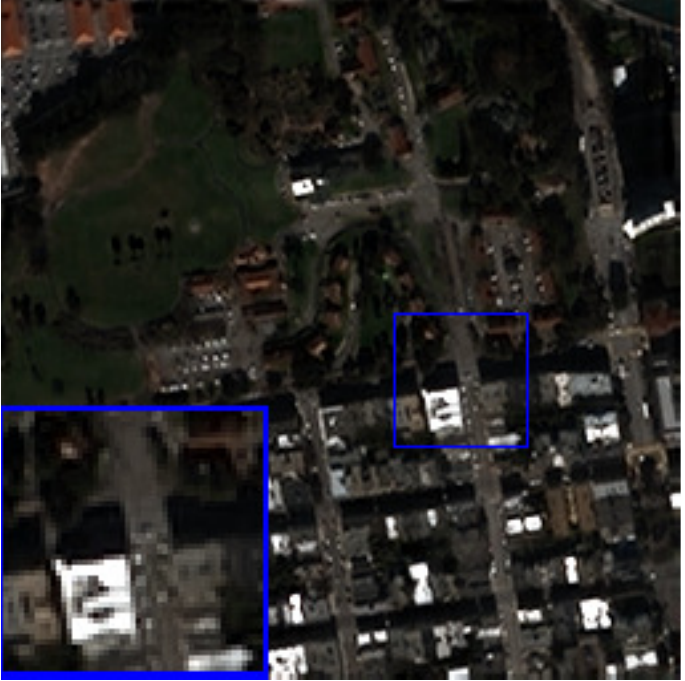}&
		\includegraphics[width=0.85in,height=0.85in]{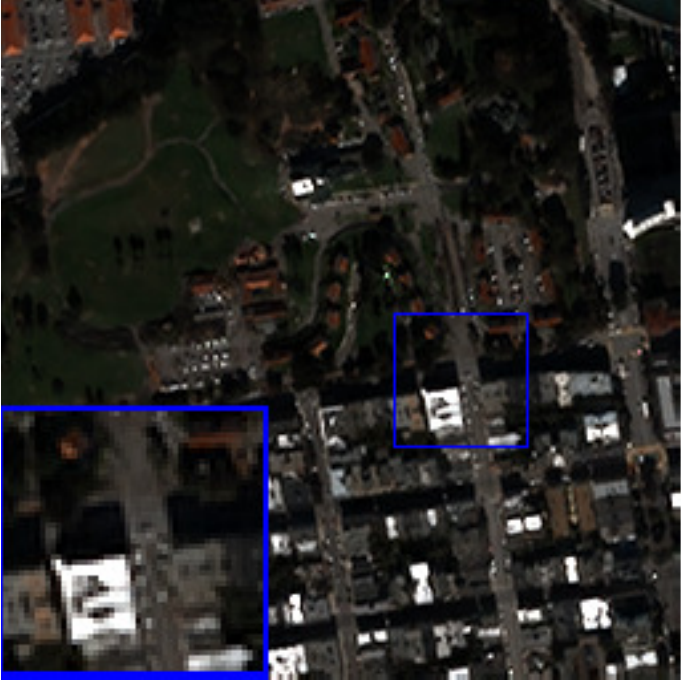}&
		\includegraphics[width=0.85in,height=0.85in]{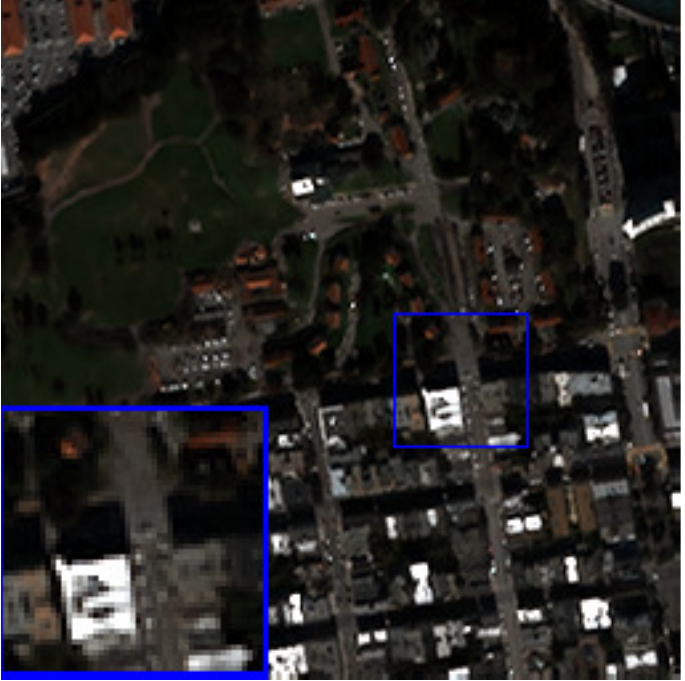}\\
		EXP&PNN&DiCNN1& PanNet&BDPN& DMDNet& FusionNet& TDNet\\
		
		{\includegraphics[width=0.85in,height=0.85in]{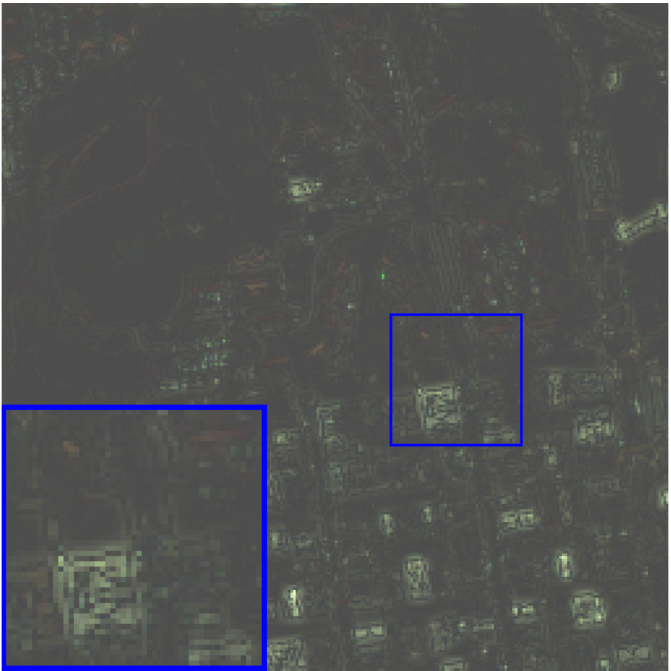}}&
		\includegraphics[width=0.85in,height=0.85in]{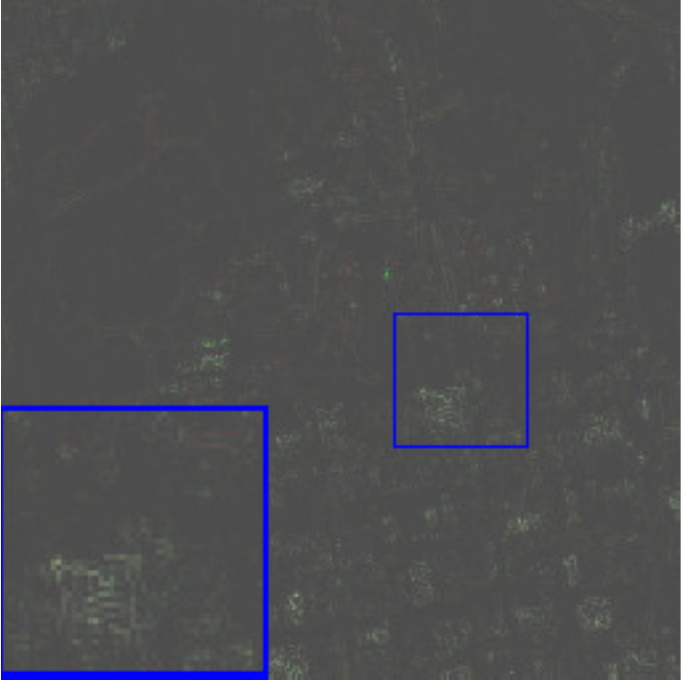}&
		\includegraphics[width=0.85in,height=0.85in]{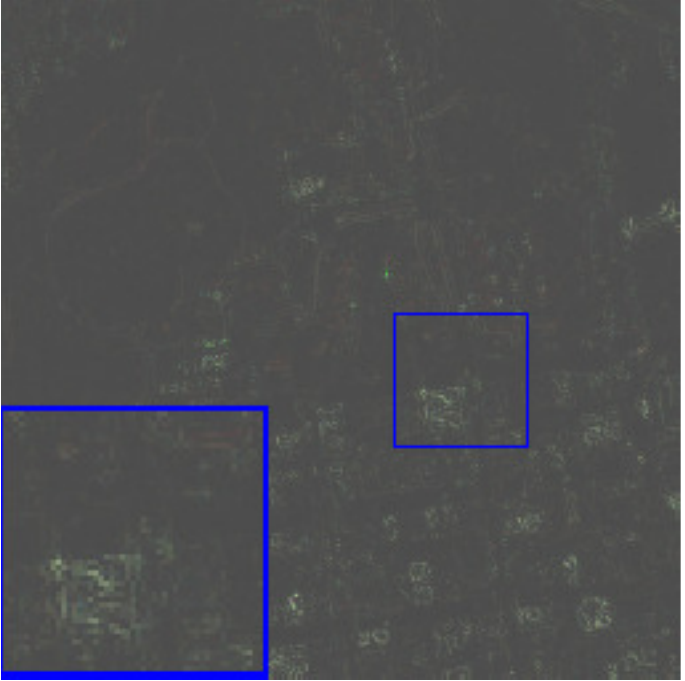}&
		\includegraphics[width=0.85in,height=0.85in]{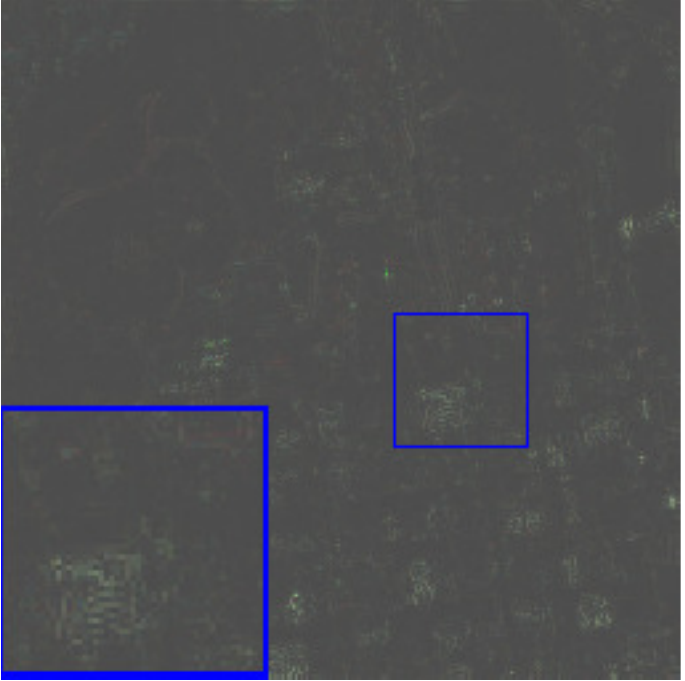}&
		\includegraphics[width=0.85in,height=0.85in]{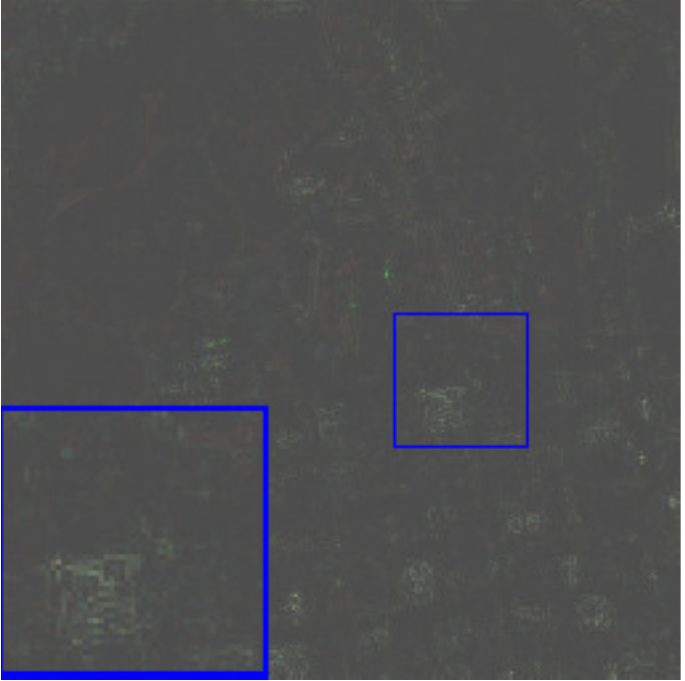}&
		\includegraphics[width=0.85in,height=0.85in]{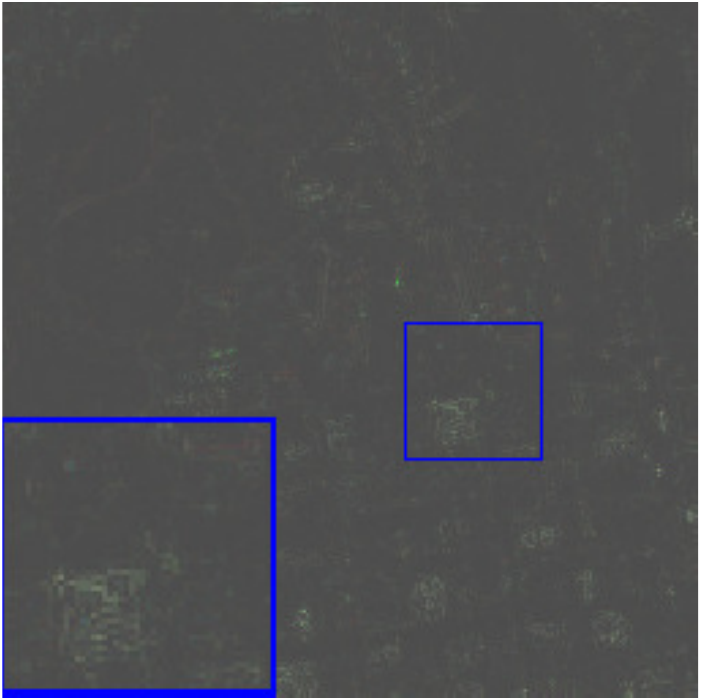}&		
		\includegraphics[width=0.85in,height=0.85in]{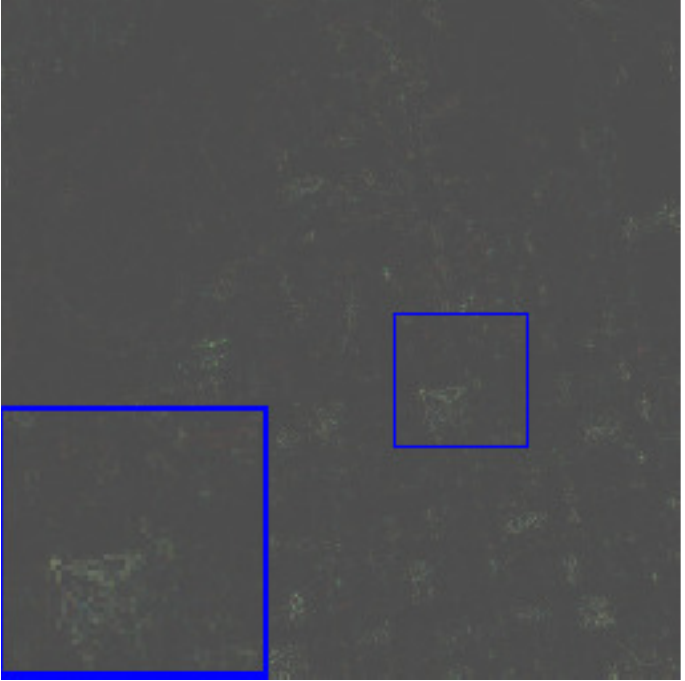}&
		\includegraphics[width=0.85in,height=0.85in]{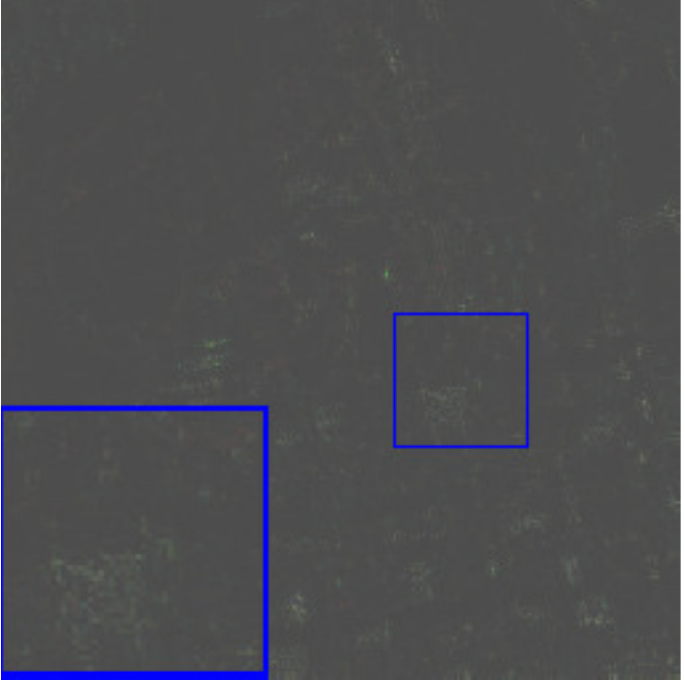}\\
		EXP&PNN&DiCNN1& PanNet&BDPN& DMDNet& FusionNet& TDNet
	\end{tabular}
	
	\caption{Visual comparisons between the TDNet and the seven DL-based methods on the Indianapolis datasets (sensor: QuickBird). For orderly display, we show the GT image in Fig.~\ref{ref}.}
	\label{fig:qb_data1_new}
\end{figure*}

The datasets used in the above experiments are all data with 8 spectral bands acquired by the same sensor (\textit{i.e.}, WorldView-3). In this section, we will focus on assessing performance on 4-band datasets acquired by the GaoFen-2 and the QuickBird sensors. The dataset simulation for the two sensors is the same as that of the 8-band WorldView-3 datasets mentioned in Sect. \ref{Sec-Train}. For the GaoFen-2 dataset, we downloaded the data acquired over Beijing from the website\footnote{Data link: \url{http://www.rscloudmart.com/dataProduct/sample}} and we simulated 21,607 training samples (PAN size, $64\times 64$). Besides, 81 testing samples (PAN size, $256\times 256$) acquired over the city of Guangzhou are used for comparison purposes. About the QuickBird test case, a large dataset acquired over the city of Indianapolis is exploited to simulate 20,685 training samples (PAN size, $64\times 64$). Moreover, we simulated 48 testing samples with spatial size $256\times 256$ to assess the performance for all the compared approaches. More details about the generation of these test cases can be found in \cite{fusionnet}. In Figs.~\ref{fig:gf_data2_new} and~\ref{fig:qb_data1_new}, we show the performance comparing all the five DL-based approaches\footnote{Note that, since the traditional methods, \textit{i.e.}, MRA and CS methods, have obtained lower performance, for the sake of brevity, we excluded them from the analysis.}. Since it is not easy to distinguish the differences having a look at the 8-bits RGB images, we present the absolute error maps (AEMs). It is worth to be remarked that our TDNet generates more details showing less residuals. The quantitative results of Tab. \ref{tab:gf_qb} also support the conclusion that the TDNet obtains the best overall performance.

\begin{table*}[h]
	\caption{\small{Average assessment of the compared approaches for 81 GaoFen-2 testing samples and 48 QuickBird testing samples. (Bold: best; Underline: second best)} \label{tab:gf_qb}}
	\footnotesize
	\begin{center}
		\resizebox{0.68\textwidth}{!}{  
			\begin{tabular}{ l|cccc}
				\hline
				&\emph{SAM} ($\pm$ std)&\emph{ERGAS} ($\pm$ std)& \emph{Q8} ($\pm$ std)& \emph{SCC} ($\pm$ std)\\
				\hline\hline
				\multicolumn{5}{c}{\emph{Guangzhou datasets (GaoFen-2)}}\\ 
				\hline
				\textbf{PNN}  &1.6599 $\pm$ 0.3606  & 1.5707 $\pm$ 0.3243 & 0.9274 $\pm$ 0.0202  & 0.9281 $\pm$ 0.0206 \\
				\textbf{DiCNN1}  &1.4948 $\pm$ 0.3814  & 1.3203 $\pm$ 0.3543  & 0.9445 $\pm$ 0.0211  & 0.9458 $\pm$ 0.0222  \\
				\textbf{PanNet}  &1.3954 $\pm$ 0.3261 &   1.2239 $\pm$ 0.2828  & 0.9468 $\pm$ 0.0222  &  0.9558 $\pm$ 0.0123   \\
				\textbf{DMDNet}  &1.2968 $\pm$ 0.3156 &   1.1281 $\pm$ 0.2669  & 0.9529 $\pm$ 0.0218 &  0.9644 $\pm$ 0.0100   \\
				\textbf{FusionNet}  &\underline{1.1795 $\pm$ 0.2714}  & \underline{1.0023 $\pm$ 0.2271}  &\underline{0.9627 $\pm$ 0.0167}  & \underline{0.9710 $\pm$ 0.0074}\\
				\textbf{TDNet}  &\textbf{1.0926 $\pm$ 0.2645}  &\textbf{ 0.9303 $\pm$ 0.2267}  & \textbf{0.9695 $\pm$ 0.0131}  & \textbf{0.9750 $\pm$0.0132} \\
				\hline\hline
				\multicolumn{5}{c}{\emph{Indianapolis datasets (QuickBird)}}\\ 
				\hline
				\textbf{PNN}  & 5.7993 $\pm$ 0.9474 & 5.5712 $\pm$ 0.4584 & 0.8572 $\pm$ 0.1481 & 0.9023 $\pm$ 0.0489 \\
				\textbf{DiCNN1}  &5.3071 $\pm$ 0.9957  & 5.2310 $\pm$ 0.5411  & 0.8821 $\pm$ 0.1431  & 0.9224 $\pm$ 0.0506 \\
				\textbf{PanNet}  &5.3144 $\pm$ 1.0175 &   5.1623 $\pm$ 0.6814  & 0.8833 $\pm$ 0.1398  &  0.9296 $\pm$ 0.0585   \\
				\textbf{DMDNet}  &5.1197 $\pm$  0.9399  &   4.7377 $\pm$  0.6486  & 0.8907 $\pm$  0.1464 &  0.9350 $\pm$  0.0652   \\
				\textbf{FusionNet}  &\underline{4.5402 $\pm$ 0.7789} & \underline{4.0508 $\pm$ 0.2666} & \underline{0.9102 $\pm$ 0.1364}  & \underline{0.9547 $\pm$ 0.0457}\\
				\textbf{TDNet}  &\textbf{4.5047 $\pm$ 0.8022}  & \textbf{3.9799 $\pm$ 0.2326}  & \textbf{0.9123 $\pm$ 0.1452}  & \textbf{0.9551 $\pm$ 0.0652}\\
				\Xhline{1pt}
				\textbf{Ideal value}&\bf{0}&\bf{0}&\bf{1}&\bf{1}\\ 
				\Xhline{1pt}
		\end{tabular}}
	\end{center}
\end{table*}

\begin{figure}[!t]
	\begin{center}
	\begin{tabular}{ccc}
		
		{\includegraphics[width=1in,height=1in]{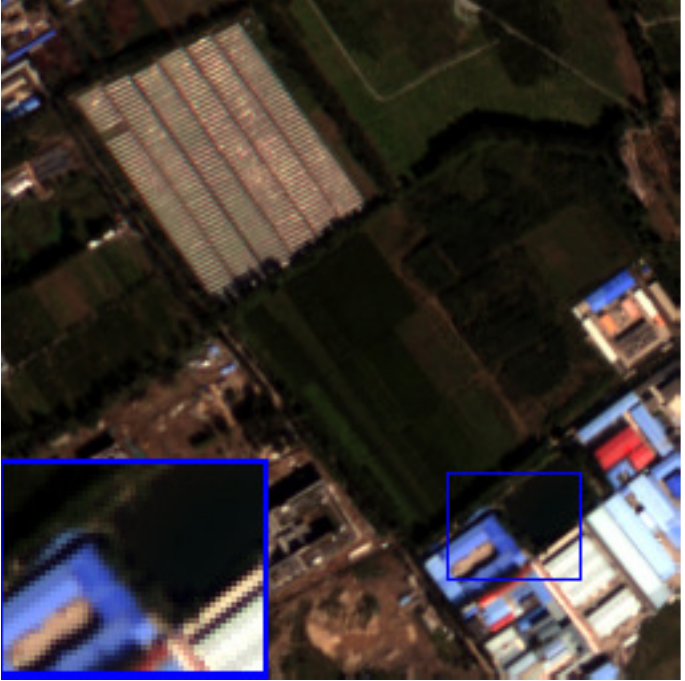}}&
		{\includegraphics[width=1in,height=1in]{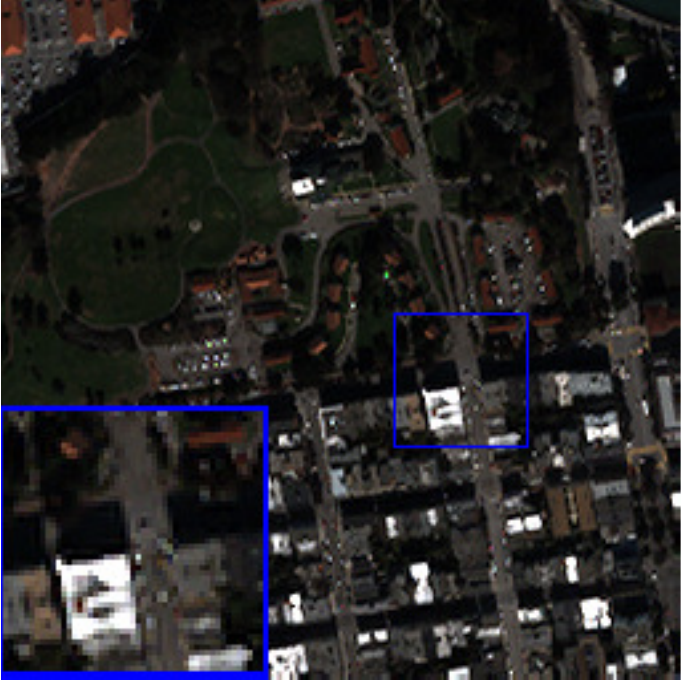}}&
		{\includegraphics[width=1in,height=1in]{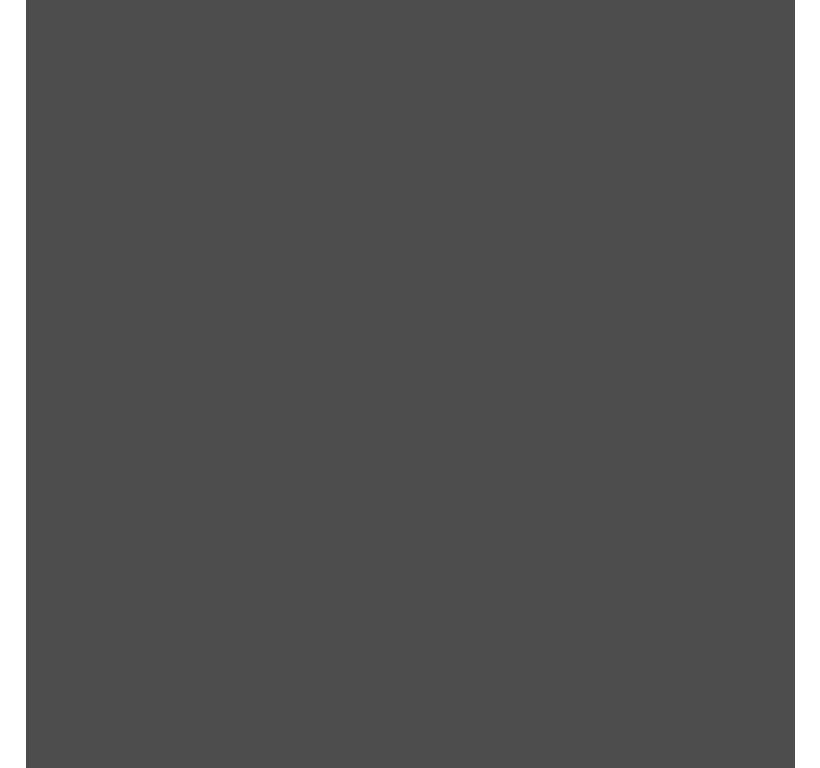}}\\
		(a)&(b)&(c)
	\end{tabular}
	\end{center}
	\caption{Reference (GT) image for 4-band datasets. (a) Guangzhou dataset (sensor: GaoFen-2) (b) Indianapolis dataset (sensor: QuickBird) (c) The AEM on GT images. }\label{ref}
\end{figure}
\subsubsection{The Hyper-Parameter in the Loss function}\label{sec:gamma}
\begin{figure}[h]
	
	\begin{center}
		{\includegraphics[width=0.6\linewidth]{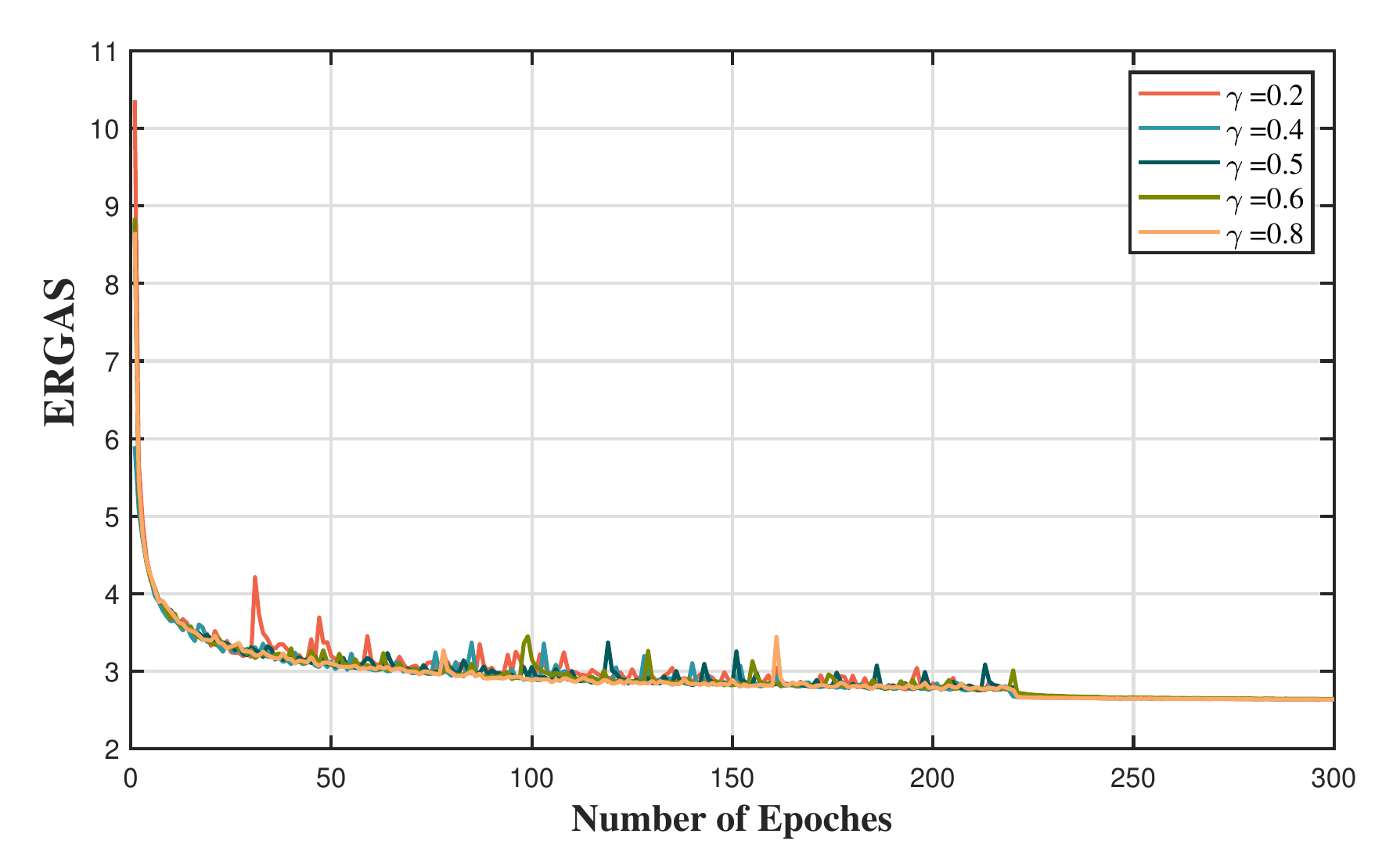}}
	\end{center}
	
	\caption{ERGAS averaged on WorldView-3 test cases for $\gamma = \{0.2, 0.4, 0.5, 0.6, 0.8\}$. Since the convergences are poor when $\gamma = 0$ and $\gamma = 1$, we decided to avoid plotting them.}\label{fig:lambda}
\end{figure}

As described in Sect. \ref{lossfuction}, the loss function consists of two parts, in which the hyper-parameter $\gamma$ weights the two sub-loss functions. Obviously, the higher the value of $\gamma$, the more the importance to the first level. The goal is to generate the final fusion image closer to the reference image. Therefore, it is worth exploring how a change in the value of $\gamma$ can lead to better results. In our experiment, $\gamma$ is set to different values. Fig.~\ref{fig:lambda} shows the changes in the ERGAS index varying $\gamma$ and increasing the epochs. When  $\gamma$ = 0 or $\gamma$ = 1, the convergence is poor. Thus, we discard these two cases. Fig.~\ref{fig:lambda} shows comparable results, in terms of the convergence speed and values, varying $\gamma$ (\textit{i.e.}, assuming the values 0.2, 0.4, 0.5, 0.6, 0.8). Thus, we choose $\gamma = 0.4$ for the training of the proposed TDNet.

%

\begin{table*}[!t]
	\caption{Comparison of the number of parameters (NOPs), the iterations, the training times, and the testing times for all the DL-based approaches. (the training times unit is hours: minutes \& the testing times unit is seconds)} \label{tab:para_flops}
	\footnotesize
	\begin{center}
		\begin{tabular}{c|ccccccc}
			\Xhline{1pt}
			& \textbf{PNN}&	\textbf{DiCNN1} &\textbf{PanNet} &\textbf{BDPN} &\textbf{DMDNet}	 &\textbf{FusionNet} & \textbf{TDNet}\\
			\hline
			\textbf{Iterations} &\textbf{$1.12\times 10^6$}  &\textbf{$3\times 10^5$} &\textbf{$2.4\times 10^5$} & \textbf{$2.7\times 10^5$}  &  \textbf{$2.5\times 10^5$}  &\textbf{$1.4\times 10^5$}  &\textbf{$8.2\times 10^4$} (300 epoches)\\
			\hline	
			\textbf{Training times} & 	25: 15  &	7: 06 &  4: 32  &  46:19 &  5: 27	 & 2: 21  & 	6: 30 \\
			\hline
			\textbf{Testing times} & 	0.0778  &0.0799 & 0.0811 &  0.0912  &  0.0852	 & 0.0812  &  0.0861\\
			\hline
			\textbf{NoPs}	&\textbf{$3.1\times 10^5$}   &\textbf{$1.8\times 10^5$}  & \textbf{$2.5\times 10^5$}   &   \textbf{$15.2\times 10^5$}  &  \textbf{$3.2\times 10^5$} &\textbf{$2.3\times 10^5$}&\textbf{$5.5\times 10^5$} \\
			\Xhline{1pt}
		\end{tabular}
	\end{center}
\end{table*}

\subsubsection{The Computational Analysis} \label{sec:para}
Tab.~\ref{tab:para_flops} reports the training time and the number of parameters (NoPs) for all the compared DL-based methods. The maximum number of iterations shown in the table is the optimal one for training the network. TDNet gets a relatively large amount of parameters, mainly due to the structure of the MSCB. However, the final training time of TDNet is less than that of PNN and DiCNN1 because of a less iteration number for the convergence.  Besides, TDNet is able to achieve a satisfying trade-off between effectiveness and complexity. We perform the evaluation on 1258 testing samples with size 256 $\times$ 256 acquired by the WorldView-3 sensor, as described in Sect.~\ref{dataset}. Comparisons on average testing time are shown in Tab.~\ref{tab:para_flops}. It is can be seen that the testing time of TDNet is not much longer than the compared deep learning methods, and significantly shorter than BDPN, which is also a double-level structure. Leveraging on the special structure, our network can fully fuse the complementary information from different sources in a more reasonable way, which leads to good results with the tolerable computational burden.

\subsubsection{Structure Discussion and Improvements Analysis} \label{sec:dis}
 In fact: the bidirectional, double-branch network structure and feature pyramid~\cite{Lai_2017_CVPR} has appeared in several previous significant works~\cite{hui2016depth,2019BDPN}, and has been proved that can implement feature extraction and image fusion hierarchically and more effectively. In particular, it is necessary to emphasize the distinction between the proposed TDNet and BDPN~\cite{2019BDPN}. Aiming at making full use of the high-frequency information in PAN images, BDPN extracts the multilevel details from PAN images and directly injects them into the upsampled LRMS images. Differently, we focus more on the mapping relations among images. Specifically, the extracted high-frequency information is adopted as the input of the fusion branch in multiple stages, and the non-linear ``pixel-to-pixel'' mapping is learned in the designed MRAB, which ensures a reasonable fusion. Besides, we choose the multi-scale convolution module (MSCB) to be used as a component of the network, which could achieve the purpose of increasing the receptive field while avoiding deep convolution layers of the TDNet. This can also explain why the number of parameters of TDNet is much smaller than that of the BDPN. 

\section{Conclusions}\label{sec:conclusion} 
In this paper, we propose a novel deep neural network architecture for pansharpening, the so-called triple-double network (TDNet), by taking into account the following three double-type structures, \textit{i.e.}, double-level, double-branch, and double-direction. By exploiting the structure of the TDNet, the spatial details of the panchromatic image can be fully exploited and progressively injected into the LRMS image yielding a final MS image with high spatial resolution. Motivated by the traditional MRA formula, an effective MRA block was integrated into the TDNet. Furthermore, the MSCB with few ResNet blocks and some multi-scale convolution kernels was also used to deepen and widen the network, aiming to effectively enhance the feature extraction and robustness of the proposed TDNet. Extensive experiments on reduced and full resolution examples, acquired by WorldView-3, QuickBird, and Gaofen-2 sensors, demonstrate the superiority of the proposed method. In addition, several ablation studies and discussions are also conducted to corroborate the effectiveness of the proposed TDNet.

\section{Acknowledge}
The second and the third authors are supported by NSFC (61702083, 61772003), Key Projects of Applied Basic Research in Sichuan Province (Grant No. 2020YJ0216), and National Key Research and Development Program of China (Grant No. 2020YFA0714001).

\bibliographystyle{unsrt}  
\bibliography{tdnet_ArXiv}

\end{document}